\documentclass[utf8]{frontiersFPHY}
\pdfoutput=1
\setcitestyle{square} 
\usepackage{hyperref}       % hyperlinks
\hypersetup{
    colorlinks=true,
    linkcolor=blue,
    filecolor=magenta,      
    urlcolor=blue,
}
\usepackage{url}
\usepackage{booktabs}       
\usepackage{amsmath}       
\usepackage{nicefrac} 
\usepackage{microtype} 
\usepackage{lipsum}
\usepackage{graphicx}
\usepackage{amsmath}
\usepackage{multirow}
\usepackage{xspace}
\usepackage{setspace}
\usepackage{glossaries-prefix}
\usepackage[capitalise,noabbrev]{cleveref}
\usepackage{todonotes}
\usepackage{float}
\usepackage[toc,page]{appendix}
\usepackage{tcolorbox}

\newcommand{\unit}[1]{\ensuremath{\text{\,#1}}\xspace}
\newcommand{\hlsfml}{\texttt{hls4ml}\xspace}
\newcommand{\conifer}{\texttt{Conifer}\xspace}

\begin{document}

\title[Fast Machine Learning in Science]{Applications and Techniques for Fast Machine Learning in Science}
\def\keyFont{\fontsize{8}{11}\helveticabold }
\def\firstAuthorLast{McCarn Deiana, Tran, {et~al.}} 

\def\Authors{
Editors: \textnormal{
Allison McCarn Deiana\,$^{1}$ (coordinator), 
Nhan Tran\,$^{2,3}$ (coordinator), 
Joshua Agar\,$^{4}$,
Michaela Blott\,$^{5}$
Giuseppe Di Guglielmo\,$^{27}$,
Javier Duarte\,$^{23}$,
Philip Harris\,$^{15}$,
Scott Hauck\,$^{24}$,
Mia Liu\,$^{25}$,
Mark S. Neubauer\,$^{26}$, 
Jennifer Ngadiuba\,$^{2}$,
Seda Ogrenci-Memik\,$^{3}$,
Maurizio Pierini\,$^{6}$}

Report Contributors: \textnormal{
Thea Aarrestad\,$^{6}$,
Steffen B\"{a}hr\,$^{9}$,
J\"{u}rgen Becker\,$^{9}$,
Anne-Sophie Berthold\,$^{29}$,
Richard J. Bonventre \,$^{11}$,
Tom\'as E. M\"uller Bravo\,$^{18}$,
Markus Diefenthaler\,$^{41}$,
Zhen Dong\,$^{19}$,
Nick Fritzsche\,$^{29}$,
Amir Gholami\,$^{19}$,
Ekaterina Govorkova\,$^{6}$,
Kyle J Hazelwood\,$^{2}$,
Christian Herwig \,$^{2}$,
Babar Khan\,$^{21}$,
Sehoon Kim\,$^{19}$,
Thomas Klijnsma\,$^{2}$,
Yaling Liu\,$^{4}$,
Kin Ho Lo\,$^{8}$,
Tri Nguyen\,$^{15}$,
Gianantonio Pezzullo\,$^{10}$,
Seyedramin Rasoulinezhad\,$^{22}$,
Ryan A. Rivera\,$^{2}$,
Kate Scholberg\,$^{13}$,
Justin Selig\,$^{32}$
Sougata Sen\,$^{16}$,
Dmitri Strukov\,$^{20}$,
William Tang\,$^{7}$,
Savannah Thais\,$^{7}$,
Kai Lukas Unger\,$^{9}$,
Ricardo Vilalta\,$^{17}$,
Belina von Krosigk\,$^{14,9}$,
Thomas K. Warburton\,$^{12}$}

Community endorsers: \textnormal{
Maria Acosta Flechas\,$^{2}$,
Anthony Aportela\,$^{23}$,
Thomas Calvet\,$^{31}$,
Leonardo Cristella\,${6}$,
Daniel Diaz\,$^{23}$,
Caterina Doglioni\,$^{37}$,
Maria Domenica Galati\,${34}$,
Elham E Khoda\,$^{24}$,
Farah Fahim\,$^{2}$,
Davide Giri\,$^{27}$,
Benjamin Hawks\,$^{2}$,
Duc Hoang\,$^{15}$,
Burt Holzman\,$^{2}$,
Shih-Chieh Hsu\,$^{24}$,
Sergo Jindariani\,$^{2}$,
Iris Johnson\,$^{2}$,
Raghav Kansal\,$^{23}$,
Ryan Kastner\,$^{23}$,
Erik Katsavounidis\,$^{15}$,
Jeffrey Krupa\,$^{15}$,
Pan Li\,$^{25}$,
Sandeep Madireddy\,$^{40}$,
Ethan Marx\,$^{15}$,
Patrick McCormack\,$^{15}$
Andres Meza\,$^{23}$,
Jovan Mitrevski\,$^{2}$,
Mohammed Attia Mohammed\,$^{36}$,
Farouk Mokhtar\,$^{23}$,
Eric Moreno\,$^{15}$,
Srishti Nagu\,$^{35}$,
Rohin Narayan\,$^{1}$,
Noah Palladino\,$^{15}$,
Zhiqiang Que\,$^{38}$,
Sang Eon Park$^{15}$,
Subramanian Ramamoorthy\,$^{28}$,
Dylan Rankin\,$^{15}$,
Simon Rothman\,$^{15}$,
Ashish Sharma\,$^{30}$,
Sioni Summers\,$^{6}$,
Pietro Vischia\,$^{33}$,
Jean-Roch Vlimant\,$^{39}$,
Olivia Weng\,$^{23}$}
}

\def\Address{\footnotesize
$^{1}$Southern Methodist University, Dallas, TX 75205, USA,
$^{2}$Fermi National Accelerator Laboratory, Batavia, IL 60510, USA,
$^{3}$Northwestern University, Evanston, IL 60208, USA,
$^{4}$Lehigh University, University, Bethlehem, PA 18015, USA,
$^{5}$Xilinx Research, Dublin, D24 T683, Ireland,
$^{6}$European Organization for Nuclear Research (CERN), Meyrin, Switzerland,
$^{7}$Princeton University, Princeton, NJ 08544, USA,
$^{8}$University of Florida, Gainesville, FL 32611, USA,
$^{9}$Karlsruhe Institute of Technology, 76131 Karlsruhe, Germany,
$^{10}$Yale University, New Haven, CT 06520, USA,
$^{11}$Lawrence Berkeley National Laboratory, Berkeley, CA 94720, USA,
$^{12}$Iowa State University, Ames, IA 50011, USA,
$^{13}$Duke University, Durham, NC 27708, USA,
$^{14}$Universit{\"a}t  Hamburg,  22761  Hamburg,  Germany,
$^{15}$Massachusetts Institute of Technology, Cambridge, MA 02139, USA,
$^{16}$Birla Institute of Technology and Science, Pilani, Goa 403726, India,
$^{17}$University of Houston, Houston TX 77204, USA,
$^{18}$University of Southampton, Southampton SO17 1BJ, United Kingdom,
$^{19}$University of California Berkeley, Berkeley, CA 94720, USA,
$^{20}$University of California Santa Barbara, Santa Barbara, CA 93106, USA,
$^{21}$Technical University Darmstadt, Darmstadt 64289, Germany,
$^{22}$University of Sydney, Camperdown NSW 2006, Australia,
$^{23}$University of California San Diego, La Jolla, CA 92093, USA,
$^{24}$University of Washington, Seattle WA 47907, USA,
$^{25}$ Purdue University, West Lafayette IN 47907, USA,
$^{26}$University of Illinois Urbana-Champaign, Champaign IL 61820, USA,
$^{27}$Columbia University, New York, NY 10027, USA,
$^{28}$University of Edinburgh, Edinburgh EH8 9YL, United Kingdom ,
$^{29}$Technische Universit\"{a}t Dresden, 01062 Dresden, Germany,
$^{30}$Indian Institute of Technology Madras, Chennai 600 036, India,
$^{31}$Centre de Physique des Particules de Marseille, 13009 Marseille, France,
$^{32}$Cerebras Systems, Sunnyvale CA 94085, USA ,
$^{33}$Universit\'{e} Catholique de Louvain, B-1348 Louvain-la-Neuve, Belgium,
$^{34}$University of Groningen, 9747 AG Groningen, Netherlands,
$^{35}$Lucknow University, Lucknow 226007, U.P., India,
$^{36}$Center for High Energy Physics (CHEP-FU), Fayoum University, El-Fayoum, Egypt,
$^{37}$Lund University, SE-223 623 Lund, Sweden,
$^{38}$Imperial College London, London SW7 2BX, UK,
$^{39}$California Institute of Technology, Pasadena, CA 91125, USA,
$^{40}$Argonne National Laboratory, Lemont, IL 60439, USA,
$^{41}$Thomas Jefferson National Accelerator Facility, Newport News, VA 23606, USA}

\def\corrAuthor{Allison McCarn Deiana, Nhan Tran}
\def\corrEmail{adeiana@smu.edu, ntran@fnal.gov}
\onecolumn
% \firstpage{1}

\author[\firstAuthorLast ]{\Authors} %This field will be automatically populated
\address{} %This field will be automatically populated
\correspondance{} %This field will be automatically populated
\extraAuth{} % Don't delete this

\maketitle

\clearpage
\begin{abstract}
\section{}
In this community review report, we discuss applications and techniques for \textit{fast} machine learning (ML) in science---the concept of integrating power ML methods into the real-time experimental data processing loop to accelerate scientific discovery. 
The material for the report builds on two workshops held by the Fast ML for Science community and covers three main areas: applications for fast ML across a number of scientific domains; techniques for training and implementing performant and resource-efficient ML algorithms; and computing architectures, platforms, and technologies for deploying these algorithms.  
We also present overlapping challenges across the multiple scientific domains where common solutions can be found.  
This community report is intended to give plenty of examples and inspiration for scientific discovery through integrated and accelerated ML solutions.  
This is followed by a high-level overview and organization of technical advances, including an abundance of pointers to source material, which can enable these breakthroughs.  
\tiny
 \keyFont{ \section{Keywords:} fast machine learning} 
\end{abstract}

\clearpage
% \tableofcontents
\begin{spacing}{1.3}
\makeatletter
  \null\hfill\textbf{\Large\contentsname}\hfill\null\par
  \@mkboth{\MakeUppercase\contentsname}{\MakeUppercase\contentsname}%
  \@starttoc{toc}
\makeatother
\end{spacing}

\clearpage

\begin{spacing}{1.3}
\section*{Foreword}
% \textcolor{blue}{tl;dr version}
Machine learning (ML) is making a huge impact on our society and daily lives through advancements in computer vision, natural language processing, and autonomous vehicles, among others.  
ML is also powering scientific advances which can lead to future paradigm shifts in a broad range of domains, including particle physics, plasma physics, astronomy, neuroscience, chemistry, material science, and biomedical engineering.  
Scientific discoveries come from groundbreaking ideas and the capability to validate those ideas by testing nature at new scales---finer and more precise temporal and spatial resolution.  
This is leading to an explosion of data that must be interpreted, and ML is proving a powerful approach. 
The more efficiently we can test our hypotheses, the faster we can achieve discovery.  
To fully unleash the power of ML and accelerate discoveries, it is necessary to embed it into our scientific process, into our instruments and detectors.

It is in this spirit that the Fast Machine Learning for Science community\footnote{\url{fastmachinelearning.org}} has been built.  
Two workshops have also been organized through this growing community and are the source for this report.  
The community brings together an extremely wide-ranging group of domain experts who would rarely interact as a whole. 
One of the underlying benefits of ML is the portability and general applicability of the techniques that can enable experts from seemingly unrelated domains to find a common language. 
Scientists and engineers from particle physicists to networking experts and biomedical engineers are represented and can interact with experts in fundamental ML techniques and compute systems architects.  

This report aims to summarize the progress in the community to understand how our scientific challenges overlap and where there are potential commonalities in data representations, ML approaches, and technology, including hardware and software platforms. 
Therefore, \textbf{the content of the report includes the following: descriptions of a number of different scientific domains including existing work and applications for embedded ML; potential overlaps across scientific domains in data representation or system constraints; and an overview of state-of-the-art techniques for efficient machine learning and compute platforms, both cutting-edge and speculative technologies}.  

Necessarily, such a broad scope of topics \textit{cannot} be comprehensive. 
For the scientific domains, we note that the contributions are \textit{examples} of how ML methods are currently being or planned to be deployed.  
We hope that giving a glimpse into specific applications will inspire readers to find more novel use-cases and potential overlaps. 
The summaries of state-of-the-art techniques we provide relate to rapidly developing fields and, as such, may become out of date relatively quickly.  
The goal is to give non-experts an overview and taxonomy of the different techniques and a starting point for further investigation.
To be succinct, we rely heavily on providing references to studies and other overviews while describing most modern methods.

We hope the reader finds this report both instructive and motivational.  
Feedback and input to this report, and to the larger community, are welcome and appreciated.  

\begin{flushright} 
\textit{
Sincerely, \\The Editors
}
\end{flushright}

\pagebreak
\section{Introduction}
% \textcolor{red}{Section Editors: ...}
In pursuit of scientific advancement across many domains, experiments are becoming exceedingly sophisticated in order to probe physical systems at increasingly smaller spatial resolutions and shorter timescales.  
These order of magnitude advancements have lead to explosions in both data volumes and richness leaving domain scientists to develop novel methods to handle growing data processing needs.  

Simultaneously, machine learning (ML), or the use of algorithms that can learn directly from data, is leading to rapid advancements across many scientific domains~\cite{Carleo:2019ptp}. 
Recent advancements have demonstrated that deep learning (DL) architectures based on structured deep neural networks are versatile and capable of solving a broad range of complex problems. 
The proliferation of large datasets like ImageNet~\cite{imagenet}, computing, and DL software has led to the exploration of many different DL approaches each with their own advantages. 

In this review paper, we will focus on the fusion of ML and experimental design to solve critical scientific problems by accelerating and improving data processing and real-time decision-making. 
We will discuss the myriad of scientific problems that require fast ML, and we will outline unifying themes across these domains that can lead to general solutions. 
Furthermore, we will review the current technology needed to make ML algorithms run fast, and we will present critical technological problems that, if solved, could lead to major scientific advancements. 
An important requirement for such advancements in science is the need for openness.  
It is vital for experts from domains that do not often interact to come together to develop transferable solutions and work together to develop open-source solutions.  

Much of the advancements within ML over the past few years have originated from the use of heterogeneous computing hardware. 
In particular, the use of graphics processing units (GPUs) has enabled the development of large DL algorithms~\cite{gpus1,gpus2,gpus3}.  
The ability to train large artificial intelligence (AI) algorithms on large datasets has enabled algorithms that are capable of performing sophisticated tasks. 
In parallel with these developments, new types of DL algorithms have emerged that aim to reduce the number of operations so as to enable fast and efficient AI algorithms. 

\begin{tcolorbox}
%Can we define here: What is Fast (which is somewhat vague)?  Here we mean accelerated and/or real-time beyond traditional computing paradigms. 
Within this review paper, we refer to the concept of \textit{\textbf{Fast Machine Learning in Science}} as the integration of ML into the experimental data processing infrastructure to enable and accelerate scientific discovery. Fusing powerful ML techniques with experimental design decreases the ``time to science" and can range from embedding real-time feature extraction to be as close as possible to the sensor all the way to large-scale ML acceleration across distributed grid computing datacenters. 
The overarching theme is to lower the barrier to advanced ML techniques and implementations to make large strides in experimental capabilities across many seemingly different scientific applications.  
Efficient solutions require collaboration between domain experts, machine learning researchers, and computer architecture designers.  
\end{tcolorbox}
 
\begin{figure}[tbh!]
    \centering
    \includegraphics[width=0.6\textwidth]{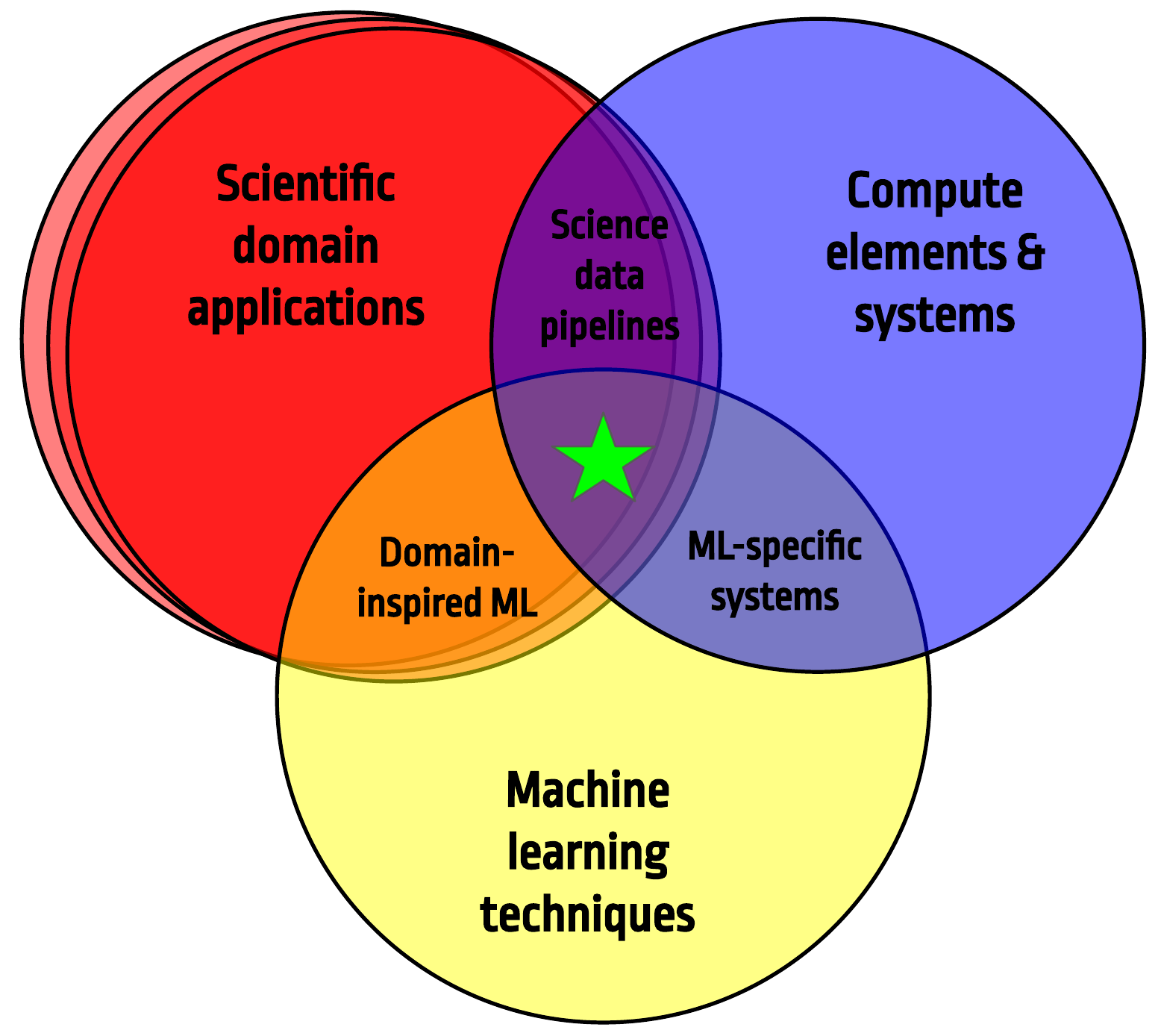}
    \caption{The concept behind this review paper is to find the confluence of domain-specific challenges, machine learning, and experiment and computer system architectures to accelerate science discovery.}
    \label{fig:intro}
\end{figure}

This paper is a review of the second annual Fast Machine Learning conference~\cite{FMLConf2} and will build on the materials presented at this conference. 
It brings together experts from multiple scientific domains ranging from particle physicists to material scientists to health monitoring researchers with machine learning experts and computer systems architects. 
Figure~\ref{fig:intro} illustrates the spirit of the workshop series on which this paper is inspired and the topics covered in subsequent sections.  

As ML tools have become more sophisticated, much of the focus has turned to building very large  algorithms that solve complicated problems, such as language translation and voice recognition. 
However, in the wake of these developments, a broad range of scientific applications have emerged that can benefit  greatly from the rapid developments underway. 
Furthermore, these applications have diversified as people have to come to realize how to adapt their scientific approach so as to take advantage of the benefits originating from the AI revolution. 
%Fast Machine Learning within Science has a broad range of applications. 
This can include the capability of AI to classify events in real time, such as the identification of a collision of particles or a merger of gravitational waves. 
It can also include systems control, such as the response control from feedback mechanisms in plasmas and particle accelerators. 
The latency, bandwidth, and throughput restrictions and the reasons for such restrictions differ within each system. However, in all cases, accelerating ML is a driver in the design goal. 

The design of low latency algorithms differs from other AI implementations in that we must tailor specific processing hardware to the task at hand to increase the overall algorithm performance. 
In particular, certain processor cores have been configured for optimized sparse matrix multiplications. 
Others have been optimized to maximize the total amount of compute. 
Processor design, and the design of algorithms around processors,  often referred to as hardware AI co-design, is the focus of the work in this review. 
For example, in some cases, ultra-low latency inference times are needed to perform scientific measurements. 
One must efficiently design the algorithm to optimally utilize the hardware constraints available while preserving the algorithm performance within desired experimental requirements. This is the essence of hardware AI co-design.

The contents of this review are laid out as follows.  
In the Section~\ref{sec:apps}, we will explore a broad range of scientific problems where Fast ML can act as a disruptive technology to the status quo and lead to a significant change in how we process data.  
Domain experts from seemingly different domains are examined.  
In Section~\ref{sec:overlaps}, we describe  data representations and experimental platform choices are common to many types of experiments.  
We will connect how Fast ML solutions can be generalized to low latency, highly resource-efficient, and domain-specific deep learning inference for many scientific applications. 
Finally in Section~\ref{sec:technolog_sota}, to achieve this requires optimized hardware-ML co-design from the algorithm design to the system architecture.  
We provide an overview of state-of-the-art techniques to train neural networks optimized for both performance and speed, survey various compute architectures to meet the needs of the experimental design and outline software solutions that optimize and enable the hardware deployment.    

The goal of this paper is to bring together scientific opportunities, common solutions,
and state-of-the-art technology into one single narrative. 
We hope this can contribute to accelerating the deployment of potentially transformative ML solutions to a broad range of scientific fields going forward.

\pagebreak
\section{Exemplars of domain applications}
\label{sec:apps}
%\textcolor{red}{Section Editors: ...}
As scientific ecosystems grow rapidly in their speed and scale, new paradigms for data processing and reduction need to be integrated into system-level design.  
In this section, we explore requirements for accelerated and sophisticated data processing.  Implementations of fast machine learning can appear greatly varied across domains and architectures but yet can have similar underlying data representations and needs for integrating machine learning.  
We enumerate here a broad sampling of scientific domains across seemingly unrelated tasks including their existing techniques and future needs.  
This will then lead to the next section where we discuss overlaps and common tasks.

In this section, we first have a detailed description of examples of Fast ML techniques being deployed at experiments for the Large Hadron Collider.  
Much rapid development has occurred for these experiments recently and gives an exemplar for how broad advancements can be made across various aspects of a specific domain.  
Then the following subsections will be briefer but lay out key challenges and areas of existing and potential applications of Fast ML across a number of other scientific domains.  

\subsection{Large Hadron Collider}

The Large Hadron Collider (LHC) at CERN is the world's largest and highest-energy particle accelerator, where collisions between bunches of protons occur every 25 ns. 
To study the products of these collisions, several detectors are located along the ring at interaction points. 
The aim of these detectors is to measure the properties of the Higgs boson~\cite{Aad:2012tfa,Chatrchyan:2012ufa} with high precision and to search for new physics phenomena beyond the standard model of particle physics.
Due to the extremely high frequency of 40 MHz at which proton bunches collide, the high multiplicity of secondary particles, and the large number of sensors, the detectors have to process and store data at enormous rates. For the two multipurpose experiments, CMS~\cite{Collaboration_2008} and ATLAS~\cite{Collaboration_2008}, comprised of tens of millions of readout channels, these rates are of the order of 100 Tb/s. 
Processing and storing this data presents severe challenges that are among the most critical for the execution of the LHC physics program. 

The approach implemented by the detectors for data processing consists of an online processing stage, where the event is selected from a buffer and analyzed in real time, and an offline processing stage, in which data have been written to disk and are more thoroughly analyzed with sophisticated algorithms. 
The online processing system, called the \emph{trigger}, reduces the data rate to a manageable level of 10\unit{Gb/s} to be recorded for offline processing. 
The trigger is typically divided into multiple tiers. 
Due to the limited size of the on-detector buffers, the first tier (Level-1 or L1) utilizes FPGAs and ASICs capable of executing the filtering process with a maximum latency of $\mathcal{O}(1)~\mu$s. 
At the second stage, the high-level trigger (HLT), data are processed on a CPU-based computing farm located at the experimental site with a latency of up to 100 ms. 
Finally, the complete offline event processing is performed on a globally distributed CPU-based computing grid.

Maintaining the capabilities of this system will become even more challenging in the near future.
In 2027, the LHC will be upgraded to the so-called High-Luminosity LHC (HL-LHC) where each collision will produce 5--7 times more particles, ultimately resulting in a total amount of accumulated data that will be one order of magnitude higher than achieved with the present accelerator. 
At the same time, the particle detectors will be made larger, more granular, and capable of processing data at ever-increasing rates. Therefore, the physics that can be extracted from the experiments will be limited by the accuracy of algorithms and computational resources. 

Machine learning technologies offer promising solutions and enhanced capabilities in both of these areas, thanks to their capacity for extracting the most relevant information from high-dimensional data and to their highly parallelizable implementation on suitable hardware.
It is expected that a new generation of algorithms, if deployed at all stages of data-processing systems at the LHC experiments, will play a crucial part in maintaining, and hopefully improving, the physics performance. In the following sections, a few examples of the application of machine learning models to physics tasks at the LHC are reviewed, together with novel methods for their efficient deployment in both the real-time and offline data processing stages.

\subsubsection{Event reconstruction}
\label{sec:lhceventreco}

The reconstruction of proton-proton collision events in the LHC detectors involves challenging pattern recognition tasks, given the large number ($\mathcal{O}(1000)$) of secondary particles produced and the high detector granularity. Specialized detector sub-systems and algorithms are used to reconstruct the different types and properties of particles produced in collisions. 
For example, the trajectories of charged particles are reconstructed from space point measurements in the inner silicon detectors, and the showers arising from particles traversing the calorimeters are reconstructed from clusters of activated sensors. 

Traditional algorithms are highly tuned for physics performance in the current LHC collision environment, but are inherently sequential and scale poorly to the expected HL-LHC conditions. 
It is thus necessary to revisit existing reconstruction algorithms and ensure that both the physics and computational performance will be sufficient. 
Deep learning solutions are currently being explored for pattern recognition tasks, as a significant speedup can be achieved when harnessing heterogeneous computing and parallelizable and efficient ML that exploits AI-dedicated hardware. In particular, modern architectures such as graph neural networks (GNNs) are being explored for the reconstruction of particle trajectories, showers in the calorimeter as well as of the final individual particles in the event. 
Much of the following work has been conducted using the TrackML dataset~\cite{trackml}, which simulates a generalized detector under HL-LHC-like pileup conditions. 
Quantifying the performance of these GNNs in actual experimental data is an ongoing point of study.

For reconstructing showers in calorimeters, GNNs have been found to predict the properties of the original incident particle with high accuracy starting from individual energy deposits. The work in~\cite{Gray:2020mcm} proposes a graph formulation of pooling to dynamically learn the most important relationships between data via an intermediate clustering, and therefore removing the need for a predetermined graph structure. When applied to the CMS electromagnetic calorimeter, with single detector hits as inputs to predict the energy of the original incident particle, a 10\% improvement is found over the traditional boosted decision tree (BDT) based approach.

GNNs have been explored for a similar calorimeter reconstruction task for the high-granularity calorimeters that will replace the current design for HL-LHC. 
The task will become even more challenging as such detectors will feature irregular sensor structure and shape (e.g. hexagonal sensor cells for CMS~\cite{collaboration:2017gbu}), high occupancy, and an unprecedented number of sensors. 
For this application, architectures such as \textsc{EdgeConv}~\cite{DBLP:abs-1801-07829} and \textsc{GravNet/GarNet}~\cite{Qasim:2019otl} have shown promising performance in the determination of the properties of single showers, yielding  excellent energy resolution and high noise rejection~\cite{Ju:2020xty}. 
While these preliminary studies were focused on scenarios with low particle multiplicities, the scalability of the clustering performance to more realistic collision scenarios is still a subject of active development.

%Given the clear and impactful use cases for GNNs in HEP, the next step is efficient hardware acceleration. However, GNNs pose interesting challenges for FPGA and ASIC based acceleration because of their large, variably sized inputs and outputs, and there is research ongoing~\cite{Duarte:2019fta} to discover proper schemes for optimization.
%A proof-of-concept workflow using the NVidia Triton Inference Server and the \textsc{SONIC} client~\cite{Krupa:2020bwg,Wang:2020fjr} in the CMS software stack is already available.

\begin{figure}[ht!p]
    \centering
    \includegraphics[width=0.99\textwidth]{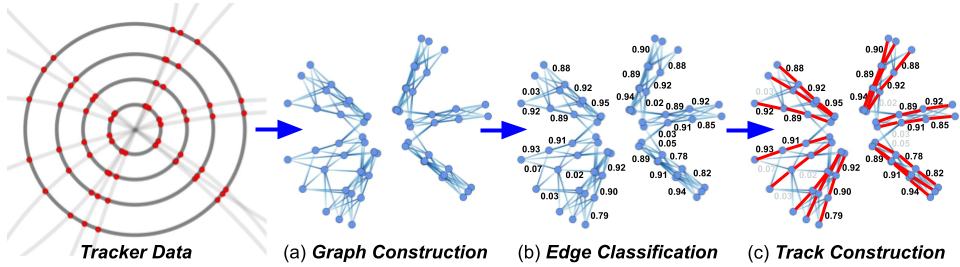}
    \caption{High-level overview of the stages in a GNN-based tracking pipeline. Only a subset of the typical edge weights are shown for illustration purposes.}
    \label{fig:gnn_steps}
\end{figure}

GNNs have also been extensively studied for charged particle tracking (the task of identifying and reconstructing the trajectories of individual particles in the detector)~\cite{exatrk_19,duarte_vlimant, heptrkx,dl_tracking}. 
The first approaches to this problem typically utilized edge-classification GNNs in a three-step process: graphs are constructed by algorithmically constructing edges between tracker hits in a point cloud, the graphs are processed through a GNN to predict edge weights (true edges that are part of true particle trajectories should be highly weighted and false edges should be lowly rated), and finally, the selected edges are grouped together to generate high-weight sub-graphs which form full track candidates, as shown in Figure~\ref{fig:gnn_steps}.

There have been several studies building upon and optimizing this initial framework. 
The ExaTrkX collaboration has demonstrated performance improvements by incorporating a recurrent GNN structure \cite{exatrk_19} and re-embedding graphs prior to training the GNNs~\cite{embedding}. 
Other work has shown that using an Interaction Network architecture~\cite{battaglia2016interaction} can substantially reduce the number of learnable parameters in the GNN \cite{dezoort2021charged}; the authors also provide comprehensive comparisons between different graph construction and track building algorithms. 
Recent work has also explored alternate approaches that combine graph building, GNN inference, and track construction into a single algorithm that is trainable end-to-end; in particular, instance segmentation architectures have generated promising results~\cite{thais2021instance}.

Finally, a novel approach based on GNNs~\cite{Pata:2021oez} has been proposed as an alternative solution to the so-called particle-flow algorithm that is used by LHC experiments to optimally reconstruct each individual particle produced in a collision by combining information from the calorimeters and the tracking detectors~\cite{Sirunyan:2017ulk}. The new GNN algorithm is found to offer comparable performance for charged and neutral hadrons to the existing reconstruction algorithm. 
At the same time, the inference time is found to scale approximately linearly with the particle multiplicity, which is promising for its ability to maintain computing costs within budget for the HL-LHC. 
Further improvements to this original approach are currently under study, including an event-based loss, such as the object condensation approach. 
Second, a complete assessment of the physics performance remains to be evaluated,  including reconstruction of rare particles and other corners of the phase space. 
Finally, it remains to be understood how to optimize and coherently interface this with the ML-based approach proposed for tasks downstream and upstream in the particle-level reconstruction.

\subsubsection{Event simulation}
\label{sec:lhceventsim}

The extraction of results from LHC data relies on a detailed and precise simulation of the physics of proton-proton collisions and of the response of the detector. 
In fact, the collected data are typically compared to a reference model, representing the current knowledge, in order to either confirm or disprove it. Numerical models, based on Monte Carlo (MC) methods, are used to simulate the interaction between elementary particles and matter, while the Geant4 toolkit is employed to simulate the detectors. These simulations are generally very CPU intensive and require roughly half of the experiment’s computing resources, with this fraction expected to increase significantly for the HL-LHC. 

Novel computational methods based on ML are being explored so as to perform precise modeling from particle interactions to detector readouts and response while maintaining feasible computing budgets for HL-LHC. 
In particular, numerous works have focused on the usage of generative adversarial networks or other state-of-the-art generative models to replace computationally intensive fragments of MC simulation, such as modeling of electromagnetic showers~\cite{Paganini:2017dwg,Paganini:2017hrr,deOliveira:2017pjk}, reconstruction of jet images~\cite{Musella:2018rdi} or matrix element calculations~\cite{Bendavid:2017zhk}. 
In addition, the usage of ML generative models on end-to-end analysis-specific fast simulations have also been investigated in the context of Drell-Yan~\cite{Hashemi:2019fkn}, dijet~\cite{DiSipio:2019imz} and W+jets~\cite{Chen:2020uds} production. 
These case-by-case proposals serve as proof-of-principle examples for complementary data augmentation strategy for LHC experiments.

\subsubsection{Heterogeneous computing}

State-of-the-art deep learning models are being explored for the compute-intensive reconstruction of each collision event at the LHC. However, their efficient deployment within the experiments' computing paradigms is still a challenge, despite the potential speed-up when the inference is executed on suitable AI-dedicated hardware. In order to gain from a parallelizable ML-based translation of traditional and mostly sequential algorithms, a heterogeneous computing architecture needs to be implemented in the experiment infrastructure. 
For this reason, comprehensive exploration of the use of CPU+GPU~\cite{Krupa:2020bwg} and CPU+FPGA~\cite{Duarte:2019fta,Rankin:2020usv} heterogeneous architectures was made to achieve the desired acceleration of deep learning inference within the data processing workflow of LHC experiments. These works demonstrated that the acceleration of machine learning inference ``as a service'' represents a heterogeneous computing solution for LHC experiments that  potentially requires minimal modification to the current computing model. 

In this approach, the ML algorithms are transferred to a co-processor on an independent (local or remote) server by reconfiguring the CPU node to communicate with it through asynchronous and non-blocking inference requests.
With the inference task offloaded on demand to the server,  the CPU can be dedicated to performing other necessary tasks within the event.
As one server can serve many CPUs, this approach has the advantage of increasing the hardware cost-effectiveness to achieve the same throughput when comparing it to a direct-connection paradigm.
It also facilitates the integration and scalability of different types of co-processor devices, where the best one is chosen for each task. 

Finally, existing open-source frameworks that have been optimized for fast DL on several different types of hardware can be exploited for a quick adaptation to LHC computing. In particular, one could use the Nvidia Triton Inference Server within a custom framework, so-called Services for Optimized Network Inference on Co-processors (SONIC), to enable remote gRPC calls to either GPUs or FPGAs within the experimental software, which then only has to handle the input and output conversion between event data format and inference server format. 
The integration of this approach within the CMS reconstruction software has been shown to lead to a significant overall reduction in the computing demands both at the HLT and offline.

\subsubsection{Real-time analysis at 40 MHz}
% Thea: Adding only plots/figures related to talks given at the FastML 2020 workshop, but do discuss other applications.
Bringing deep learning algorithms to the Level-1 hardware trigger is an extremely challenging task due to the strict latency requirement and the resource constraints imposed by the system. Depending on which part of the system an algorithm is designed to run on, a latency down to $\mathcal{O}(10)~$ns might be required.  
With $\mathcal{O}(100)~$ processors running large-capacity FPGAs, processing thousands of algorithms in parallel, dedicated FPGA-implementations are needed to make ML algorithms as resource-efficient and fast as possible.
To facilitate the design process and subsequent deployment of highly parallel, highly compressed ML algorithms on FPGAs, dedicated open-source libraries have been developed: \hlsfml and \conifer. 
The former, \hlsfml, provides conversion tools for deep neural networks, while \conifer aids the deployment of Boosted Decision Trees (BDTs) on FPGAs. Both libraries, as well as example LHC applications, will be described in the following.
\begin{figure*}[htb]
    \centering
    \includegraphics[width=0.89\textwidth]{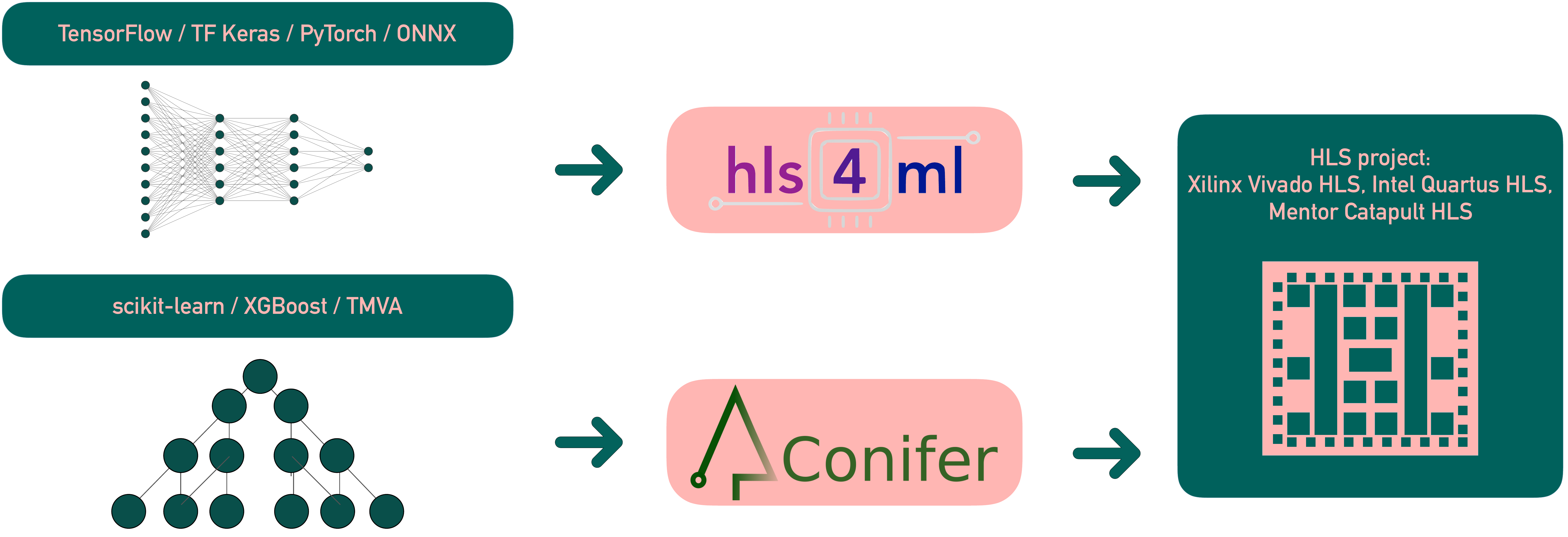}
    \caption{Two dedicated libraries for the conversion of Machine Learning algorithms into FPGA or ASIC firmware: \hlsfml for deep neural network architectures and \conifer for Boosted Decision Tree architectures. Models from a wide range of open-source ML libraries are supported and may be converted using three different high-level synthesis backends.}
    \label{figs2:libraries}
\end{figure*}

The \hlsfml library~\cite{Duarte:2018ite,aarrestad2021fast,DiGuglielmo:2020eqx,Coelho:2020zfu} converts pre-trained ML models into ultra low-latency FPGA or ASIC firmware with little overhead required. 
Integration with the Google QKeras library~\cite{qkeras} allows users to design aggressively quantized deep neural networks and train them quantization-aware~\cite{Coelho:2020zfu} down to 1 or 2 bits for weights and activations~\cite{DiGuglielmo:2020eqx}. 
This step results in highly resource-efficient equivalents of the original model, sacrificing little to no accuracy in the process. The goal of this joint package is to provide a simple two-step approach going from a pre-trained floating point model to FPGA firmware.
The \hlsfml library currently provides support for several commonly used neural network layers like fully connected, convolutional, batch normalization, pooling, as well as several activation functions.
%Recently, support for large convolutional layers has also been added~\cite{aarrestad2021fast}.
These implementations are already sufficient to provide support for the most common architectures envisioned for deployment at L1.
Some first examples of machine learning models designed for the L1 trigger are based on fully connected layers, and they are proposed for tasks such as the reconstruction and calibration of final objects or lower-level inputs like trajectories, vertices, and calorimeter clusters~\cite{CERN-LHCC-2020-004}. 
One example of a convolutional NN (CNN) architecture targeting the L1 trigger is a dedicated algorithm for the identification of long-lived particles~\cite{Alimena_2020}. Here, an attempt is made to efficiently identify showers from displaced particles in a high-granularity forward calorimeter. The algorithm is demonstrated to be highly efficient down to low energies while operating at a low trigger rate.
Traditionally, cut-based selection algorithms have been used for these purposes, in order to meet the limited latency- and resource budget. However, with the advent of tools like \hlsfml and QKeras, ML alternatives are being explored to improve the sensitivity to such physics processes while maintaining latency and resources in the available budget.

%Other examples include variational auto-encoders~(VAE) for anomaly detection, as described in Section 3.7.2.1 of Ref.~\cite{CERN-LHCC-2020-004} and in Ref.~\cite{vaemining}.
%Here, a variational auto-encoder is  designed to correctly identify and trigger on non-Standard Model like events, with the ultimate goal of providing a stream of highly anomalous events. The algorithm is designed to run in the Level-1 Global Trigger, using as input a subset of particles fed into the system, like missing transverse energy and four-vectors of the five highest-$p_{T}$ jets. Jets are then classified as anomalous by the algorithm based on some cut on the degree of anomaly, ultimately decided upon based on the available bandwidth.

More recently, (variational) auto-encoders (VAEs or AEs) are being considered for the detection of ``anomalous'' collision events, i.e. events that are not produced by standard physics processes but that could be due instead to unexpected processes not yet explored at colliders. 
Such algorithms have been proposed for both the incoming LHC run starting in 2022 as well as for the future high-luminosity runs where more granular information will be available. 
The common approach uses global information about the event, including a subset of individual produced particles or final objects such as jets as well as energy sums. 
The algorithm trained on these inputs is then used to classify the event as anomalous if surpassing a threshold on the degree of anomaly (typically the loss function), ultimately decided upon the available bandwidth.
%Currently, work is ongoing to improve and to deploy these anomaly detection algorithms in the CMS hardware triggering system for Run 3. Two possibilities are simultaneously being explored: Deployment in the Level-1 hardware trigger, and deployment in a dedicated {\em 40 MHz scouting system}.
%Scouting at 40 MHz consists of acquiring the L1 trigger data at the full bunch crossing rate of 40 MHz and analyzing certain interesting topologies at the full rate~\cite{CERN-LHCC-2020-004}. A proof-of-principle system, connected directly to the L1 trigger through 25 GB/s optical links and fed to a buffering/analysis system consisting of FPGAs and GPUs, has been successfully tested using inputs from the CMS Global Muon Trigger. This system would provide an excellent platform to test online anomaly detection algorithms before integration in the hardware trigger, and could also be used to run anomaly detection at a higher rate than what would be possible in the Global Trigger. 
Deploying a typical variational autoencoder is impossible in the L1-trigger since the bottleneck layer involves Gaussian random sampling. 
The explored solution is therefore to only deploy the encoder part of the network and do inference directly from the latent dimension. Another possibility is to deploy a simple auto-encoder with the same architecture and do inference computing the difference between output and input. 
However, this would require buffering a copy of the input for the duration it takes the auto-encoder to process the input.
For this reason, the two methods are being considered and compared in terms of accuracy over a range of new physics processes, as well as latency and resources.
%The two methods will be compared in terms of accuracy, latency and resource-consumption. Two possible architectures are currently being investigated: fully-connected and convolutional. Benchmarking is performed on a dataset simulated using the \texttt{DELPHES3} software~\cite{deFavereau:2013fsa} to emulate the CMS detector effects, but the model will eventually be trained on minimum bias data. A wide range of beyond standard model signal samples are used for comparing the performance of different models, and the one which performs the best on average for the different signal hypotheses, will be deployed in the CMS Level-1 trigger in Run 3.

%With the \hlsfml+QKeras workflow and the TensorFlow Pruning API, the goal is to be able to cater for all the different FPGA- and ASIC- based experiments at LHC (and other areas which require ultra low latency inference).
%However, some problems require more complicated architectures and NN layers. Specific layers can therefore easily be added to the \hlsfml library.
Finally, another interesting aspect of the \hlsfml tool is the capability for users to easily add custom layers that might serve a specific task not captured by the most common layers supported in the library. One example of this is compressed distance-weighted graph networks~\cite{garnet}, where a graph network block called a \emph{GarNet layer} takes as input a set of V vertices, each of which has $F_{in}$ features, and returns the same set of vertices with $F_{out}$ features. 
To keep the dimensionality of the problem at a manageable level, the input features of each vertex are encoded and aggregated at $S$ aggregators. Message-passing is only performed between vertices and a limited set of aggregators, and not between all vertices, significantly reducing the network size. 
In Ref.~\cite{garnet}, an example task of pion and electron identification and energy regression in a 3D calorimeter is studied. 
A total inference latency of $\mathcal{O}(100)~$ns is reported, satisfying the L1 requirement of $\mathcal{O}(1)~\mu$s latency. 
The critical resource is digital signal processing (DSP) units, where 29\% of the DSPs are in use by the algorithm. 
This can be further reduced by taking advantage of quantization-aware training with QKeras. 
Another example of a GNN architecture implemented on FPGA hardware using \hlsfml is presented in Ref.~\cite{heintz2020accelerated}. 
This work shows that a compressed GNN can be deployed on FPGA hardware within the latency and resources required by L1 trigger system for the challenging task of reconstructing the trajectory of charged particles.
%This GNN is designed to identify the trajectory of charged particles, an extremely challenging task where current algorithms scale worse than quadratically in the number of hits. Even for this complicated task, a latency of ${\cal O}(1)~\mu$s can be achieved for a compressed GNN architecture. The algorithm is described in detail in Section~\ref{sec:lhceventreco} above.
%Thes are important results as demonstrating that more advanced and custom neural network architectures can be successfully deployed in the Level-1 trigger system. 

In many cases, the task to be performed is simple enough that a boosted decision tree (BDT) architecture suffices to solve the problem. 
As of today, BDTs are still the most commonly used ML algorithm for LHC experiments. 
To simplify the deployment of these, the library {\tt Conifer}~\cite{Summers:2020xiy} has been developed. In {\tt Conifer}, the BDT implementation targets extreme low latency inference by executing all trees, and all decisions within each tree, in parallel. 
BDTs and random forests can be converted from scikit-learn~\cite{scikit-learn}, XGBoost~\cite{XGBoost}, and TMVA~\cite{TMVA}, with support for more BDT training libraries planned.

There are several ongoing projects at LHC which plan to deploy BDTs in the Level-1 trigger using {\tt Conifer}. One example is a BDT designed to provide an estimate of the {\em track quality}, by learning to identify tracks that are reconstructed in error, and do not originate from a real particle~\cite{bdt_tq}. 
%Both a DNN and BDT architecture is explored and the results are shown in Figure~\ref{figs2:bdt_tq}.
%\begin{figure}[hbt]
%    \centering
%    \includegraphics[width=0.70\textwidth]{figures/bdt_trackquality.pdf}
%    \caption{Accuracy, latency and resource usage for a track quality algorithm using a BDT or a DNN %architecture~\cite{bdt_tq}.}
%    \label{figs2:bdt_tq}
%\end{figure}
While the accuracy and resource usage are similar between a BDT and a DNN, the latency is significantly reduced for a BDT architecture. 
The algorithm is planned to be implemented in the CMS Experiment for the data-taking period beginning in 2022.
%Another example is a BDT for electron and photon identification for the CMS High Granularity Endcap Calorimeter, described in Section 3.2.3 of Ref.~\cite{CERN-LHCC-2020-004}. Signal candidates are distinguished from clusters originating from pileup using five longitudinal and four lateral shower shape variables.

Rather than relying on open source libraries such as \hlsfml or \conifer, which are based on high-level synthesis tools from FPGA vendors, other approaches are being considered based directly on hardware description languages, such as VHDL~\cite{Nottbeck:2019rqu,Fritzsche2020}. 
One example is the application of ML for the real-time signal processing of the ATLAS Liquid Argon calorimeter~\cite{atlas1996atlas}. 
It has been shown that with upgraded capabilities for the HL-LHC collision environment the conventional signal processing, which applies an optimal filtering algorithm~\cite{Cleland:2002rya}, will lose its performance due to the increase of overlapping signals. 
More sophisticated DL methods have been found to be more suitable to cope with these challenges being able to maintain high signal detection efficiency and energy reconstruction. 
More specifically, studies based on simulation~\cite{madysa-chep} of dilated convolutional neural networks showed promising results. 
An implementation of this architecture for FPGA is designed using VHDL~\cite{Fritzsche2020} to meet the strict requirements on latency and resources required by the L1 trigger system. 
The firmware runs with a multiple of the bunch crossing frequency to reuse hardware resources by implementing time-division multiplexing while using pipeline stages, the maximum frequency can be increased. 
Furthermore, DSPs are chained up to perform the MAC operation in between two layers efficiently. In this way, a core frequency of more than 480 MHz could be reached, corresponding to twelve times the bunch crossing frequency.

\subsubsection{Bringing ML to detector front-end}

While LHC detectors grow in complexity to meet the challenging conditions of higher-luminosity environments, growing data rates prohibit transmission of full event images off-detector for analysis by conventional FPGA-based trigger systems.
As a consequence, event data must be compressed on-detector in low-power, radiation-hard ASICs while sacrificing minimal physics information.

Traditionally this has been accomplished by simple algorithms, such as grouping nearby sensors together so that only these summed ``super-cells'' are transmitted, sacrificing the fine segmentation of the detector.
Recently, an autoencoder-based approach has been proposed, relying instead on a set of machine-learned radiation patterns to more efficiently encode the complete calorimeter image via a CNN.
Targeting the CMS high-granularity endcap calorimeter (HGCal)~\cite{collaboration:2017gbu} at the HL-LHC, the algorithm aims to achieve higher-fidelity electromagnetic and hadronic showers, critical for accurate particle identification.

The on-detector environment (the ECON-T concentrator ASIC~\cite{collaboration:2017gbu}) demands a highly-efficient CNN implementation; a compact design should be thoroughly optimized for limited-precision calculations via quantization-aware training tools~\cite{qkeraspaper}.
Further, to automate the design, optimization, and validation of the complex NN circuit, HLS-based tool flows~\cite{Duarte:2018ite} may be adapted to target the ASIC form factor.
Finally, as the front-end ASIC cannot be completely reprogrammed in the manner of an FPGA, a mature NN design is required from the time of initial fabrication.
However, adaptability to changing run conditions and experimental priorities over the lifetime of the experiment motivate the implementation of all NN weights as configurable registers accessible via the chip's slow-control interface.

%%%%%%%%%%%%%%%%%%%%%%%%%%%%%%%%%%%%%%%%%%%%%%%%%%%%%%%%%%%%%%%%%
%%%%%%%%%%%%%%%%%%%%%%%%%%%%%%%%%%%%%%%%%%%%%%%%%%%%%%%%%%%%%%%%%

\subsection{High intensity accelerator experiments}
\subsubsection{ML-based Trigger System at the Belle II Experiment}
\subparagraph*{Context:} The Belle II experiment in Japan is engaged in the search for physics phenomena that cannot be explained by the Standard Model. Electrons and positrons are accelerated at the SuperKEKB particle accelerator to collide at the interaction point located inside of the Belle II detector. 
The resulting decay products are continually measured by the detector’s heterogeneous sensor composition. 
The resulting data is then stored offline for detailed analysis. 

\subparagraph*{Challenges:} Due to the increasing luminosity (target luminosity is $8\times10^{35}\mathrm{cm}^{-2}\mathrm{s}^{-1}$) most of the recorded data is from unwanted but unavoidable background reactions, rather than electron-positron annihilation at the interaction point.  Not only is storing all the data inefficient due to the high background rates, but it is also not feasible to build an infrastructure that stores all the generated data. A multilevel trigger system is used as a solution to decide online which recorded events are to be stored. 

\subparagraph*{Existing and Planned Work:} The Neural Network z-Vertex Trigger (NNT) described used at Belle II is a deadtime-free level 1 (L1) trigger that identifies particles by estimating their origin along the beampipe. 
For the whole L1 trigger process, from data readout to the decision, a real-time 5\unit{$\mu$s} time budget is given to avoid dead-time \cite{Lai_2020}. 
Due to the time cost of data pre-processing and transmission, the NNT needs to provide a decision within 300\unit{ns} processing time.

The task of the NNT is to estimate the origin of a particle track so that it can be decided whether it originates from the interaction point or not. 
For this purpose, a multilayer perceptron (MLP) implemented on a Xilinx Virtex 6 XC6VHX380T FPGA is used. 
The MLP consists of three layers with 27 input neurons, 81 hidden layer neurons and two output neurons. 
Data from the Belle II's central drift chamber (CDC) is used for this task, since it is dedicated to the detection of particle tracks. 
Before being processed by the network, the raw detector data is first combined into a 2D track based on so-called track segments, which are groupings of adjacent active sense wires. 
%For the processing, five different networks are used, which are trained to compensate missing sensor data. 
%The sense wires of the CDC are grouped in nine so-called super layer (SL) which are divided into alternating axial and stereo layers. 
%Since not all SLs are always providing data to be used for the estimation, the NNT selects an appropriate network online to compensate this. 
The output of the NNT delivers the origin of the track in $z$, along the beampipe, as well as the polar angle $\theta$. 
With the help of the z-vertex, the downstream global decision logic (GDL) can decide whether a track is from the interaction point or not. 
In addition, the particle momentum can be detected using the polar angle $\theta$~\cite{baehr2019low}. 

The networks used in the NNT are trained offline. 
The first networks were trained with plain simulated data because no experimental data were available. 
For more recent networks, reconstructed tracks from the experimental data are used. 
For the training the iRPROP algorithm is used which is an extension of the RPROP backpropagation algorithm. Current results show a good correlation between the NNT tracks and reconstructed tracks. 
Since the event rate and the background noise are currently still tolerable, the z-cut, i.e., the allowed estimated origin of a track origin in order to be kept, is chosen at $\pm 40$\,cm. 
With increasing luminosity and the associated increasing background, this z-cut can be tightened. 
Since the new Virtex Ultrascale based universal trigger board (UT4) is available for the NNT this year, an extension of the data preprocessing is planned. 
This will be done by a 3D Hough transformation for further efficiency increases. 
It has already been shown in simulation that a more accurate resolution and larger solid angle coverage can be achieved~\cite{Skambraks_2020}.

%%%%%%%%%%%%%%%%%%%%%%%%%%%%%%%%%%%%%%%%%%%%%%%%%%%%%%%%%%%%%%%%%
%%%%%%%%%%%%%%%%%%%%%%%%%%%%%%%%%%%%%%%%%%%%%%%%%%%%%%%%%%%%%%%%%

\subsubsection{Mu2e}
\subparagraph*{Context:} The Mu2e experiment at Fermilab will search for the charged lepton flavor violating process of neutrino-less $\mu \to e$ coherent conversion in the field of an aluminum nucleus. 
About $7\cdot 10^{17}$ muons, provided by a dedicated muon beamline in construction at Fermilab, will be stopped in 3 years in the aluminum target. 
The corresponding single event sensitivity will be $2.5\cdot 10^{-17}$. 
To detect the signal $e^-$ ($p=105$\unit{MeV}), Mu2e uses a detector system made of a straw-tube tracker and a crystal electromagnetic calorimeter~\cite{MU2E}. 

\subparagraph*{Challenges:} The trigger system is based on detector Read Out Controllers (ROCs) which stream out continuously the data, zero-suppressed, to the Data Transfer Controller units (DTCs). The proton pulses are delivered at a rate of about 600 kHz and a duty cycle of about 30\% (0.4 s out of 1.4 s of the booster-ring delivery period). Each proton pulse is considered a single event, with the data from each event then grouped at a single server using a 10~Gbps Ethernet switch. Then, the online reconstruction of the events starts and makes a trigger decision.  The trigger system needs to satisfy the following requirements: (1)  provide efficiency better than 90\% for the signals; (2)  keep the trigger rate below a few kHz -- equivalent to 7 Pb/year; (3) achieve a processing time $<5$~ms/event. Our main physics triggers use the information of the reconstructed tracks to make the final decision.  

\subparagraph*{Existing and Planned Work:} The current strategy is to perform the helix pattern recognition and the track reconstruction with the CPUs of the DAQ servers, but so far this design showed limitations in matching the required timing performance~\cite{pezzullo_gianantonio_2020_4088480}. 
Another idea that the collaboration started exploring is to perform the early stage of the track reconstruction on the ROC and DTC FPGA using the High Level Synthesis tool (HLS) and the \texttt{hls4ml} package. 
The Mu2e helix pattern-recognition algorithms~\cite{pezzullo_gianantonio_2020_4088480} are a natural fit for these tools for several reasons: they use neural-networks to clean up the recorded straw-hits from hits by low-momentum electrons ($p<10$\unit{MeV}) and they perform large combinatorics calculations when reconstructing the helicoidal electron trajectory. 
This R\&D is particularly important for the design of the trigger system of the planned upgrade of Mu2e~\cite{abusalma2018expression}, where we expect to: (i) increase the beam intensity by at least a factor of 10, (ii) increase the duty cycle to at least 90\%, and (iii) increase the number of detector's channels to cope with the increased occupancy.

%%%%%%%%%%%%%%%%%%%%%%%%%%%%%%%%%%%%%%%%%%%%%%%%%%%%%%%%%%%%%%%%%
%%%%%%%%%%%%%%%%%%%%%%%%%%%%%%%%%%%%%%%%%%%%%%%%%%%%%%%%%%%%%%%%%

\subsection{Materials Discovery}
\subsubsection{Materials Synthesis}

\subparagraph*{Context:} Advances in electronics, transportation, healthcare, and buildings require the synthesis of materials with controlled synthesis-structure-property relationships. 
To achieve application-specific performance metrics, it is common to design and engineer materials with highly ordered structures. 
This directive has led to a boom in non-equilibrium materials synthesis techniques. 
Most exciting are additive synthesis and manufacturing techniques, for example, 3d-printing\cite{Wang2020-tv,Parekh2016-vj,Visser2015-hy,Ligon2017-dg,Zarek2016-dw} and thin film deposition\cite{Chrisey1994-gw,Richter1990-ml,Yoshino2000-oo,Kelly2000-xk,Marvel2013-cd,George2010-pb,Park2001-so}, where complex nanoscale architectures of materials can be fabricated. 
To glean insight into synthesis dynamics, there has been a trend to include in situ diagnostics to observe synthesis dynamics\cite{Ojeda-G-P2017-la,Egelhoff1989-pr,Thomas1999-ij,Langereis2007-dj}.
There is less emphasis  on automating the downstream analysis to turn data into actionable information that can detect anomalies in synthesis, guide experimentation, or enable closed-loop control. Part of the challenge with automating analysis pipelines for in situ diagnostics is the highly variable nature and multimodality of the measurements and the sensors. 
A system might measure many time-resolved state variables (time-series) at various locations (e.g., temperature, pressure, energy, flow rate, etc.)\cite{Hansen1999-an}. 
Additionally, it is common to measure time-resolved spectroscopic signals (spectrograms) that provide, for instance, information about the dynamics of the chemistry and energetic distributions of the materials being synthesized\cite{Cooks2018-jm,Termopoli2019-gb,Dauchot1995-cu,Aubriet2002-ln}. 
Furthermore, there are a growing number of techniques that leverage high-speed temporally-resolved imaging to observe synthesis dynamics\cite{Trigub2017-xw,Ojeda-G-P2018-cv}.

\subparagraph*{Challenges:} Experimental synthesis tools and in situ diagnostic instrumentation are generally semi-custom instruments provided by commercial vendors. 
Many of these vendors rely on proprietary software to differentiate their products from their competition. In turn, the closed-nature of these tools and even data schemas makes it hard to utilize these tools fully. The varied nature and suppliers for sensors compounds this challenge. Integration and synchronization of multiple sensing modalities require a custom software solution. 
However, there is a catch-22 because the software does not yet exist. 
Researchers cannot be ensured that the development of analysis pipelines will contribute to their ultimate goal to discover new materials or synthesize materials with increased fecundity. Furthermore, there are significant workforce challenges as most curriculums emphasize Edisonian rather than computational methods in the design of synthesis. There is an urgent need for multilingual trainees fluent in typically disparate fields.

\subparagraph*{Existing and Planned Work:} Recently, the materials science community has started to embrace machine learning to accelerate scientific discovery\cite{Butler2018-qo,Schmidt2019-dz,Ramprasad2017-wp}. 
However, there have been growing pains. The ability to create highly overparameterized models to solve problems with limited data provides a false sense of efficacy without the generalization required for science. 
Machine learning model architectures designed for natural time-series and images are ill-posed for physical processes governed by equations. 
In this regard, there is a growing body of work to embed physics in machine learning models, which serve as the ultimate regularizers. 
For instance, rotational~\cite{Oxley2020-hg,Kalinin2020-xl} and Euclidean equivariance~\cite{Smidt_undated-oh,Smidt2020-sh} has been built into the model architectures, and methods to learn sparse representations of underlying governing equations have been developed\cite{Kaheman2020-zt,De_Silva2020-ef,Champion2019-kh}.

Another challenge is that real systems have system-specific discrepancies that need to be compensated\cite{Kaheman2019-yu}. For example, a precursor from a different batch might have a slightly different viscosity that needs to be considered.  There is an urgent need to develop these foundational methods for materials synthesis. Complementing these foundational studies, there has been a growing body of literature emphasizing post-mortem machine-learning-based analysis of in situ spectroscopies\cite{Provence2020-ro,Trejo2019-ph}. 
As these concepts become more mature, there will be an increasing emphasis on codesign of synthesis systems, machine learning methods, and hardware for on-the-fly analysis and control. 
This effort towards self-driving laboratories is already underway in wet-chemical synthesis where there are minimal dynamics, and thus, latencies are not a factor\cite{MacLeod2020-mv,Langner2020-ds}. 
Future efforts will undoubtedly focus on controlling dynamic synthesis processes where millisecond-to-nanosecond latencies are required.

\subsubsection{Scanning Probe Microscopy}

\subparagraph*{Context:} Touch is the first sense humans develop. Since the atomic force microscope’s (AFM) invention in 1985\cite{Binnig1986-ig}, humans have been able to “feel” surfaces with atomic level resolution with pN sensitivity. 
AFMs rely on bringing an atomically sharp tip mounted on a cantilever into contact with a surface. By scanning this tip nanometer-to-atomically resolved images can be constructed by measuring the angular deflection of a laser bounced off the cantilever. This detection mechanism provides high-precision sub-angstrom measures of displacement. 

By adding functionality to the probe (e.g., electrical conductivity\cite{Benstetter2009-oe}, resistive heaters\cite{King2005-sc}, single-molecule probes\cite{Oberhauser2002-cs}, and N-V centers\cite{Ariyaratne2018-hg}), scanning probe microscopy (SPM) can measure nanoscale functional properties, including electrical conductivity\cite{Seidel2010-uv,Gomez-Navarro2005-pu}, piezoresponse\cite{Jesse2011-tv}, electrochemical response\cite{Jesse2012-gh}, magnetic force\cite{Kazakova2019-dj}, magnetometry\cite{Casola2018-ms}, and much more. 
These techniques have been expanded to include dynamics measurements during a tip-induced perturbation that drives a structural transformation. These methods have led to a boom in new AFM techniques, including fast-force microscopy\cite{Benaglia2018-js}, current-voltage spectroscopies\cite{Holstad2020-kq}, band-excitation-based spectroscopies\cite{Jesse2018-jw}, and full-acquisition mode spectroscopies\cite{Somnath2015-qk}. 
What has emerged is a data deluge where these techniques are either underutilized or under-analyzed.

\subparagraph*{Challenges:} The key practical challenge is that it takes on days-to-weeks to analyze data from a single measurement properly. 
As a result, experimentalists have little information on how to design their experiments. 
There is even minimal feedback on whether the experiments have artifacts (e.g., tip damage) that would render the results unusable. The number of costly failed experiments is a strong deterrent to conducting advanced scanning probe spectroscopies and developing even more sophisticated imaging techniques. 
There is a significant challenge in both the acceleration and automation of analysis pipelines. 

\subparagraph*{Existing and Planned Work:} In materials science, scanning probe microscopy has quickly adopted machine learning. 
Techniques for linear and nonlinear spectral unmixing provide rapid visualization and extraction of information from these datasets to discover and unravel physical mechanisms~\cite{Ziatdinov2020-nt,Collins2020-na,Kalinin2021-gp,Collins2020-ml}. 
The ease of applying these techniques has led to justified concerns about the overinterpretation of results and overextension of linear models~\cite{Griffin2020-mc} to highly nonlinear systems. 
More recently, long-short term memory autoencoders were controlled to have non-negative and sparse latent spaces for spectral unmixing. 
By traversing the learned latent space, it has been possible to draw complex structure-property relationships~\cite{Agar2019-eq,Holstad2020-kq}. 
There are significant opportunities to accelerate the computational pipeline such that information can be extracted on practically relevant time scales by the experimentalist on the microscope. 

Due to the high velocity of data, up to GB/s, with sample rates of 100,000 spectra, extracting even cursory information will require the confluence of data-driven models, physics-informed machine learning, and AI hardware. 
As a tangible example, in band-excitation piezoresponse force microscopy, the frequency-dependent cantilever response is measured at rates up to 2,000 spectra-per-second. 
Extracting the parameters from these measurements requires fitting the response to an empirical model. Using least-squares fitting throughput is limited to $\sim50$-fits/core-minute, but neural networks provide an opportunity to accelerate analysis and better handle noisy data~\cite{Borodinov2019-pn}. 
There is an opportunity to deploy neural networks on GPU or FPGA hardware accelerators to approximate and accelerate this pipeline by orders of magnitude.

%\subsubsection{Electron Microscopy}
%\subsubsection{X-ray Spectroscopy}
\subsection{Fermilab Accelerator Controls}

\subparagraph*{Context:} 
            The Fermi National Accelerator Laboratory (Fermilab) is dedicated to investigating matter, energy, space, and time \cite{fermilab_about}. 
            For over 50 years, Fermilab's primary tool for probing the most elementary nature of matter has been its vast accelerator complex. Spanning a number of miles of tunnels, the accelerator complex is actually multiple accelerators and beam transport lines each representing different accelerator techniques and eras of accelerator technologies. 
            In its long history, Fermilab's accelerator complex has had to adapt to the mission, asking more of the accelerators than they were designed for and often for purposes they were never intended. 
            This often resulted in layering new controls on top of existing antiquated hardware. 
            Until recently, accelerator controls focused mainly on providing tools and data to the machine operators and experts for tuning and optimization. 
            Having recognized the future inadequacies of the current control system and the promise of new technologies such as ML, the Fermilab accelerator control system will be largely overhauled in the coming years as part of the Accelerator Controls Operations Research Network (ACORN) project~\cite{acorn_paper}. 
            
\subparagraph*{Challenges:} 
            The accelerator complex brings unique challenges for machine learning. Particle accelerators are immensely complicated machines, each consisting of many thousands of variable components and even larger data sources. 
            Their large size and differing types, resolution, and frequency of data mean collecting and synchronizing data is difficult. 
            Also, as one might imagine, control and regulation of beams  that travel at near light speeds is always a challenge. 
            Maintaining and upgrading the accelerator complex controls is costly.
            For this reason, much of the accelerator complex is a mixture of obsolete, new and cutting edge hardware. 
            % Particle accelerators are immensely complicated machines, each with thousands of variable components. 
            % Keeping older accelerators running with higher demand than was ever designed for. 
            % Re-purposing accelerators for tasks they were never designed for. Overlaying advanced control algorithms atop of now obsolete control infrastructure. 
            % New accelerators having very tight design specs, machine protection critical. \cite{7454846}
            % Readings for realtime regulation coming from large distances and latencies
            % Readings coming from very different hardware types with varying resolutions and frequency
            
\subparagraph*{Existing and Planned Work:} 
            Traditional accelerator controls have focused on grouping like elements so that particular aspects of the beam can be tuned independently. However, many elements are not always completely separable. 
            Magnets, for example, often have higher-order fields that affect the beam in different ways than is the primary intent. 
            Machine learning has made it finally possible to combine previously believed to be unrelated readings and beam control elements into new novel control and regulation schemes. 
            
            One such novel regulation project is underway for the Booster Gradient Magnet Power Supply (GMPS). GMPS controls the primary trajectory of the beam in the Booster~\cite{operations_booster_rookie_book}. 
            The project hopes to increase the regulation precision of GMPS ten-fold. 
            When complete, GMPS would be the first FPGA online ML-model-based regulation system in the Fermilab accelerator complex~\cite{john2021realtime}. 
            The promise of ML for accelerator controls is so apparent to the Department of Energy that a call for accelerator controls using ML was made to the national labs \cite{doe_foa_lab_20-2261}. Of the two proposals submitted by Fermilab and approved by the DOE is the Real-time Edge AI for Distributed Systems (READS) project. READS is actually two projects. 
            The first READS project will create a complimentary ML regulation system for slow extraction from the Delivery Ring to the future Mu2e experiment~\cite{bartoszek2015mu2e}. 
            The second READS project will tackle a long-standing problem with de-blending beam losses in the Main Injector (MI) enclosure. The MI enclosure houses two accelerators, the MI and the Recycler. 
            During normal operation, high intensity beams exist in both machines. 
            One to use ML to help regulate slow spill in the Delivery ring to Mu2e, and another to develop a real-time online model to de-blend losses coming from the Recycler and Main Injector accelerators which share an enclosure. Both READS projects will make use of FPGA online ML models for inference and will collect data at low latencies from distributed systems around the accelerator complex~\cite{seiya2021accelerator}.
            % Booster GMPS AI \cite{john2021realtime}
            % Mu2e spill regulation 
            % MI/RR loss deblending 
            % PIP2 SRF quench protection

\subsection{Neutrino and direct dark matter experiments}
\subsubsection{Accelerator Neutrino Experiments}\label{sec:nuaccel}

\subparagraph*{Context:} Accelerator neutrino experiments detect neutrinos with energies ranging from a few tens of MeV up to about 20\unit{GeV}. 
The detectors can be anywhere from tens of meters away from the neutrino production source, to as far as away as 1500\unit{km}. 
For experiments with longer baselines it is common for experiments to consist of both a near ($\sim$1\unit{km} baseline) and a more distant far detector (100's\unit{km} baseline). 
Accelerator neutrino experiments focused on long-baseline oscillations use highly pure muon neutrino beams, produced by pion decays in flight. 
By using a system of magnetic horns it is possible to produce either a neutrino, or antineutrino beam. This ability is particularly useful for CP-violation measurements.   
Other experiments use pions decaying at rest, which produce both muon and electron flavors.
    
The primary research goal of many accelerator neutrino experiments is to perform neutrino oscillation measurements; the process by which neutrinos created in one flavor state are observed interacting as different flavor states after traveling a given distance. 
Often this takes the form of measuring electron neutrino appearance and muon neutrino disappearance. The rate of oscillation is energy-dependent, and so highly accurate energy estimation is essential. 
Another key research goal for accelerator neutrinos is to measure neutrino cross-sections, which in addition to accurate energy estimation requires the identification of the particles produced by the neutrino interaction. 
    
\subparagraph*{Challenges:} Accelerator neutrino experiments employ a variety of detector technologies. 
These range from scintillator detectors such as NOvA (liquid), MINOS (solid), and MINERvA (solid), to water 
Cherenkov detectors such as T2K, and finally liquid argon time projection chambers such as MicroBooNE, ICARUS, and DUNE. 
Pion decay-at-rest experiments (COHERENT, JSNS$^2$) use yet different technologies (liquid and solid scintillators, as well as solid-state detectors). The individual challenges and solutions are unique to each experiment, though common themes do emerge.
    
    Neutrino interactions are fairly uncommon due to their low cross-section. 
    Some experiments can see as few as one neutrino interaction per day. This, combined with many detectors being close to the surface, means that analyses have to be highly efficient whilst achieving excellent background rejection. 
    This is true both in online data taking and offline data analysis.
    
    As experiments typically have very good temporal and/or spatial resolution it is often fairly trivial to isolate entire neutrino interactions. 
    This means that it is then possible to use image recognition tools such as CNNs to perform classification tasks. 
    As a result, many experiments initially utilized variants of GoogLeNet, though many are now transitioning to use GNNs and networks better able to identify sparse images. 

\subparagraph*{Existing and Planned Work:} As discussed in Section~\ref{sec:nu_astro}, DUNE will use machine learning in its triggering framework to handle its immense data rates and to identify candidate interactions, for both traditional neutrino oscillation measurements and for candidate solar and supernova events. 
Accelerator neutrino experiments have successfully implemented machine learning techniques for a number of years, the first such example being in 2017~\cite{Adamson_2017}, where the network increased the effective exposure of the analysis by 30\%. Networks aimed at performing event classification are common across many experiments, with DUNE having recently published a network capable of exceeding its design sensitivity on simulated data and which includes outputs that count the numbers of final state particles from the interaction~\cite{Abi_2020}.
    
Experiments are becoming increasingly cognizant of the dangers of networks learning features of the training data beyond what is intended. For this reason, it is essential to carefully construct training datasets such that this risk is reduced. 
However, it is not possible to correct or quantify bias which is not yet known; therefore the MINERvA experiment has explored the use of a domain adversarial neural network~\cite{Perdue_2018} to reduce unknown biases from differences in simulated and real data. 
The network features a gradient reversal layer in the domain network (trained on data), thus  discouraging the classification network (trained on simulation) to learn from any features that behave differently between the two domains. 
A more robust exploration of the machine learning applied to accelerator neutrino experiments can be found here in Ref.~\cite{Psihas_2020}.

\subsubsection{Neutrino Astrophysics} \label{sec:nu_astro}

\subparagraph*{Context:} Neutrino astrophysics spans a wide range of energies, with neutrinos emitted from both steady-state and transient sources with energies from less than MeV to EeV scale.  
Observations of astrophysical neutrinos are valuable both for the understanding of neutrino sources and for probing fundamental physics.  
Neutrino detectors designed for observing these tend to be huge scale (kilotons to megatons).  
Existing detectors involve a diverse range of materials and technologies for particle detection; they include Cherenkov radiation detectors in water and ice, liquid scintillator detectors and, liquid argon time projection chambers.
  
Astrophysical neutrinos are one kind of messenger contributing to the thriving field of \textit{multimessenger astronomy}, in which signals from neutrinos, charged particles, gravitational waves, and photons spanning the electromagnetic spectrum are observed in coincidence.  
This field has had some recent spectacular successes~\cite{Abbott_2017,AaAc2018,GrFo2020}.  
For multimessenger transient astronomy, time is of the essence for sharing data and locating sources.  
\textit{Directional information} from the neutrinos is critically valuable, to allow prompt location of the source by other messengers.

Potential interesting transient astrophysical sources include sources of ultra-high energy neutrinos, as well as nearby stellar core collapses.   
Neutrinos in the multi-GeV and higher range are emitted from distant cosmic sources, including kilonovae and blazars, and cubic-km-scale water-based Cherenkov detectors such as IceCube at the South Pole can produce fast alerts from  single neutrino observations.

Core-collapse supernovae are another promising use case for fast machine learning.  
These are copious sources of few tens of MeV-scale neutrinos, which are emitted in a burst lasting a few tens of seconds~\cite{Scholberg:2012id,Mirizzi:2015eza}. 
The neutrinos are prompt after core collapse (as will be gravitational waves) but observable electromagnetic radiation will not emerge for anywhere from tens to 10$^6$\unit{s}, depending on the nature of the progenitor and its envelope~\cite{Kistler:2012as}.  
Low-latency information is therefore immensely valuable.  
Core-collapse supernovae are rare events within the distance range observable by current and near-future neutrino detectors.  
They occur only every several decades, which makes prompt and robust detection especially important.  
The SuperNova Early Warning System~\cite{Antonioli:2004zb,Kharusi:2020ovw} aims to provide a prompt alert from a coincidence of burst detections.  
However, pointing information from neutrinos is relatively difficult to extract promptly.  Detectors with the capability for prompt pointing thanks to the anisotropy of neutrino interactions (i.e. the interaction products that remember where the neutrino came from) offer the best prospects, but these need to be able to select neutrino events from background and reconstruct their directions with very low latency.

% Skip if this is too long
Presupernova neutrinos are another interesting possibility. In the final stages of stellar burning, one expects a characteristic uptick in neutrino luminosity and average energy, producing observable events in detectors for nearby progenitors.  
This could give a warning of hours or perhaps days before core collapse for the nearest progenitors. For this case, fast selection of neutrino-like events and reconstruction of their directional information for background reduction is needed.

\subparagraph*{Challenges:}

  The challenges, in general, are fast selection and reconstruction of neutrino event (interaction) information.  
  The specifics of the problem depend on the particular detector technology, but in general, the charged particle products of a neutrino interaction will have a distinctive topology or other signature and must be selected from a background of cosmic rays, radiologicals, or detector noise.
  Taking as an example a liquid argon time projection chamber like the Deep Underground Neutrino Experiment (DUNE), neutrino-induced charged particles produce charge and light signals in liquid argon.   
  Supernova neutrino interactions appear as small (tens of cm spatial scale) stubs and blips~\cite{Abi:2020lpk, Abi:2020evt}. 
  The recorded neutrino event information from the burst can be used to reconstruct the supernova direction to $\sim$5--10$^\circ$ for core collapse at 10\unit{kpc} distance~\cite{ajpointingtalk,Abi:2020evt}. 
  The neutrino events need to be selected from a background of radioactivity and cosmogenics, as well as detector noise, requiring background reduction of many orders of magnitude.  
  Total data rate amounts to $\sim$40\unit{Tb/s}.  
  The detector must take data for a decade or more at this rate, with near-continuous uptime. 
  
  For steady-state signals such as solar neutrinos, triggering on individual events in the presence of large backgrounds is a challenge that can be addressed with machine learning.  
  For burst signals, the triggering is a different problem: the general strategy is to read out all information on every channel within a tens-of-seconds time window, for the case of a triggered burst.  
  This leads to the subsequent problem of sifting the signal events and reconstructing sufficient information on a very short timescale to point back to the supernova.  
  The required timescale is minutes, or preferably seconds.  
  Both the event-by-event triggering and fast directional reconstruction can be addressed with fast machine learning.

  % Cross reference to previous section?

\subparagraph*{Existing and Planned Work:}

  There are a number of existing efforts towards the use of machine learning for particle reconstruction in neutrino detectors including water Cherenkov, scintillator, and liquid argon detectors. 
  These overlap to some extent with the efforts described in Sec.~\ref{sec:nuaccel}.  
  Efforts directed specifically towards real-time event selection and reconstruction are ramping up.
  Some examples of ongoing efforts can be found in Refs.~\cite{Abi_2020,Drielsma:2021jdv,Qian:2021vnh,Acciarri:2020ond,Abratenko:2020pbp,Wang:2020fjr,Psihas_2020}.  

\subsubsection{Direct Detection Dark Matter Experiments}

\subparagraph*{Context:} Direct dark matter (DM) search experiments take advantage of the vastly abundant DM in the universe and are searching for direct interactions of DM particles with the detector target material. 
The various target materials can be separated into two main categories, crystals and liquid noble gases, though other material types are subject to ongoing detector R\&D efforts~\cite{alexander2016dark, Schumann_2019}.

One of the most prominent particle DM candidates is the WIMP (weakly interacting massive particle), a thermal, cold DM candidate with an expected mass and coupling to Standard Model particles at the weak scale~\cite{Jungman_1996}. 
However, decades of intensive searches both at direct DM and at collider experiments have not yet been able to discover\footnote{The DAMA/NaI and subsequent DAMA/LIBRA experiment, claim the direct observation of DM particles in the galactic halo \cite{Bernabei_2013}, but the results are in tension with negative results from similar experiments~\cite{Schumann_2019}.} the vanilla WIMP while excluding most of the parameter space of the simplest WIMP hypothesis~\cite{Schumann_2019}. 
This instance has lead to a shift in paradigm for thermal DM towards increasingly lower masses well below 1\unit{GeV} (and thus the weak scale) \cite{B_hm_2004} and as low as a few keV, i.e. the warm DM limit \cite{Weinberg_2015}. 
Thermal sub-GeV DM is also referred to as light dark matter (LDM). 
Other DM candidates that are being considered include non-thermal, bosonic candidates like dark photons, axions and axion-light particles (ALPs)~\cite{Peccei:2006as, Svrcek:2006yi, Holdom:1985ag}.

The most common interactions direct DM experiments are trying to observe are thermal DM scattering off either a nucleus or an electron and the absorption of dark bosons under the emission of an electron. 
The corresponding signatures are either nuclear recoil or electron recoil signatures.

\subparagraph*{Challenges:} In all mentioned interactions, and independent of the target material, a lower DM mass means a smaller energy deposition in the detector and thus a signal amplitude closer to the baseline noise. 
Typically, the baseline noise has non-Gaussian contributions that can fire a simple amplitude-over-threshold trigger even if the duration of the amplitude above threshold is taken into account. 
The closer the trigger threshold is to the baseline, the higher the rate of these spurious events. 
In experiments which cannot read out raw data continuously and which have constraints on the data throughput, the hardware-level trigger threshold has thus to be high enough to significantly suppress accidental noise triggers.
%which can mean several $\sigma$ baseline resolution above the baseline.

In the hunt for increasingly lower DM masses, however, an as-low-as-possible trigger threshold is highly desirable, calling for a more sophisticated and extremely efficient event classification at the hardware trigger level. 
Particle-induced events have a known, and generally constant, pulse-shape while non-physical noise ``events" (e.g. induced by the electronics) generally have a varying pulse-shape which is not necessarily predictable. 
A promising approach in such a scenario is the use of machine learning techniques for most efficient noise event rejection in real-time allowing to lower the hardware-level trigger threshold, and thus the low mass reach in most to all direct DM searches, while remaining within the raw data read-out limitations imposed by the experimental set-up.

\subparagraph*{Existing and Planned Work:} Machine learning is already applied by various direct DM search experiments \cite{Khosa_2020, Szydagis:2021hfh, Simola_2019}, especially in the context of offline data analyses. 
However, it is not yet used to its full potential within the direct DM search community. 
Activities in this regard are still ramping up but with increasing interest, efforts, and commitment. 
Typical offline applications to date are the reconstruction of the energy or position of an event and the classification of events (e.g. signal against noise or single-scattering against multiple-scattering). 
In parallel R\&D has started on real-time event classification within the FPGA-level trigger architecture of the SuperCDMS experiment \cite{Agnese_2017} with the long-term goal of lowering the trigger threshold notably closer to the baseline noise without triggering on spurious events. 
While these efforts are being conducted within the context of SuperCDMS the goal is a modular trigger solution for easier adaption to other experiments.

\subsection{Electron-Ion Collider}

\subparagraph*{Context:}
        The Electron-Ion Collider (EIC) will support the exploration of nuclear physics over a wide range of center-of-mass energies and ion species, using highly-polarized electrons to probe highly-polarized light ions and unpolarized heavy ions. 
        The frontier accelerator facility will be designed and constructed in the U.S. over the next ten years. The requirements of the EIC are detailed in a white paper~\cite{Accardi:2012qut}, the 2015 Nuclear Physics Long Range Plan~\cite{Geesaman:2015fha}, and an assessment of the science by the National Academies of Science~\cite{NAS:2018eic}. 
        The EIC's high luminosity and highly polarized beams will push the frontiers of particle accelerator science and technology and will enable us to embark on a precision study of the nucleon and the nucleus at the scale of sea quarks and gluons, over all of the kinematic range that is relevant as described in the EIC Yellow Report~\cite{AbdulKhalek:2021gbh}. 

\subparagraph*{Challenges:}
        While the event reconstruction at the EIC is likely easier than the same task at present LHC or RHIC hadron machines, and much easier than for the High-Luminosity LHC, which will start operating two years earlier than the EIC, possible contributions from machine backgrounds form a challenge. 
        The expected gain in CPU performance in the next ten years as well as the possible improvement in the reconstruction software from the use of AI and ML techniques give a considerable margin to cope with higher event complexity that may come by higher background rates. 
        Software design and development will constitute an important ingredient for the future success of the experimental program at the EIC. 
        Moreover, the cost of the IT related components, from software development to storage systems and to distributed complex e-Infrastructures can be raised considerably if a proper understanding and planning is not taken into account from the beginning in the design of the EIC. 
        The planning must include AI and ML techniques, in particular for the compute-detector integration at the EIC, and training in these techniques. 
        
\subparagraph*{Existing and Planned Work:}
        
         Accessing the EIC physics of interest requires an unprecedented integration of the interaction region (IR) and detector designs. 
         The triggerless DAQ scheme that is foreseen for the EIC will extend the highly integrated IR-detector designs to analysis. 
         A seamless data processing from DAQ to analysis at the EIC would allow to streamline workflows, e.g., in a combined software effort for the DAQ, online, and offline analysis, as well as to utilize emerging software technologies, in particular fast ML algorithms, at all levels of data processing. 
         This will provide an opportunity to further optimize the physics reach of the EIC. 
         The status and prospects for ``AI for Nuclear Physics'' have been discussed in a workshop in 2020~\cite{Bedaque:2021bja}. 
         Topics related to fast ML are intelligent decisions about data storage and (near) real-time analysis. 
         Intelligent decisions about data storage are required to ensure the relevant physics is captured. 
         Fast ML algorithms can improve the data taken through data compactification, sophisticated triggers, and fast online analysis. 
         At the EIC, this could include automated alignment and calibration of the detectors as well as automated data-quality monitoring. 
         A (near) real-time analysis and feedback enables quick diagnostics and optimization of experimental setups as well as significantly faster access to physics results. 
         
\subsection{Gravitational Waves}
\subparagraph*{Context:}
        As predicted by Einstein in 1916, gravitational waves are fluctuations in the gravitational field which within the theory of general relativity manifest as a change in the spacetime metric. 
        These ripples in the fabric of spacetime travel at the speed of light and are generated by changes in the mass quadruple moment, as, for example, in the case of two merging black holes~\cite{PhysRevLett.116.061102}.
        To detect gravitational waves, the LIGO/Virgo/KAGRA collaborations employ a network of kilometer-scale laser interferometers~\cite{aLIGO, Acernese_2014, Affeldt_2014, PhysRevD.88.043007}.
        An interferometer consists of two perpendicular arms; as the gravitational wave passes through the instrument, it stretches one arm while compressing the other in an alternating pattern dictated by the gravitational wave itself. 
        Such length difference is then measured from the laser interference pattern.
        
        Gravitational waves are providing a unique way to study fundamental physics, including testing the theory of general relativity at the strong field regime, the speed of propagation and polarization of gravitational waves, the state of matter at nuclear densities, formation of black holes, effects of quantum gravity and more.
        They have also opened up a completely new window for observing the Universe and in a complementary way to one enabled by electromagnetic and neutrino astronomy.
        This includes the study of populations, including their formation and evolution, of compact objects such as binary black holes and neutron stars, establish the origin of gamma-ray bursts (GRBs), measure the expansion of the Universe independently of electromagnetic observations, and more~\cite{PhysRevLett.119.161101}. 
        
\subparagraph*{Challenges:}
        In the next observing run in 2022, LIGO, Virgo, and KAGRA will detect an increasing number of gravitational-wave candidates. 
        This poses a computational challenge to the current detection framework, which relies on matched-filtering techniques that match parameterized waveforms (templates) from simulations into the gravitational-wave time series data~\cite{PhysRevLett.116.061102, Vas2001, PhysRevD.44.3819}.
        Matched filtering scales poorly as the low-frequency sensitivity of the instrument improves and the search parameter space of the gravitational wave expands to cover spin effects and low mass compact objects.
        To estimate the physical properties of the gravitational wave, stochastic Bayesian posterior samplers, such as Markov-chain Monte Carlo and Nested Sampling, have been used until now. 
        Such analysis approaches can take up hours to days to complete~\cite{Abbott_2016}.
        The latency introduced by the current search and parameter estimation pipeline is non-negligible and can hinder electromagnetic follow-ups of time-sensitive sources like binary neutron stars, supernovae, and other, yet unknown, systems.
        
        Observations of gravitational-wave transients are also susceptible to environmental and instrumental noise.
        Transient noise artifacts can be misidentified as a potential source, especially when the gravitational-wave transients have an unknown morphology (e.g. supernovae, neutron star glitches).
        Line noise in the noise spectrum of the instruments can affect the search for  continuous gravitational waves (e.g. spinning neutron stars) and stochastic gravitational waves (e.g astrophysical background of gravitational waves from unresolved compact binary systems).
        These noise sources are difficult to simulate, and current noise subtraction techniques are insufficient to remove the more complex noise sources, such as non-linear and non-stationary ones. 

\subparagraph*{Existing and Planned Work:}
        In recent years, machine learning algorithms have been explored in different areas of gravitational-wave physics~\cite{Cuoco_2020}.
        CNNs have been applied to detect and categorize compact binary coalescence gravitational waves~\cite{PhysRevLett.120.141103, Kim_2015, PhysRevD.101.083006, George_2018, Gebhard_2019}, burst gravitational waves from core-collapse supernovae~\cite{Astone_2018, Chan_2020, Iess_2020}, and continuous gravitational waves~\cite{Dreissigacker_2019, Beheshtipour_2020}.
        Besides, recurrent neural networks (RNNs) based autoencoders have been explored to detect gravitational wave using an unsupervised strategy~\cite{moreno2021source}. FPGA-based RNNs are also explored to show the potential in low-latency detection of gravitational wave~\cite{que2021accelerating}.
        Applications of ML in searches of other types of gravitational waves, such as generic burst and stochastic background, are currently being explored.
        Moreover, probabilistic and generative ML models can be used for posterior sampling in gravitational-wave parameter estimation and achieve comparable performance to Bayesian sampler on mock data while taking significantly less time to complete~\cite{shen2019deterministic, gabbard2020bayesian,  PhysRevLett.124.041102}.
        ML algorithms are also being used to improve the gravitational-wave data quality and subtract noise.
        Transient noise artifacts can be identified and categorized from their time-frequency transforms and constant-Q transforms~\cite{Zevin_2017, Razzano_2018} or through examining hundreds of thousands of LIGO's auxiliary channels~\cite{iDQ2013}.
        These auxiliary channels can also be used to subtract quasi-periodic noise sources (e.g. spectral lines)~\cite{PhysRevD.101.042003, Ormiston_2020}.
        Although ML algorithms have shown a lot of promise in gravitational-wave data analysis, many of these algorithms are still at the proof-of-concept stage and have not yet been successfully applied in real-time analysis.
        Current efforts seek to create a computational infrastructure for low-latency analysis, improve the quality of the training data (e.g. expanding the parameter space, using a more realistic noise model), and better quantify the performance of these algorithms on longer stretches of data.

\subsection{Biomedical engineering}

\subparagraph*{Context:} We have seen an explosion of biomedical data, such as biomedical images, genomic sequences, and protein structures, due to the advances in high-resolution and high-throughput biomedical devices. 
AI-augmented reality-based microscopy~\cite{Chen2019-ze} enables automatic analysis of cellular images and real-time characterization of cells. 
Machine learning is used \textit{in-silico} prediction of fluorescent labels, label-free rare cell classification, morphology characterization, and RNA sequencing~\cite{Christiansen2018-eu,Wang2020-lr,Siu2020-kd,Tang2018-mj,Li2020-cx}. 
For in-situ cell sorting, real-time therapy response prediction, and augmented reality microscope-assisted diagnosis~\cite{Chen2019-ze,Nitta2018-bc,Sakellaropoulos2019-tq}, it is important to standardize and optimize data structure in deep learning models to increase speed and efficiency. 
Various machine-learning-based algorithms for detecting hemorrhage and lesions, accelerating diagnosis, and enhancing medical video and image quality have also been proposed in biopsy analysis and surgery assistance.
        
\subparagraph*{Challenges:} A major challenge for clinical application of ML is inadequate training and testing data. The medical data annotation process is both time-consuming and expensive for large image and video datasets which require expert knowledge. 
The latency of trained models' inference also introduces computational difficulties in performing real-time diagnosis and surgical operation. 
The quality of services for time-critical healthcare requires less than 300 milliseconds as real-time video communication \cite{Shukla2019-bz}. For reaching 60 frames per second (FPS) high-quality medical video, the efficiency and performance of a deep learning model become crucial.
        
\subparagraph*{Existing and Planned Work:} Many changes in ML algorithms have involved improvements to performance both in accuracy and inference speed. Some state-of-art machine learning models can reach a high speed for inference. 
For example, \textit{YOLOv3-tiny}~\cite{Adarsh2020-hq}, an object detection model commonly used for medical imaging, can process images at over 200 FPS on a standard dataset with producing reasonable accuracy. 
Currently both GPU- and FPGA-based~\cite{Satpathy2020-gs,Chang2020-ob,Zhang2020-bb}, distributed networks of wireless sensors connected to cloud ML (edge computing), and 5G-high-speed-WiFi-based ML models are deployed in medical AI applications~\cite{Chen2018-qx,Zhang2020-ze,Morocho-Cayamcela2019-gt}. 
ML models for fast diagnosis of stroke, thrombosis, colon polyps, cancer, and epilepsy have significantly reduced the time in lesion detection and clinical decision~\cite{Lee2020-oj,Nafee2020-yy,Nogueira-Rodriguez2020-zd,Bagheri2019-ee,Horie2019-hz}. 
Real-time AI-assisted surgery can improve perioperative workflow, perform video segmentation~\cite{Volkov2017-oy}, detection of surgical instruments \cite{Choi2017-iv}, and visualization of tissue deformation~\cite{Tonutti2017-vv}. 
High-speed ML is playing a critical role in digital health, \textit{i.e.}, remote diagnosis, surgery, and monitoring \cite{Zhang2020-ze}.

\subsection{Health Monitoring}

\subparagraph*{Context:} Our habits and behaviors affect our health and wellness. Unhealthy  behaviors such as smoking, consuming excessive alcohol, or medication non-adherence often has an adverse effect on our health~\cite{baker2000health,klesges1989smoking,sokol2005impact,white2013burden}. 
Traditional behavior monitoring approaches relied on self-reports, which were often biased and required intense manual labor~\cite{althubaiti2016information}. 
With the advent of mobile and wearable devices, it is gradually becoming possible to monitor various human behaviors automatically and unobtrusively. 
Over the years, researchers have either developed custom wearable hardware or have used off-the-shelf commercial devices for mobile and wearable health (mHealth) monitoring~\cite{dong2012new,parate2014risq,ali2012mpuff,sen2020annapurna,bi2018auracle,zhang2020necksense,mishra2020continuous}. The automatic and unobtrusive monitoring capability of these devices makes it possible to detect, identify and monitor behaviors, including unhealthy behaviors in a free-living setting.   
        
\subparagraph*{Challenges:} There are various challenges associated with monitoring habits and behaviors using wearable devices. Firstly, these devices should be capable of monitoring unhealthy behaviors accurately, and in real-time. 
The occurrence of these unhealthy behaviors in a free-living setting is often sparse as compared to other behaviors and thus it is important to spot them accurately, whenever they occur. 
Most existing systems take an offline ML approach of detecting these unhealthy behaviors, where the ML algorithm identifies these behaviors  well after they have occurred. An offline approach prevents providing interventions that can minimize unhealthy behaviors. 
Thus, it is necessary to develop ML approaches that can detect these behaviors online, and in real-time, so that interventions such as just-in-time adaptive interventions (JITAIs) can be delivered. 
Secondly, since these devices capture sensitive information, it is necessary to ensure that an individual's privacy is preserved. 
Privacy-preserving approaches such as locally processing the data on-device can be taken so that critical information does not leave the device. 
%Thirdly, the amount of data produced by these wearable devices is large and that requires excessive data labeling effort. An automated data labeling approach is thus necessary to reduce manual data labeling burden. 
Finally, these behaviors can occur in various heterogeneous environments and thus the health monitoring system should be agnostic to where the behavior occurs. 
Such monitoring requires developing multiple machine learning models for diverse environments.  

\subparagraph*{Existing and Planned Work:} While existing work has ventured in various directions, there is a growing need for sensing health biomarkers correctly and developing ML approaches that are fast and can accurately identify these biomarkers. 
Researchers have focused on developing novel sensing systems that can sense various health behaviors and biomarkers~\cite{holz2017glabella,pham2020wake,chun2020intraoral,bui2019ebp,li2020noninvasive,echterhoff2020personal,bedri2020fitbyte}. Historically, most of these novel sensing techniques were tested in controlled settings, but more recently researchers are ensuring that these systems can work seamlessly in free-living settings as well. 
This often requires developing multiple ML models, each catering to a specific context and environment. 
A new trend in this field has started relying on implementing models that can be implemented on-device and are both quick and accurate in detecting these behaviors. 
In addition to providing real-time interventions~\cite{nahum2018just,thomas2015behavioral}, on-device monitoring of these behaviors can reduce privacy concerns~\cite{sadek2019privacy}. 
However, since wearable devices themselves might not be capable of processing the data, federated machine learning approaches are also being explored recently by several researchers~\cite{rieke2020future}.

\subsection{Cosmology}

\subparagraph*{Context:} Cosmology is the study of the Universe's origin (big bang), evolution, and future (ultimate fate). 
The large-scale dynamics of the universe are governed by gravity, where dark matter plays an important role, and the accelerating expansion rate of the universe itself, caused by the so-called dark energy. 
A non-exhaustive list of cosmological probes includes type Ia supernovae~\cite{Riess_1998, Perlmutter_1999, Betoule_2014, Scolnic_2018, Abbott_2019}, cosmic microwave background~\cite{Fixsen_1996, Spergel_2003, Komatsu_2011, PC_2016, PC_2020}, large-scale structures (including baryon acoustic oscillation)~\cite{Eisenstein_2005, Percival_2010, Delubac_2015, Abbott_2019b}, gravitational lensing~\cite{Bacon_2000, Bacon_2003, Collett_2014, Suyu_2017, Heymans_2020} and 21\unit{cm} cosmology~\cite{McQuinn_2007, Pritchard_2012, Maartens_2015, Beardsley_2016}.
    
\subparagraph*{Challenges:} As astronomy is approaching the big data era with next-generation facilities, such as the Nancy Grace Roman Space telescope, Vera C. Rubin Observatory, and Euclid telescope, the uncertainty budget in the estimation of cosmological parameters is no longer expected to be dominated by statistical uncertainties, but rather by systematic ones; understanding such uncertainties can lead to attaining sub-percent precision. 
On the other hand, the immense stream of astronomical images will be impossible to analyze in a standard fashion (by human interaction); new automated methods are needed to extract valuable pieces of cosmological data.
    
\subparagraph*{\textbf{Existing and future work}:} Current efforts are focused on applying ML techniques to study the influence of systematic biases on available analysis methods (e.g., for purposes of fitting or modeling) or on developing new methods to overcome present limitations; for example CNNs can be adapted to spherical surfaces to generate more accurate models when producing weak lensing maps~\cite{Perraudin_2019}, or to remove noise from cosmic microwave background maps~\cite{Petroff_2020}. 
In addition, discovery and classification engines are being developed to extract useful cosmological data from next-generation facilities~\cite{Narayan_2018, Mahabal_2019, Forster_2020, Moller_2020}. 
Furthermore, ML is also being used in cosmological simulations to test new analyses and methods and to set the foundations for the first operation of such new facilities~\cite{Kamdar16, Rodriguez18, Villaescusa-Navarro20}. 
An extensive list of published ML applications in cosmology can be found in \url{https://github.com/georgestein/ml-in-cosmology}.

\subsection{Plasma Physics}

\subparagraph*{Context:} The focus of this description is on the Plasma Physics/Fusion Energy Science domain with regard to the major system constraints encountered for existing and expected algorithms and data representations when dealing with the challenge of delivering accelerated progress in AI---enabled deep machine learning prediction and control of magnetically-confined thermonuclear plasmas.  
Associated techniques have enabled new avenues of data-driven discovery in the quest to deliver fusion energy---identified by the 2015 CNN ``Moonshots for the 21st Century'' televised series as one of 5 prominent grand challenges for the world today.

\subparagraph*{Challenges:} An especially time-urgent and challenging problem is the need to reliably predict and avoid large-scale major disruptions in ``tokamak systems'' such as the EUROFUSION Joint European Torus (JET) today and the burning plasma ITER device in the near future---a ground-breaking \$25B international burning plasma experiment with the potential capability to exceed ``breakeven'' fusion power by a factor of 10 or more with “first plasma” targeted for 2026 in France.  The associated requirement is for real-time plasma forecasting with control capabilities operative during the temporal evolution of the plasma state well before the arrival of damaging disruptive events.  
High-level supervisory control of many lower-level control loops via actuators (analogous to advanced robotics operations) will be essential for ITER and future burning plasmas to protect the facility and to avoid operational limits (for magnets, walls, plasma position, stability, etc.) while optimizing performance.   

\subparagraph*{Existing and Planned Work:} In short, an overarching goal here involves developing realistic \textit{predictive plasma models of disruptions integrated with a modern plasma control system to deliver the capability to design experiments before they are performed}. 
The associated novel AI-enabled integrated modeling tool would clearly be of great value for the most efficient and safe planning of the expensive discharges in ITER and future burning plasmas. 
Verification, validation, and uncertainty quantification of associated components would include: (1) development of predictive neural net models of the plasma and actuators that can be extrapolated to burning plasma scales via advanced Bayesian reinforcement learning methods that incorporate prior information into efficient inference algorithms; (2) systematic well-diagnosed experimental validation studies of components in the integrated plasma forecasting models involving massive amounts of data from major tokamak experiments worldwide (e.g., DIII-D in the US, KSTAR \& EAST in Asia, JET in Europe, followed by JT60 SA---the large superconducting device in Japan that will precede ITER).  
This would ideally lead to a mature AI-enabled comprehensive control system for ITER and future reactors that feature integration with full pilot-plant system models. 

At present, a key challenge is to deliver significantly improved methods of prediction with better than 95\% predictive accuracy to provide advanced warning for disruption avoidance/mitigation strategies to be effectively applied before critical damage can be done to ITER.  Significant advances in the deployment of deep learning recurrent and CNNs are well illustrated in Princeton’s Deep Learning Code---``FRNN''---that have enabled the rapid analysis of large complex datasets on supercomputing systems.  
Associated acceleration of progress in predicting tokamak disruptions with unprecedented accuracy and speed is described in~\cite{plasmaref}. 
Included in this paper (and extensive references cited therein) are descriptions of FES data representation for physics features (density, temperature, current, radiation, fluctuations, etc.) and the nature of key plasma experiments featuring detectors/diagnostics with frame (event-based) level of accuracy accounting for required ``zero-D'' (scalar) and higher-dimension signals and real-time resolution recorded at manageable data rates. 
Rough future estimates indicate that ITER will likely require dealing with the challenge of processing and interpreting exabytes of complex spatial and temporal data. 

Since simulation is another vital aspect of ITER data analysis, dealing with the associated major computational expenses will demand the introduction of advanced compressional methods.  
More generally, real-time predictions based on actual first-principles simulations are important for providing insights into instability properties and particle-phase space dynamics. 
This motivates the development of an AI-based ``surrogate model''---for example, of the well-established HPC ``gyrokinetic'' particle-in-cell simulation code GTC~\cite{plasmaref2} that would be capable of accurately simulating plasma instabilities in real-time.  
Data preparation and training a surrogate model – e.g., ``SGTC''---provides a clear example of the modern task of integration/connection between modern High Performance Computing (HPC) predictive simulations with AI-enabled Deep Learning/Machine Learning campaigns. 
These considerations also serve to further illustrate/motivate the need to integrate HPC \& Big Data ML approaches to expedite the delivery of scientific discovery.

As a final note, the cited paper~\cite{plasmaref} represents the first adaptable predictive DL software trained on leadership class supercomputing systems to deliver accurate predictions for disruptions across different tokamak devices (DIII-D in the US and JET in the UK).  
It features the unique statistical capability to carry out efficient ``transfer learning'' via training on a large database from one experiment (i.e., DIII-D) and be able to accurately predict disruption onset on an unseen device (i.e., JET).  
In more recent advances, the FRNN inference engine has been deployed in a real-time plasma control system on the DIII-D tokamak facility in San Diego, CA.  
As illustrated in slides 18 through 20 of the attached invited presentation slide deck, this opens up exciting avenues for moving from passive disruption prediction to active real-time control with subsequent optimization for reactor scenarios.

\subsection{ML for Wireless Networking and Edge Computing}

\subparagraph*{Context:} Wireless devices and services have become a crucial tool for collecting and relaying big data in many scientific studies. Moreover, mobility information has proven to be extremely useful in understanding human activities and their impact on the environment and public health. The exponential growth of data traffic is placing significant pressure on the wireless infrastructure. In particular, inter-cell interference causes large variability in reliability and latency. To meet user demands for data communication and value-added AI/ML services, wireless providers must 1) develop more intelligent learning algorithms for radio resource management that adapt to complicated and ever-changing traffic and interference conditions; and 2) realize many ML/AI computations and functionalities in edge devices to achieve lower latency and higher communication efficiency.

\subparagraph*{Challenges:} Conventional implementations of ML models, especially deep learning algorithms, lag far behind the packet-level dynamics for utility.  Moreover, existing ML/AI services are often performed in the cloud for efficiency at the expense of communication overhead and higher latency.  A major challenge in the wireless networking and edge computing context is to build a computing platform that can execute complex ML models at relevant timescales ($< 10$ ms) within small cell access points.

\subparagraph*{Existing and planned work:} Researchers have proposed a variety of learning algorithms to perform specific radio resource management tasks using artificial neural networks~\cite{
calabrese2018learning,
challita2018proactive,
huang2020deep,
zhu2020toward}.
Some of the first proposals to train a NN to perform transmit power control adopts supervised learning~\cite{sun2018learning,liang2019towards}.  
More recent proposals adopt deep reinforcement learning approaches that work better with channel and network uncertainties and require little training data {\em a priori}~\cite{nasir2020deep,zhao2019deep,liang2019spectrum,meng2020power}.
A number of works are focused on the convergence of edge computing and deep learning~\cite{chen2019deep,zhang2019deep,wang2020convergence}.  
A specific set of work is on federated learning where participants jointly train their models in lieu of sending all their data to a central controller for training purposes~\cite{niknam2020federated,amiri2020federated,chen2021convergence,ren2020scheduling}.
All of the preceding work basically ends at the simulation stage for the lack of practical ML/AI solutions that are fast and computationally efficient at the same time.  More specifically, the research challenge is to develop a computing platform that can execute complex ML models at a very fast timescale ($<$ 10\unit{ms}) and can also be equipped in small cell access points.
One project with a potentially very high impact is to map intelligent radio resource management algorithms (such as that of~\cite{nasir2020deep}) onto an FPGA device suitable for deployment in a large network of connected and interfering access points.
Another interesting project is to build a federated learning system to conduct time-sensitive ML for Internet-of-Things (IoT) devices where transferring data to centralized computing facilities is latency-prohibitive.  
This opens up entirely new possibilities for low-cost closed-loop IoT devices in healthcare, smart buildings, agriculture, and transportation.

%%%%%%%%%%%%%%%%%%%%%%%%%%%%%%%%%%%%
\pagebreak
\section{Key areas of overlap}
\label{sec:overlaps}
% \textcolor{red}{Section Editors: Mia Liu (Purdue),...}
\noindent 
% \textcolor{blue}{
% The focus of these sections is to source material from Sec.~\ref{sec:apps} and lay out similarities and differences.
% }
Real-time, accelerated AI inference show promises in improving the discovery potential at current and planned scientific instruments across the domains as detailed in Sec.~\ref{sec:apps}. Design of high performant specialty systems for real-time/accelerated AI applications requires particular attention to the figure-of-merit of the target domain's ML algorithm. 
It might be dominated by its latency per inference, computational cost (e.g., power consumption), reliability, security, and ability to operate in extreme environments (e.g., radiation). 
For instance, ML might need to: trigger acquisition systems for rare events with $\sim$100\unit{ns} latency on the Large Hadron Collider~\cite{Duarte:2018ite}; analyze multi-channel ambulatory health monitors at kilohertz frequencies where wireless transfer of data is not possible due to power limitations ($\sim$50 iPhone batteries/day for data transfer) or security requirements; or to keep pace with materials spectroscopy data streams on the order of terabits per second~\cite{Hart2017-bf}.
Furthermore, real-time analysis of advanced scientific instrumentation must have an uninterrupted allocation of computing resources and patient sensitive information processed by wireless health devices must be secured. Such features and characteristics create quantifiable guidelines for understanding distinctions and commonalities among domains and applications. 
Thereby, we can coordinate efforts towards creating fundamental design principles and tools, which may address needs across seemingly disparate domains. 
Appropriate data representation is an essential first step of the design process as it determines the choice of NN architecture to be implemented in real-time systems that need to meet the performance targets outlined above. 
Prominent data representations of different scientific instruments are summarized below. 
Other areas of overlap across domains such as NN and hardware co-design tools and workflows, NN complexity reduction with quantization and pruning are also recent technology advancements in real-time/accelerated AI and therefore are outlined in Section~\ref{sec:technolog_sota}.

\subsection{Data representations}
\begin{figure*}[tbh!]
     \centering
     \includegraphics[width = 0.55\textwidth]{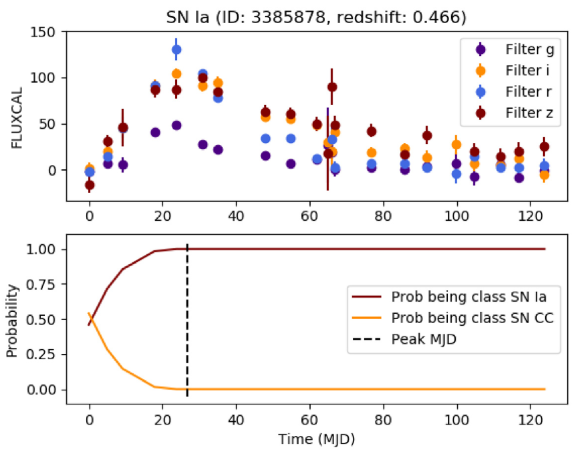}
     \caption{Simulated type Ia supernova light-curve and classification. 
     Top: calibrated flux evolution in different DES band-passes as a function of normalized time (the first photometric measurement is set to time equals zero). 
     Bottom: Baseline RNN classification probability evolution with respect of time, no host-galaxy redshift information was provided. 
     At each photometric measurement, classification probability is obtained. The maximum light of the simulated supernova is shown in a gray dashed line and the simulated redshift of the supernovae is shown on the top $z = 0.466$. 
     We highlight that redshift is not used for this classification but can improve results. 
     Our baseline RNN classifies this light-curve as type Ia SN with great accuracy before maximum light, it only requires a handful of photometric epochs. \cite{supernova_2019}.}
     \label{fig:supernova}
 \end{figure*}

Data representation used in a particular domain influences both the computation system and data storage. One global classification for data representations across domains can be considered as being into raw versus reconstructed data. 
The data representation often varies depending on the stage of the reconstruction and the upstream steps in the data processing pipeline. 
Existing applications include fully connected NNs that often take pre-processed expert feature variables as inputs or CNNs when the data is of image nature. 
On-going development of domain knowledge-inspired NN algorithms could further take advantage of the expert features in the accuracy and efficiency as detailed below.
To fully exploit the power of advanced NNs and bring it closer to data creation for minimum information loss, a more suitable representation of the raw data, e.g as point clouds, needs to be employed. 
Prominent representations for raw data from different experimental and measurement systems are:
    \begin{itemize}
        \item \textbf{Spatial Data}: Used for describing physical objects in geometric space. 
        There are two main types, called vector and raster data. 
        Vector data, in turn, can be comprised of points, lines, or polygons. Raster data refers to a grid of pixels, such as images, but pixels can also represent other measurements such as intensity, charge, field strength, etc.
        \item \textbf{Point Clouds}: Can be considered a type of spatial data. 
        This data representation is created by collating a set of spatial data, i.e., points in a 3D space, that usually form an object in space collectively. 
        \item \textbf{Temporal Data}: Used to represent the state of a system/experiment at a particular time. 
        Data collected across time, in a specific order, is classified in this manner. 
        Time-series data is a subset of this representation, where data is sampled at regular time intervals. 
        An example of time-series data can be seen in Fig.~\ref{fig:supernova}, for the specific case of supernova classification.
        \item \textbf{Spatio-Temporal Data}: Measurements and observations of a system can be collected across both the space and time dimensions. 
        In that case, the data can be considered spatio-temporal. 
        \item \textbf{Multispectral Data}: Used to represent outputs of multiple sensors that capture measurements from multiple bands of the electromagnetic spectrum. 
        Multispectral representation is commonly used in the context of imaging, involving sensors that are sensitive to different wavelengths of light. 
        This usually involves in the order of a few to 10s of spectra.    
        \item \textbf{Hyperspectral Data}: Used to represent measurements from a high number of spectra, e.g., in the order of 100s. 
        These images collected from different narrow-band spectra are combined into a so-called hyperspectral cube with three main dimensions. 
        The first two reference the 2D spatial placement (e.g., earth's surface) while the third dimension represents the complete spectrum content at each ``pixel'' location.  
    \end{itemize}  

In Table~\ref{tab:representations}, we match these data representations to scientific application domains and give a brief description.  
We highlight the data representations which are particularly important for a specific domain.  
We will give more detailed examples below.

Cost of data communication (in terms of latency) and data storage (in terms of the cost of acquiring and managing the physical storage resources) present important challenges. Particularly, application domains, which require real-time analysis and/or real-time feedback demand highly optimized data analytics solutions. 
Applications that rely on hyper-spectral data are faced with an ever-increasing rate of data input across the electromagnetic spectrum. 
High-speed data reduction is required in these domains. 
Applications that generate large-scale point clouds similarly demand efficient compression on their spatial data. 
Application domains that handle multi-spectral data with limited spatial resolution require ultra-fast reconstruction in order to enable real-time control feedback. Another challenge is posed by applications that rely on accurate analysis of streaming time-series data, yet they are forced to perform under highly limited storage and communication resources, either due to privacy and security concerns or limitations of the associated edge devices.  

Some current efforts in developing ML solutions to data processing front-ends focus on developing autoencoder based compression engines~\cite{ieee_nss_talk_1_2020, DiGuglielmo:2020eqx}. 
ML-based dimensionality reduction for hyper-spectral data is another direction which has drawn attention~\cite{Agar2019}. 
Deep learning-based approaches are investigated for image reconstruction; the field of material sciences being one of the most active fields in that regards~\cite{Schmidt_nature2019}.

\begin{table*}[]
    \centering
    \footnotesize
    \begin{tabular}{|c|c|c|c|c|c|c|}
    \hline
       Domain & Spatial & Point Cloud & Temporal & Spatio- & Multi/Hyper- & Examples\\
              &         &             &          & Temporal & spectral & \\
    \hline
        LHC & \checkmark \checkmark & \checkmark \checkmark & \checkmark & \checkmark & -- & detector reconstruction\\
        Belle-II/Mu2e & \checkmark \checkmark & \checkmark \checkmark & -- & -- & -- & track reconstruction\\
        Material Synthesis & \checkmark & -- & \checkmark & \checkmark \checkmark & \checkmark \checkmark & high-speed plasma imaging \\
        %  &  &  & &  &  & plasma plume\\
        Accelerator Controls & \checkmark & -- & \checkmark \checkmark & -- & -- & beam sensors\\
        Accelerator neutrino & \checkmark \checkmark & \checkmark \checkmark & \checkmark & \checkmark & -- & detector reconstruction \\
        Direct detection DM & \checkmark \checkmark & \checkmark \checkmark & \checkmark & \checkmark & -- & energy signatures \\
        EIC & \checkmark \checkmark & \checkmark \checkmark & \checkmark & \checkmark & -- & detector reconstruction\\
        Gravitational Waves & \checkmark & -- & \checkmark \checkmark & -- & -- & laser inference patterns\\ 
        Biomedical engineering & \checkmark \checkmark & -- & -- & \checkmark \checkmark & -- & cell and tissue images \\  
                            %   &  &  &  &  &  & genomic sequence\\  
        Health Monitoring & \checkmark & -- & \checkmark \checkmark & \checkmark & \checkmark & physiological sensor data\\         
        Cosmology & \checkmark \checkmark & \checkmark \checkmark & \checkmark \checkmark & \checkmark  & \checkmark \checkmark & lensing/radiation maps\\         
        Plasma Physics & \checkmark & -- & \checkmark \checkmark & \checkmark & -- & detector actuator signals\\         
        Wireless networking & -- & -- &  \checkmark \checkmark & -- & -- & electromagnetic spectrum \\ 
    \hline
    \end{tabular}
    \caption{Types of data representations and their relevance for the scientific domains discussed in this paper; \checkmark \checkmark = Particularly important for domain, \checkmark = Relevant for domain}
    \label{tab:representations}
    \normalsize
\end{table*}

%%%%%%
\subsubsection{Expert Feature DNNs}

% A simple approach is to multivariate anlysis is to give as input known expert features
% Positive, more interpretable, more efficient
% Negative, limits the way that we can learn from the data, computational efficiency maybe
% A lot of interest in embedding domain expertise into the nerual network learning process

One straightforward approach to building powerful domain-specific ML algorithms is to start with expert domain features and combine them in a neural network or other multivariate analysis technique.  
This embedded expertise has inherent advantages because the input features are interpretable, and correlations between features can yield insight into a particular task while optimizing performance.  
Furthermore, depending on the computational complexity of the domain features, the computation efficiency of such a machine learning approach can be greater than the direct use of raw features.  
However, the downside is that, by using expert features, we rely entirely on the informativeness of such new features.    

Therefore, there is a lot of interest in automating the process of building informative new features from raw features.  
In image classification tasks, for example, a lot of progress has been made in extracting high-level data representations through deep neural networks DNNs~\cite{Goodfellow_2016}. 
In DNNs, layers of neurons above the original input signal are built to ensure that each new layer captures a more abstract representation of the data. 
Each layer constructs new features by forming nonlinear combinations of the features in the layer below. 
This hierarchical approach to feature construction has been effective in disentangling factors of variation in the data~\cite{Hinton_2006,Bengio_2013,Goodfellow_2016}, and has been useful to construct informative and meaningful representations. 
In astronomical images, for example, a DNN starts with low-level pixel information, gradually capturing at upper layers edges, motifs, and eventually entire objects (e.g., galaxies) to provide a broad view of the Universe~\cite{Sanchez_2018,Huertas_Company_2018}. 
The same applies to other fields of science. 
For example, detecting particles in large accelerators requires transforming low-level signals into dynamic patterns that can be ascribed to specific particles~\cite{Belayneh_2020}. 
In medical imaging, there is a need to quickly identify abnormal tissue from low-level pixel information by gradually capturing global tissue patterns~\cite{Bychkov_2018}. 
The importance of transforming the initial input data into meaningful abstract representations cannot be overstated: it remains one of the most powerful properties of modern neural network architectures.  

Several challenges exist in the construction of increasingly abstract representations using DNNs. 
One challenge is to incorporate domain knowledge (e.g., physical constraints) into the neural network model. 
This is important to address the need for excessive amounts of data when training a DNN and narrow the gap in representational bias between the model and target concept. Under scarce data but abundant domain expertise, adding domain knowledge can expedite the training process~\cite{Xie_2021}, as well as improving the model generalization performance. 
Another challenge is to develop tools for model interpretability by explaining the semantics of the representations embedded at each layer~\cite{Chakraborty_2017}. This is challenging due to the distributed representation of information in the network architecture. 

Despite the lack of a formal mechanism to attain a seamless integration between a statistical model and domain knowledge, current approaches point to interesting directions, e.g., using knowledge to add training data or to change the loss function~\cite{Vo_2017}. 
Model interpretability in DNNs has seen an upsurge in research over the past years~\cite{Chakraborty_2017}. 
Commonly, studies look at individual units and their activation patterns to elucidate what is learned across layers of neurons.

\subsubsection{Frame-based images}
\begin{figure*}[tbh!]
    \centering
    \includegraphics[width = 0.85\textwidth]{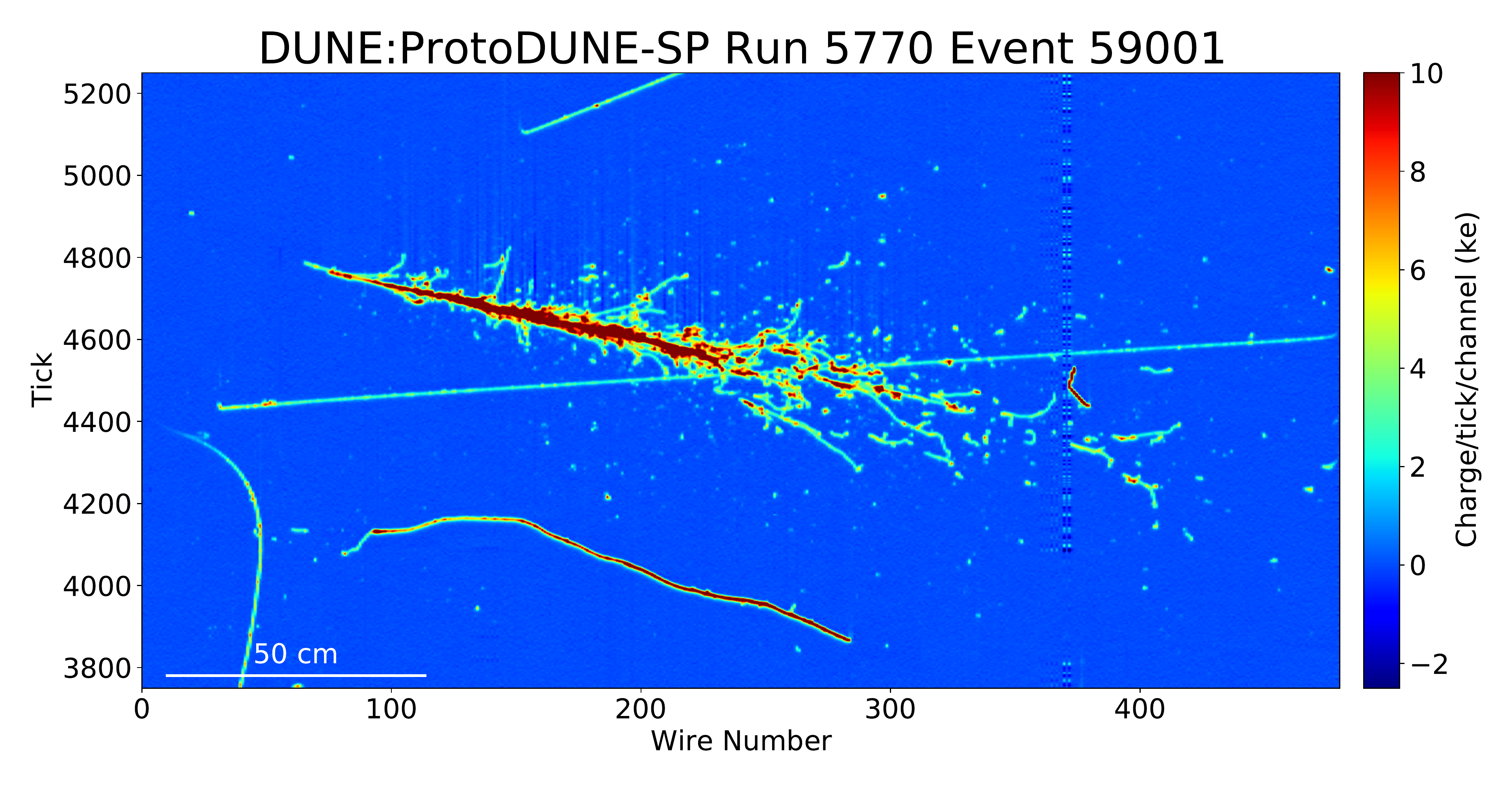}
    \caption{A 6\unit{GeV/c} electron event in the ProtoDUNE detector. 
    The x-axis shows the wire number. 
    The y-axis shows the time tick in the unit of 0.5$\mu s$. 
    The color scale represents the charge deposition.\cite{}}
    \label{fig:repframe}
\end{figure*}

Frame-based images are a suitable representation of the experimental data in multiple domains such as neutrino detection with time projection chambers in particle physics. 
An example of this data representation can be seen in Fig.~\ref{fig:repframe} for an electron deposition in the ProtoDUNE neutrino detector.  
A spatial frame is shown by plotting the time coordinate ``Tick'' and wire position in space. 
Recent developments in neural network architectures exploit the sparsity of the images to reduce the computation complexity for real-time/accelerated ML applications. 
Other types of experimental data in HEP and many other domains can also be processed to be represented as frame-based images, although often not without information loss. 

\subsubsection{Point clouds}
\begin{figure*}[tbh!]
     \centering
     \includegraphics[width=0.85\textwidth]{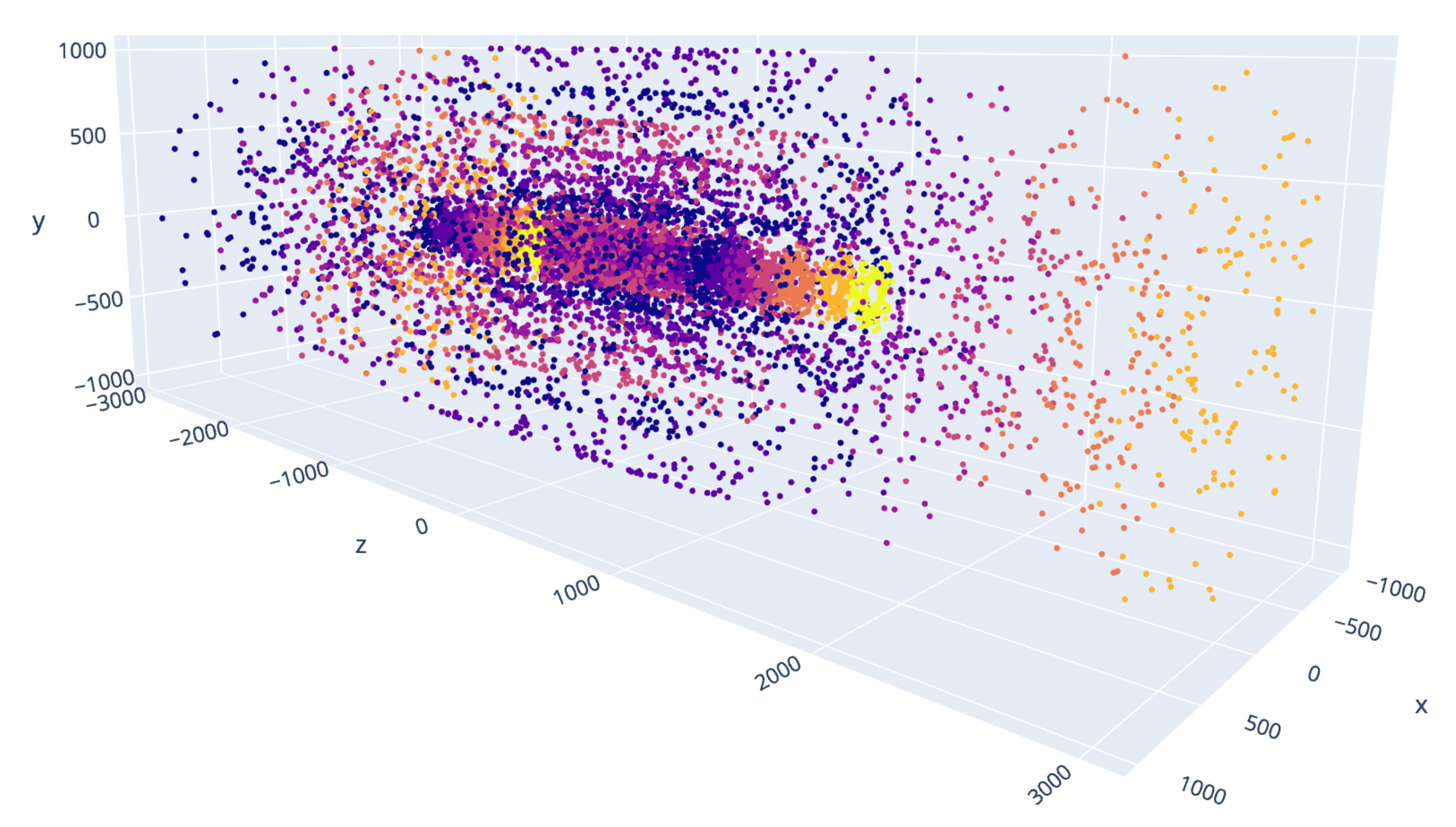}
     \caption{Visualization of particle tracking hits in 3D space from the TrackML Kaggle dataset~\cite{cmseventdisplay}}
     \label{fig:repcloud}
 \end{figure*}

 Point cloud data representation is often used in HEP, where multiple frames of event-based measurements collected by a large number of detectors are combined into a data set. 
 Across many HEP applications point clouds commonly help to represent particle jets with data sizes exceeding Pb/s. 
 More broadly, point clouds can be used to capture any 3D space event and interactions of moving parts in space. 
 A point cloud visualization of the CMS detector at the LHC is shown in Fig.~\ref{fig:repcloud}.  
 Remnants of proton-proton collisions create sensors signals in a customized and optimized detector geometry and points are illustrated in space. Various types of scan-based imaging data can be represented as point clouds. 
 Other domains such as CT and PET scanning in biomedical engineering and virtual reality also utilize this representation for imaging. 3D scanners used for product design, solid object modeling, architecture, and infrastructure design leverage point clouds as well. Many of these imaging tasks generate point clouds of sizes in the order of several GB to TB. 
 Domains sharing point cloud representation (e.g., HEP and biomedical imaging) also commonly involve spatial characteristics. 
 
 \subsubsection{Multi-/Hyperspectral Data}
\begin{figure*}[tbh!]
     \centering
     \includegraphics[width = 0.55\textwidth]{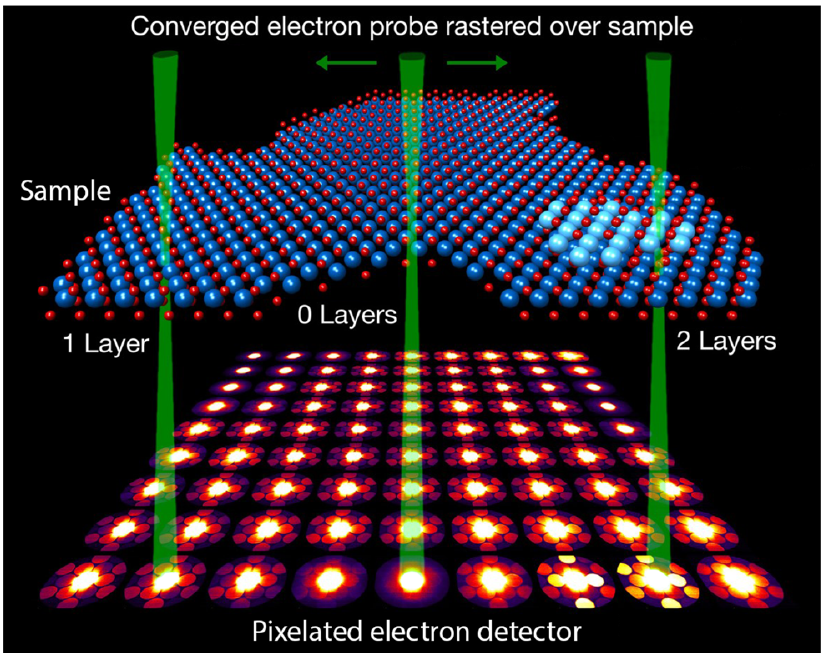}
     \caption{Experimental 4D-STEM measurement of a dichalcogenide 2D material. 
     Atomic map is inferred from the data, each diffraction pattern represents an average of $7\times 7$ experimental images, green STEM probes are labeled for regions of the sample with one layer, vacuum, and two layers~\cite{ophus_2019}.}
     \label{fig:rephyper}
 \end{figure*}

 Multispectral data is common between wireless health monitoring and wireless communication systems. 
 A set of physiological sensors, often representing different modalities, are combined into a multispectral data set for health monitoring and intervention systems. 
 For wireless communication, signal interference and network traffic conditions are captured via multispectral data. 
 Both domains capture this data across the time domain, so also exhibit temporal features. 
 Furthermore, in both domains generated data size can be considered relatively smaller (ranging from 100s of Mb/s to 10s of Gb/s), compared to the rest of the domains discussed in this article. 
 Hyperspectral data is used across many astronomy applications, medical imaging, and electron microscopy, which is used to drive many materials science design and discovery applications. 
 An example of hyperspectral data in electron microscopy is shown in Fig.~\ref{fig:rephyper}.  
 An electron probe is rastered over a sample under study and diffraction patterns are captured on a pixelated detector.  
 The pixelated detector captures many images as the electron probe is scanned across the sample.  
 Emerging multimessenger astronomy applications further emphasize the utility of hyperspectral data representations combining observations from a wide array of detectors and telescopes. 
 
\subsubsection{Time-series data}
\begin{figure*}[tbh!]
     \centering
     \includegraphics[width = 0.85\textwidth]{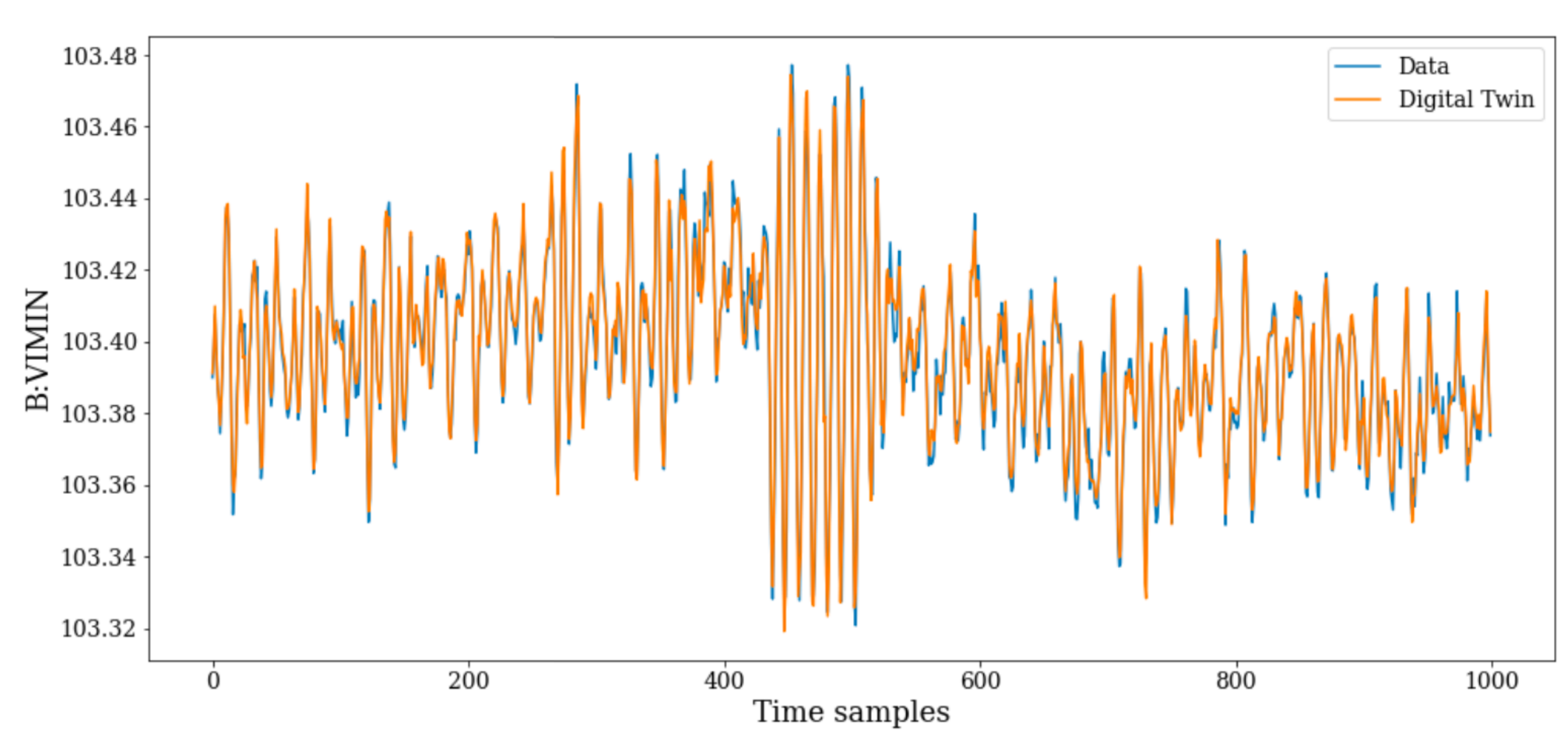}
     \caption{Selected test data (blue) versus prediction values (orange) from the Booster LSTM surrogate model for the Booster proton synchrotron complex~\cite{John:2020sak}.}
     \label{fig:reptime}
 \end{figure*}

Time-series data is common in experiments that observe dynamically evolving systems in processes such as synthesis for material discoveries or the temporal evolution of the plasma state in nuclear fusion experiments. 
It can be a measurement of high-speed temporally resolved imaging in material science or physics features (density, temperature, current, radiation, fluctuations, etc.) or spatial features of evolving plasma state, as a function of time. 
In-situ diagnostics of the time-series data can either provide alerts to terminate an experiment early that indicates undesired outcome in material science without performing the entire experiment and offline analysis that is time-consuming and computationally expensive, thus improves the experiment operation efficiency and accelerates discoveries of material of desired properties. 
This is illustrated in Fig.~\ref{fig:reptime} for accelerator controls at the Fermilab Booster accelerator.  
In this application, magnet voltages that steer proton beams around a synchrotron are recorded at 15\unit{Hz} time samples.  
This study builds a digital twin which is used to simulate the Booster data. 
Furthermore, to reliably predict and avoid large-scale major disruptions in nuclear fusion experiments, real-time analysis of the time-series data is crucial in guiding the action needed in experimental prediction and control.

\subsection{System constraints}
In this section, we present an overview of desired system properties and constraints that are prevalent across a number of application domains. 
Unique challenges are arising from each scientific application based on sensing technology, the physical processes, and the timescales and data rates, and bandwidth.  
These system constraints result in specific choices of data processing platforms, often with multiple compute architectures across the data continuum, such as the choice of FPGA-based systems versus embedded processors, GPUs, or custom ASICs. 
Table~\ref{applications} summarizes several scientific application domains along with their event rates, system latency constraints and performance requirements, and deployment characteristics. 
We broadly define platforms for integration fast machine learning techniques into ``soft'', software programmable coprocessors, and ``custom'', custom embedded computing devices.  
Software-programmable systems are often preferred because they are less complex to implement while custom embedded solutions are required when software programmable systems cannot satisfy experimental throughput, bandwidth, or latency constraints.  
We will describe in further detail this distinction below.  
%Table~\ref{data} provides further details and examples of data sources representing each data type. Note that system characteristics that are common to a broad class have been listed first in bold and those that are unique within an individual domain are listed subsequently. 
Examples of these system design choices are the trigger systems for HEP include LHC reconstruction of collision events, the Belle-II experiment, the Mu2e experiment which deploy custom embedded systems. 
Meanwhile, experiments like the Electron-Ion Collider have data rates that may not require custom hardware solutions and could deploy only software programmable solutions for event reconstruction and real-time processing experiments. %although they do not utilize the trigger concept. 
%Another broad category of domains deploy ML solutions to various image processing problems. One of the major distinctions from heavily studied computer vision-based problems is that many of the science applications aim to learn temporal dynamics of a system that is governed by physical processes. 
One final distinction worth discussing concerns the nature of real-time processing and the in-situ versus post-mortem nature of the inference and analysis tasks. 
Examples that we consider in classifying tasks that have different requirements are: data reduction which primarily focuses on limiting data collection rates of experiments for offline analysis; real-time processing and data analysis which is required to extract real-time domain features of the data for tasks like filtering/triggering; and closed-loop controls where data processing provides direct feedback to the operation and continuous control of an experiment. 
These distinctions and their consequences on the computing systems is illustrated in Table~\ref{timing}  

\begin{table*}
\centering
\caption{\label{applications}
Domains and practical constraints: systems are broadly classified as soft (software-programmable computing devices: CPUs, GPUs, and TPUs) and custom (custom embedded computing devices: FPGAs and ASICs)}
\small
\begin{tabular}{ c c c c c}
\hline
 Domain &  Event Rate & Latency & Systems &  Energy-constrained\\
        % &   Rate &          &         & Constrained\\
 \hline
 \hline
\textbf{Detection and Event Reconstruction} &     &  &         & \textbf{No}  \\ 
LHC \& intensity frontier HEP  &  10s Mhz & ns-ms & Soft/custom        &   \\ 
Nuclear physics &  10s kHz & ms   & soft & \\ 
Dark matter \& neutrino physics & 10s MHz & $\mu$s   & Soft/custom & \\ 
\hline
\textbf{Image Processing}    &         &          &       &   \\ 
Material synthesis        &  10s kHz   & ms       & Soft/custom & \\ 
Scanning probe microscopy &  kHz   & ms       & Soft/custom & \\ 
Electron microscopy       &  MHz   & $\mu$s       & Soft/custom & \\ 
Biomedical engineering   &  kHz   & ms       & Soft/custom & Yes (mobile settings)\\ 
Cosmology                &            Hz   & s        & soft           &  \\ 
Astrophysics             &            kHz--MHz   & ms-us  & Soft   & Yes (remote locations)  \\ 
\hline
\textbf{Signal Processing} &           &          &     &     \\ 
Gravitational waves        &           kHz      & ms      & Soft &  \\ 
Health monitoring          &  kHz      & ms      & Custom & Yes \\ 
Communications             &  kHz      & ms      & Soft & Yes (mobile settings)\\
\hline
\textbf{Control Systems}   &  &  &   \\ 
Accelerator controls       &         kHz & ms--$\mu$s & Soft/custom  \\ 
Plasma physics             &         kHz  & ms    & Soft  
\end{tabular}
\normalsize
\end{table*}

\begin{table}
\centering
\caption{\label{timing} Classification of domains and their system requirements with respect to real-time needs.}
\small
\begin{tabular}{ c c c c}
\hline
Domain &  Real-time data reduction & Real-time analysis & Closed-loop Control \\
 \hline
 \hline
\textbf{Detection/Event Reconstruction} &    &                              &   \\ 
LHC                                        & Yes & Yes & No \\ 
Nuclear Physics                            & Yes & No                           & No \\ 
Dark Matter - Neutrino                     & Yes  & No                          & No  \\ 
\hline
\textbf{Image Processing}    &   &              &   \\ 
Material Synthesis        & Yes       & Yes   & Yes \\ 
Scanning Probe Microscopy & Yes       &    &          \\ 
Electron Microscopy       & Yes       &    &      \\ 
Biomedical Engineering   & Yes        &    &      \\ 
Cosmology                & Yes        &  No   & No        \\ 
Astrophysics             & Yes       &  No   & No  \\ 
\hline
\textbf{Signal Processing} &  &          &          \\ 
Gravitational Waves        & Yes         &  No      & No      \\ 
Health Monitoring          & Yes & Yes   & Yes      \\ 
Communications             & Yes & Yes   & Yes      \\
\hline
\textbf{Control Systems}   &     &  &   \\ 
Accelerator Controls       & Yes & Yes & Yes  \\ 
Plasma Physics             & Yes & Yes & Yes      
\end{tabular}
\normalsize
\end{table}

\subsubsection{Software programmable coprocessors}
Historically, the first attempts at addressing the computational needs of the problems reviewed in this article have been through software-programmable systems. CPU-based local clusters or cloud services as well as cloud computing resources utilizing GPU or TPU-based hardware accelerators are utilized in different applications. One particular concept explored by the HEP community is the GPU as a Service (GPUaaS) model~\cite{krupa2020gpu}. 
This can further be expanded into the Machine Learning as a Service concept, similarly explored within HEP~\cite{kuznetsov2020mlaas4hep}. 
These paradigms involve the implementation of machine learning modules to solve a set of physics problems, which are then transferred to GPU or TPU accelerators and accessed by the local CPU ``client'' of the native experimental system. 

One of the major system constraints is the computational capacity, which can be defined in terms of a number of floating point operations as far as neural network implementations are concerned. 
Real-time machine learning methods require an ever-increasing rate of computational capacity as it directly impacts the \textit{latency per task}. 
The \textit{task} could be a trigger for LHC, reconstruction of an event in accelerator experiments or astrophysics, material synthesis, reconstruction of an image captured by an electron microscope, etc. 
Extreme parallelism would be desired to provide the highest capacity possible to minimize latency and maximize throughput. 
In a processor-based system, this can be addressed by increasing the size of the compute cluster. 
Naturally, facility costs impose a limit on the scale of these clusters. 
Another constraint is the available amount of storage coupled with the cost of data movement across the memory hierarchy. 
In the majority of the use cases, the latency involved with moving data from the front-end (detectors, microscopes, sensors, etc.) dominates the total latency. 
One of the prominent performance constraints is related to the utilization and subsequent latency of the network that links the front-end with the back-end. Current limitations on the speed of data movement renders the CPU/GPU cluster-based systems unable to meet the real-time requirements.               
\subsubsection{Custom embedded computing devices}
As the latency and throughput constraints are coupled with challenging practical energy constraints, efforts have been directed towards specialized computing systems to address the hard real-time needs. 
An increasingly attractive paradigm is to design components that are finely optimized for specific steps in the data capture workflow. 
These components can be mapped onto FPGA devices or they can be designed and manufactured as an application-specific integrated circuit (ASIC). 
In the LHC and accelerator domains, there is a rich set of FPGA-based demonstrations of front-end data processing systems, which meet microsecond latencies. 
These systems are in charge of tasks such as triggering, event reconstruction, and anomaly detection. Direct and naive implementations of neural networks to perform inference for these tasks can fail to meet the latency requirements since they often incur significant resource utilization. 
The highest achievable FPGA clock frequency and inference latency is correlated with the resource utilization and percentage occupancy of the device. 
Co-design techniques developed for these applications particularly specialize in extreme quantization and pruning (with an awareness of accuracy) so that resource requirements can be controlled aggressively to ensure inference latency targets. 
These optimizations push the resource usage envelope as far as down as 10s of percent of the FPGA device in order to meet the system constraints and yet demonstrate implementations with high inference accuracy.

Some other applications (e.g., accelerator controls, biomedical and health applications) impose less stringent latency expectations, in the order of ms, where the urgency for resource minimization is alleviated. 
Hence, the focus of the system design can shift from extreme resource economy to enhanced sophistication in the algorithms that are being mapped to the device. 
Inference models can now include deep(er) learning models coupled with advanced video and signal processing engines, as well as local privacy-preserving processing tasks (applicable particularly to mobile health and networking and communication applications). 

For mobile and IoT-based deployment of the edge devices, resource efficiency emerges as an important factor as it impacts energy consumption.
However, in these applications, energy efficiency can also be achieved by alternative means. 
One option would be selective powering, i.e., creating a resource-rich full-featured baseline implementation, which still comfortably meets latency constraints if energy was not an issue, and introducing power gating or standby features to modulate energy consumption during periods of low/no activity. 

There are system constraints, which point the designers to a custom ASIC solution in addition to or in place of FPGA devices. ASICs can address extreme form factor considerations, integration of computation with sensing (e.g., smart photon detectors) into compact front-end devices, tight integration with other mixed-signal or analog functionalities, radiation hardening requirements, and ultra-low energy budgets.

%%%%%%%%%%%%%%%%%%%%%%%%%%%%%%%%%%%%
\pagebreak
% \addtocontents{toc}{\setcounter{tocdepth}{5}}
\addtocontents{toc}{\protect\setcounter{tocdepth}{5}}
\section{Technology State-of-the-Art}
\label{sec:technolog_sota}
% \noindent 
% \textcolor{blue}{
% The content of this section will be dependent on the interests of contributors and what they want to write about.
% }
In this section, we aim to give an overview of technologies and techniques for building fast ML algorithms. 
This requires {\it codesign}: building algorithms with hardware in mind and providing efficient platforms for programming the hardware.  Section~\ref{sec:deploy} and Section~\ref{sec:train} focus on neural network design and training for efficient implementation in hardware. 
In Section~\ref{sec:cmos} and Section~\ref{sec:beyondcmos}, we classify our discussion of ML hardware compute platforms into two categories: ``Conventional CMOS Hardware'' and ``Emerging Beyond CMOS Hardware.''  
The former will address nearer-term hardware solutions, while the latter will focus on the speculative end of the spectrum. Meanwhile, because the area of programming new hardware is rapidly moving, we lay out an example of the options and challenges for one device family: FPGAs. 
This is presented in Sec.~\ref{sec:codesign}, and from the details for FPGAs we hope the reader also gets a sense of the fundamental approaches for designing software for emerging hardware.

\subsection{Systematic Methods for the Efficient Deployment of ML Models}
\label{sec:deploy}

As discussed in Section~\ref{sec:apps}, many ML problems in science
require low latency, often with
constrained resources. However, most of the current state-of-the-art NN models have prohibitively high latency with a large
memory footprint and energy consumption. 
For this reason, practitioners have been forced to use sub-optimal models (e.g. shallow NNs)
with non-ideal accuracy to avoid this latency problem.
There is a large body of literature that has
focused on solving this problem by making NN models more efficient
(in terms of latency, memory footprint, and energy consumption). These efforts could be broadly categorized as follows:
(i) Designing new efficient NN architectures;
(ii) NN and hardware co-design;
(iii) Quantization (low precision inference);
(iv) Pruning and sparse inference; and
(v) Knowledge distillation.
Here we briefly discuss each of these approaches.

% --------------------------------------------------------------------------------
\paragraph*{Designing new efficient NN architectures}
One line of research has been focused on finding new NN models
that are efficient by design. A notable early work is SqueezeNet~\cite{iandola2016squeezenet},
a new NN model without any expensive Fully Connected layers,
along with a new lightweight \emph{Fire module}, that resulted
in a $50\times$ smaller model as compared to AlexNet, but with the
same accuracy. Later on, several new innovations were made in
efficient NN architecture design.
One focus has been to find efficient layers/operators.
Notable works are
group convolutions~\cite{ioannou2017deep}, 
depthwise convolutions~\cite{howard2017mobilenets},
spatial separable convolutions~\cite{mamalet2012simplifying},
shuffle layers~\cite{ma2018shufflenet},
and shift convolutions~\cite{wu2018shift}, to name a few.

Another focus has been to find similar substitutes to \emph{Fire module} that are more efficient and result in better
accuracy/generalization. Notable works include
residual networks~\cite{he2016deep} (originally designed
to solve issues with vanishing gradients, but these structures are generally more efficient
than non-residual architectures), densely connected networks~\cite{huang2017densely},
squeeze-and-excite modules~\cite{hu2018squeeze},
and inverted residual blocks~\cite{sandler2018mobilenetv2}.

These classical techniques mostly found new architecture modules
through a manual design search. 
This is not scalable, and as such recent approaches have proposed automated methods that use neural architecture search (NAS). 
NAS methods automatically find the right NN architecture for a given constraint
of model size, depth/width, and/or latency.
The high-level approach here is to train a probabilistic \emph{SuperNet}
that includes all possible combinations of NN architectures within the prescribed
constraints, but with learnable probabilities. 
After this SuperNet is trained, one can sample an architecture from its learned probability distribution.
Notable works include RL based methods~\cite{zoph2016neural}, efficient NAS~\cite{pham2018efficient}, MNasNet~\cite{tan2019mnasnet}, DARTS~\cite{liu2018darts}, and Differentiable NAS~\cite{wu2019fbnet}.

% \begin{outline}
%     \1 SqueezeNet ~\cite{iandola2016squeezenet};
%     \1 MobileNet family~\cite{howard2017mobilenets,sandler2018mobilenetv2,howard2019searchin}
%     \1 ShuffleNet~\cite{ma2018shufflenet}, ShiftNet~\cite{wu2018shift}, DenseNet~\cite{huang2017densely}
%     \1 EfficientNet~\cite{tan2019efficientnet}
%     \1 Discuss the methodology of AutoML/DNAS and how it works
% \end{outline}

% --------------------------------------------------------------------------------
\paragraph*{NN and hardware co-design}
Another promising line of work has been to tailor the NN architecture for a specific hardware
platform, and/or co-design them together. This is quite promising for
configurable hardware such as FPGAs. The importance of hardware-aware NN design is that
the cost of performing different types of operations varies for different hardware.
For example, hardware that has a dedicated cache hierarchy can execute bandwidth
bound operations much more efficiently than hardware without a cache hierarchy.
Notable works in this area include SqueezeNext~\cite{gholami2018squeezenext}, where both the NN and the hardware
accelerator were co-designed with a manual tuning approach.
More recent works have proposed to automate hardware-aware design
through NAS.
Notable works include ProxylessNAS~\cite{cai2018proxylessnas}, OnceForAll~\cite{cai2019once}, FBNet~\cite{wu2019fbnet},
and MobileNetV3~\cite{howard2019searching}.

% --------------------------------------------------------------------------------
\paragraph*{Quantization (low precision inference)}
A common solution is to compress NN models with quantization~\cite{asanovic1991experimental, hubara2016binarized, rastegari2016xnor, zhou2017incremental, zhou2016dorefa, cai2017deep, jacob2018quantization, zhang2018lq, choi2018pact, wang2019haq, dong2019hawq, chin2020one, cai2020zeroq, gholami2021survey},
where low bit-precision is used for weights/activations.
A notable work here is Deep Compression~\cite{han2015deep}, which used
quantization to compress the model footprint of the
SqueezeNet model discussed above, bringing its size to 500x smaller than AlexNet. 
In quantization, the model size is reduced without changing the original network architecture,
and it could potentially permit the use of low-precision matrix multiplication or convolution. Therefore, both the memory footprint
and the latency could be improved.

The quantization methods can be broadly classified into two categories of \emph{Post-Training Quantization} (PTQ),
and \emph{Quantization-Aware Training} (QAT). In PTQ, a pre-trained model in single precision is quantized to low precision
without any fine-tuning or re-training~\cite{banner2018post,meller2019same,choukroun2019low,zhao2019improving,fang2020post,fang2020near,lee2018quantization,nagel2019data,cai2020zeroq,hawks2021ps}. As such, these quantization methods are typically very fast, and, in some cases, do not even require any training data~\cite{cai2020zeroq, haroush2020knowledge,nagel2019data}.
However, PTQ often leads to high accuracy degradation, especially for low precision quantization.
To address this, some quantization methods adopt QAT to re-train the model after the quantization, so that the parameters can
get adjusted. 
This approach often results in higher accuracy, but at the cost of longer
time associated with re-training the model~\cite{courbariaux2015binaryconnect,lin2015neural,hubara2016binarized,rastegari2016xnor,zhou2016dorefa,zhu2016trained,cai2017deep,hou2016loss,gysel2018ristretto,huang2021codenet,zhou2018explicit}.

Another differentiator is the use of \emph{simulated quantization} (aka fake quantization), versus \emph{integer-only} quantization~\cite{jacob2018quantization,yao2020hawqv3,kim2021bert,lin2016fixed}. In the former, 
the weights/activations are stored in low precision, but they are cast to higher precision during
inference. In the latter, there is no casting involved, and the multiplication and accumulation also happen
in low precision. Using integer-only quantization has the advantage that one can speed up
inference by using low-precision logic for multiplication and addition, besides reducing the memory footprint of the model.

% For some cases, it is possible to map a simulated quantization graph to low precision
% multipliers. However, this could lead to discrepancy which can become large, especially for low precision quantization~\cite{yao2020hawqv3}.

\begin{figure*}
\centering
\includegraphics[width=0.8\linewidth]{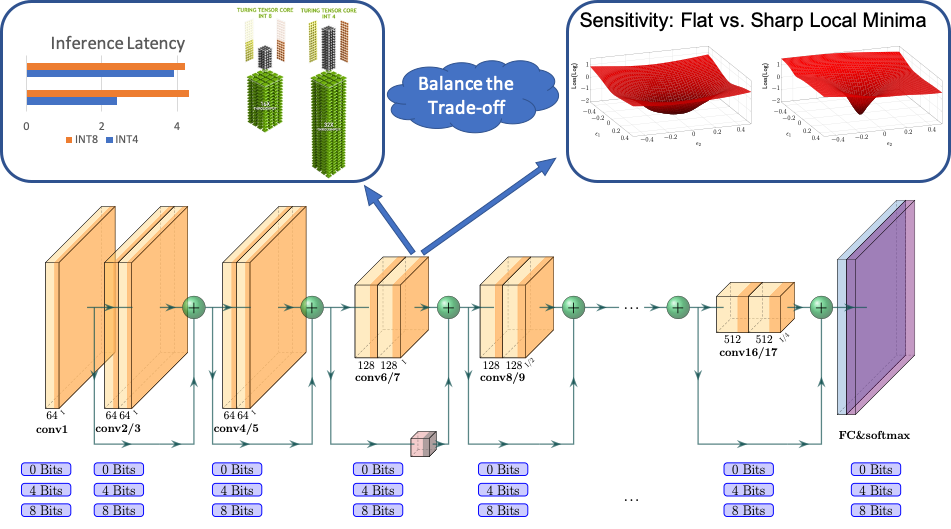}
\caption{The illustration of hardware-aware quantization and pruning. 
A given NN model can be compressed by using low precision quantization instead
of single precision. The extreme case is to use 0-bit quantization which is 
equivalent to removing/pruning the corresponding neurons.
The goal of compression is to find the best bit-precision setting for
quantization/pruning to reduce model footprint/latency on a target hardware
with minimal generalization loss.
}
\label{fig:pruning_quantization}
\end{figure*}

Another distinction is \emph{hardware-aware quantization}. 
Similar to NN architecture design, quantization can also be tailored for specific hardware platforms.
This becomes important for mixed-precision quantization~\cite{zhou2018adaptive, wang2018haq, wu2018mixed, hawq, shen2020q, hawqv2, dong2021hao, yao2020hawqv3}.
The reason is that certain operations in the NN model may benefit more from low precision
quantization than others, based on whether they are bandwidth bound or compute-bound.
As such, as schematically
illustrated in Figure~\ref{fig:pruning_quantization},
one must determine the best precision setting based on the tradeoff between the potential footprint/latency gain and the sensitivity to accuracy degradation.

% ----------------------------------------------------------
% ----------------------------------------------------------

% --------------------------------------------------------------------------------
\paragraph*{Pruning and sparse inference}

Another approach reducing the memory footprint and computational cost of NNs is to apply
pruning, which could be thought of as quantization to 0-bits. In pruning, neurons with small \emph{saliency} (sensitivity) are removed, which results in a sparse computational graph~\cite{lecun1990optimal}. 
Here, neurons with small saliency are those whose removal should minimally affect the model output/loss function.
Pruning methods can be broadly categorized into unstructured pruning~\cite{lecun1990optimal,hassibi1993second,dong2017learning, lee2018snip, xiao2019autoprune, park2020lookahead}, and structured pruning~\cite{luo2017thinet, he2018amc, yu2018nisp, lin2018accelerating, huang2018data, zhao2019variational}.
Unstructured pruning removes neurons without any structure.
With this approach, one can remove most of the NN parameters with little impact on the
generalization performance of the model.
However, this approach leads to sparse matrix operations which are hard to accelerate 
and are typically memory-bounded~\cite{buluc2008challenges,gale2019state,hoefler2021sparsity,blalock2020state}.
This can be addressed with structured pruning, where a
group of parameters (e.g., an output channel) is removed. However, the challenge here is that high degrees of structured
pruning often lead to significant accuracy degradation.

In both approaches, the key question is to find which parameters to prune.
A simple and popular approach is magnitude-based pruning~\cite{chauvin1989back,hanson1988comparing,mozer1988skeletonization,li2016pruning,he2017channel,liu2017learning,he2019filter,lin2020hrank}.
In this approach, the magnitude of parameters is used as the pruning metric.
The assumption here is that small parameters are not important and can be removed.

An important problem with magnitude-based pruning methods is that parameters with small magnitudes can actually be quite sensitive.
It is easy to see this through a second-order Taylor series expansion, where the perturbation is dependent on not just the weight magnitude but
also the Hessian~\cite{lecun1990optimal}. As such there are several works that use
second-order based pruning~\cite{lecun1990optimal,hassibi1993optimal,hassibi1993second,wang2019eigendamage,yu2021hessian}.

Finally, we should mention that it is possible to 
combine pruning and quantization together to compress the NN model.
In fact, pruning could be viewed as quantization to 0-bits. The recent work
of ~\cite{hawks2021ps} proposes a quantization-aware pruning
method and applies to high energy physics problems;  It reports
better results than pruning or quantization alone.

% \begin{outline}
% \1 \textbf{Pruning}~\cite{han2016deep, hassibi1993second, optimalbraindamage, dong2017learning} removes unnecessary nodes/edges from the model for inference. These edges 
%     may be important for training the model, but can be removed without hurting
%     generalization for inference.
%     \2 Current challenges:
%         \3 Pruning beyond 50\% typically results in significant accuracy degradation
%         \3 Need to explore coupled pruning and quantization
%         \3 Unstructured pruning has high generalization but inefficient for HW deployment. Need to study new methods for unstructured pruning that can be deployed efficiently.
% \end{outline}
% --------------------------------------------------------------------------------

\paragraph*{Knowledge distillation}
Model distillation~\cite{romero2014fitnets, hinton2015distilling, mishra2017apprentice, li2017learning, yim2017gift, polino2018model, ahn2019variational, yin2020dreaming} trains a large model and then uses it as a teacher to train a compact model. Instead of using class labels during the training of the student model, the key idea of model distillation is to leverage the soft probabilities produced by the teacher, 
which can guide/help the student training.
% As an example in crystal structure recognition, for an input lattice with the label \textit{Fluorite structure}, classes such as \textit{Antifluorite structure} can also have high probabilities according to the teacher model. Teaching the compact student model to distinguish \textit{Antifluorite structure} (in addition to \textit{Fluorite structure}) from other irrelevant classes can lead to potential performance improvement.

Previous methods of knowledge distillation focus on exploring different knowledge sources. Refs.~\cite{hinton2015distilling, li2017learning, park2019relational} use logits (the soft probabilities) as the source of knowledge, while Refs.~\cite{romero2014fitnets, yim2017gift, ahn2019variational} try to leverage the knowledge from intermediate layers. The choices of teacher models are also well studied, where Refs.~\cite{you2017learning, tarvainen2017mean} use multiple teacher models to jointly supervise the student model, while Refs.~\cite{crowley2018moonshine, zhang2019your} apply self-distillation without an extra teacher model. Other previous efforts apply knowledge distillation with different settings on different applications. Refs.~\cite{lopes2017data, nayak2019zero, yin2020dreaming} study data-free knowledge distillation, and Refs.~\cite{wang2018kdgan, wang2020minegan} combine knowledge distillation with GANs.

A major challenge of knowledge distillation methods is to achieve a high compression ratio. 
Compared to quantization and pruning which can usually maintain accuracy at $4\times$ compression, knowledge distillation methods tend to have non-negligible accuracy degradation at those compression levels. 
But these two approaches are orthogonal, and recent works have
shown that their combination can result in high accuracy/compression~\cite{polino2018model,mishra2017apprentice,yao2020hawqv3,mao2020ladabert}.
It should be mentioned that current distillation methods are mostly applied to classical ML problems, and few works have looked into their application in Science AI problems.

% ------------------------------------------------------------
% ------------------------------------------------------------
% ------------------------------------------------------------
% ------------------------------------------------------------
\subsection{Systematic Neural Network Design and Training}
\label{sec:train}

There is currently no analytical approach to find the right NN architecture for
a given task and training dataset.
Originally, designing the NN architecture was mostly a manual task with
intuitions that were often ad-hoc. However, in recent years there has been
a lot of innovations in automating the NN architecture design process, which is 
referred to as Neural Architecture Search~\cite{zoph2016neural,pham2018efficient,tan2019mnasnet,liu2018darts,wu2019fbnet,cai2018proxylessnas,cai2019once}.

NAS could be viewed as a hyperparameter tuning problem, where the hyperparameters are
the design choices for a NN architecture. This could include
width, depth, types of operations, etc. The main challenge
is that the search space for the operation types scales exponentially with the number
of layers.
As such, one has to still include some high-level intuition about the NN architecture
to limit the search space. 

After limiting the search space, the general NAS process is as follows:
A candidate architecture is sampled from the set of all
possible architectures and is then trained for a number of epochs on the training dataset.
The accuracy is then used 
as the metric to evaluate how good that candidate architecture is. Then based
on this reward, the probability distribution of sampling architectures is updated.
This process needs to be repeated for many different candidate architectures (sometimes exceeding
hundreds of thousands). 
Inherently, this leads to another problem related to tuning the optimization hyper-parameters
for each candidate architecture.
For example, if a good architecture is sampled from the NAS but
is trained with sub-optimal hyperparamters, then the error will be high and the NAS algorithm will reduce the likelihood
of sampling that architecture which is not the desired property.

% \textcolor{red}{[S.H. I don't understand this statement]
% AG: What we meant here is that even if NAS sampled a good architecture but you trained it with bad hyperparamters you would get bad errors and the NAS algorithm will reduce the likelihood of sampling that architecture which is not what we want.

As a result, \emph{scalability} has become an integral concern for any procedure in the presence of ``big data.'' 
One main class of procedures for which scalability has become indispensable is in
numerical optimization algorithms, which are the core of training methods. 
There is a large body of literature on designing
efficient  numerical optimization/training methods~\cite{reddi2018adaptive,shazeer2018adafactor,zhang2019lookahead,park2020lookahead,zhuang2020adabelief,liu2020variance,ginsburg2020stochastic,yao2020adahessian,ma2020apollo,gupta2018shampoo} as well as efficient NAS algorithms to 
search for the right NN architecture~\cite{zoph2016neural,pham2018efficient,tan2019mnasnet,liu2018darts,wu2019fbnet}.

For the optimization, the goal is to design new methods that require fewer iterations to converge and are
more robust to hyper-parameter tuning.
One notable advancement here is the ability to apply second-order methods
without the need for forming the second-order operator~\cite{yao2020adahessian,yao2019pyhessian,gupta2018shampoo,reddi2018adaptive}.
It has been shown that the performance and robustness of these methods are higher than
first-order optimization methods on classical ML problems (e.g. in computer vision or natural language processing).
Interestingly, some recent results for Physics Informed Neural Networks (PINN)~\cite{raissi2019physics} have
found that first-order methods work significantly sub-par to (quasi) second-order methods.
This could potentially provide opportunities to adapt or redesign some of the second-order algorithms for Science problems.

For the NAS algorithms, the goal is similar, which is to find methods
that require evaluating fewer candidate architectures, with less manual restriction or tuning
of the search space.
Another goal is to design transferable NAS algorithms that can be trained on a small problem
and then transferred to larger problems that are more expensive~\cite{cai2018proxylessnas,cai2019once}.

In summary, the core of designing NN architecture is to have a fast method of sampling architectures (through NAS),
and the fast training of the sampled architectures (through fast and robust optimization algorithms).

% Challenges with existing methods for training and designing NNs:

% \begin{outline}
% \1 Mostly empirical with trial and error $=>$ brute-force hyperparameter tuning
%     \2 Need to design systematic neural architecture search methods $=>$ This is basically a monte-carlo search method, need to borrow ideas from MCMC community.
% \1 For science applications we need to impose physics constraints, such as fundamental laws of physics such as conservation of mass and energy.
%     \2 Need to impose this during the NAS search above. One approach is to hard-code this, another is to add a penalty term as done in PINNs.
% \1 Lack of scalable training methods. As a result training time and designing NNs is very costly
%     \2  Hard to scale training to supercomputers, due to the inherent problems with large batch SGD training
%     \2 Need to look beyond SGD, and into second order methods

% \1 Not interpretable. 
%     \2  Not clear why a particular depth/width has the best accuracy, and if accuracy obtained on the test dataset is due to overfitting to non-robust features or if it has actually learned to extract meaningful features fro the data.
% \1 Current methods do not provide uncertainty quantification.
%     \2 How would have the results been changed if there was noise in the input data?

% \end{outline}

%%%%%%%%%%%%%%%%%%%%%%%%%%%%%%%%%%%%%%%%%%%%%%%%%%%%%%%%%%%%
%%%%%%%%%%%%%%%%%%%%%%%%%%%%%%%%%%%%%%%%%%%%%%%%%%%%%%%%%%%%
%%%%%%%%%%%%%%%%%%%%%%%%%%%%%%%%%%%%%%%%%%%%%%%%%%%%%%%%%%%%

\subsection{Hardware Architectures: Conventional CMOS}
\label{sec:cmos}
%In the previous section, we considered the application requirements for deployment of \acp{cnn} and quantified their associated compute and memory demands, which are huge and growing beyond the limits to where standard silicon-based semiconductors can scale. 
As the prevalence and demands for machine learning rapidly continue to grow, it is increasingly important that we design machine learning algorithms efficiently and simultaneously deploy them on complementary and powerful hardware platforms. The compute and memory demands of NN deployments are huge and growing beyond the limits to where standard silicon-based semiconductors can scale.  The reasons behind the scalability challenges in the semiconductor industry are as follows:
Firstly, as we approach the End of Moore's Law, transistor cost has been exponentially rising due to rising chip design costs with shrinking technology nodes (as published by Xilinx and Gartner in 2011 already~\cite{trimberger2018three}). Furthermore, with the end of Dennard scaling, we've encountered considerable thermal challenges as power density no longer remains constant between node generations. To mitigate the challenges of increasing thermal density, chips are now designed to conditionally deliver power to groups of transistors, effectively throttling or "turning off" parts of a chip. This technique has come to be known as creating dark silicon ~\cite{esmaeilzadeh2011dark}.

To overcome these challenges and provide sufficient compute capabilities, many disruptive approaches have been proposed. For example, Cerebras Systems~\cite{cerebras} has brought to market the first computer system which employs \textbf{wafer scale integration}.  where chips are built from complete wafers rather than individual dies. Such a technique brought with it substantial engineering challenges in regards to power delivery, packaging, and cooling. Exploring the other dimension, foundries are investigating true \textbf{3D chip stacking} as was presented at HotChips'2019 by TSMC~\cite{TSMC}. Even \textbf{analog computing}~\cite{aspinity,yuzuguler2019analog}, \textbf{quantum computing}~\cite{dwave} and \textbf{in-memory computing}~\cite{neuroblade,essera2016convolutional} are investigated as well. %All of these are on the speculative end of the spectrum. 

% \begin{figure}[h!]
% \centering
% \includegraphics[width=0.7\linewidth]{figures/cloud.png}
% \caption{Innovative approaches for acceleration of CNN workloads}
% \label{fig:cloud}
% \end{figure}

% \subsubsection{Conventional CMOS hardware}

Less risky approaches focus on moving away from traditional von Neumann architectures, using specialization of compute architectures to provide the necessary performance scaling and energy efficiency. 
Due to the specialization, the devices become increasingly heterogeneous.
A huge range of devices has emerged that all try to address this problem in different ways, whereby the key challenge is:
How do we loop transform and unfold the algorithms best to maximize data reuse and compute efficiency, minimize memory bottlenecks, and limit power consumption while meeting real-time requirements? 
%In the following, we restrict our discussion to the devices that can be realistically used today, and thereby focus on \textbf{specialized architectures}. 
%\textcolor{red}{[S.H. This statement seems out of place.  What is "speculative approaches" referring to?]It is important to remember though that these speculative approaches, as discussed above, will materialize in the future and we will have to deal with an even greater diversity of hardware platforms.}

The choice of hardware type and quantity often boils down to a set of constraints imposed by compute environment (datacenter, cloud, on-premise, edge, mobile), workload type (inference, training), data type (Language, Time Series, Vision, Graph, etc), ML model, usage model (online inference, batch jobs), and user-centric Service-Level Agreements (encryption level, request latency, etc). For large datacenter deployments handling various types of workloads, it is often the case that several platforms must be combined to reduce Total Cost of Ownership (ToC) across all their hardware platforms. It has therefore become increasingly necessary for owners of heterogeneous platforms to think of their systems as large-scale multi-processor computers, a trend sometimes termed Warehouse Scale Computing~\cite{wsc}. For Deep Learning hardware accelerators, these new computers generally take the form of CPU co-processors: a host CPU communicates with other entities in the datacenter, interfaces with disk memory, and formats input data which is then offloaded to the accelerator responsible for executing a user-defined compute graph, or Neural Network. 

We begin with a taxonomy of these hardware architectures and discuss their relevant characteristics when it comes to the acceleration of machine learning workloads. 
This is essential to understand how they will differ in their execution behavior, what it takes to leverage their unique features and how they can potentially benefit from previously introduced optimization techniques. 
% We will also discuss deployment options that are unique to specific architectures, and may or may not bring additional benefits to CNN inference. 
% At last, we briefly touch on other considerations such as form factors. All of these stipulate another form of design requirements for a benchmarking methodology.

%%%%%%%%%%%%%%%%%%%%%%%%%%%%%%%%%%%%%%%%%%%%%%%%%%%%%%%%%%%%
\subsubsection*{\textbf{Taxonomy of Compute Architectures for Deep Learning}}

\begin{figure*}
\centering
\includegraphics[width=0.8\linewidth]{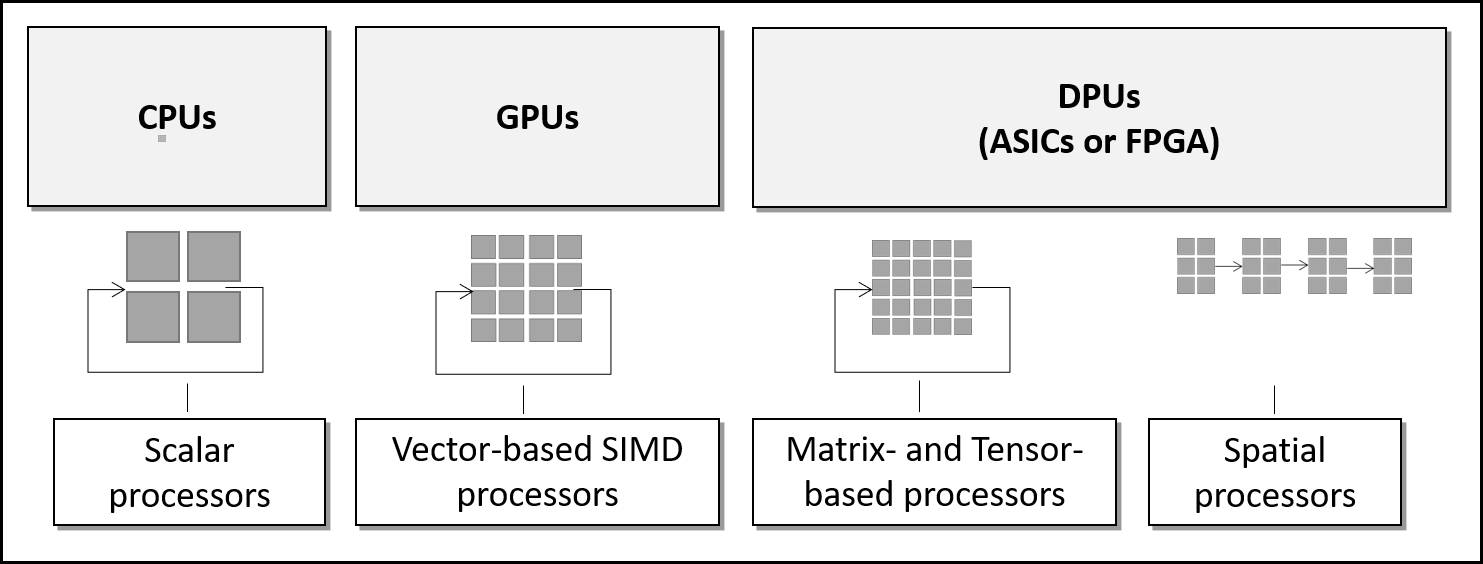}
\caption{Taxonomy of compute architectures, differentiating CPUs, GPUs and DPUs}
\label{fig:tax}
\end{figure*}

A broad range of hardware architectures to deploy machine learning algorithms exists today.
We can broadly classify them by the following criteria:

\begin{enumerate}[itemsep=-1pt]
\item Basic type of compute operation
\item Inherent support for specific numerical representations
\item External memory capacity (which is mostly relevant for training workloads)~\footnote{In these comparisons, we treat HBM and HBM2 as external memory as it is used in the same way as DDR4 or GDDR memory.}
\item External memory access bandwidth
\item Power consumption in the form of thermal design power (TDP)
\item Level of parallelism in the architecture and the degree of specialization
\end{enumerate}
 
As is shown in Figure~\ref{fig:tax}, we classify the compute architectures into scalar processors (\textbf{CPUs}), vector-based processors (\textbf{GPUs}), and so-called deep learning processing units (\textbf{DPUs}), although realistically these categories blend to some degree. 
DPUs are specialized for this application domain whereby we distinguish the more generic matrix- or tensor-based processor and a spatial processing approach. DPUs can be implemented with either ASICs or FPGAs. All of these architectures will be discussed individually below.

\paragraph*{CPUs} CPUs are widely used for ML applications and are viewed as largely serial or scalar compute engines (even though high-end variants for cloud deployment may have up to 10s of cores). They are optimized for single-thread performance, with implicitly managed memory hierarchies (with multiple levels of caches), and support floating point operations (FP64 and FP32) as well as 8bit and 16bit integer formats with dedicated vector units in most recent variants. Theoretical peak performance tops at 6.8TOPs for FP64 assuming boost clock speed (Cascade lake, 56 cores, 3.8GHz). External memory is currently primarily leveraging DDR4 memory banks with large capacities: Intel's Cascade Lake offers up to 4.5 TebiByte ($2^{40}$ Bytes) which is beyond what any of the other device categories can offer. Access is at maximum speed through high-end hardened memory controllers, offering 282~Gbps bandwidth (for example Cascade Lake with 12 DDR4 channels). 
Compared to GPUs and other HBM-enabled devices, the memory bandwidth of CPUs is lower. However, for many use cases, this can be compensated through their sophisticated cache hierarchies, combined with mature compiler tools.
%\textcolor{red}{[S.H. This sentence is confusing and should be rewritten]In regards to memory bandwidth, this is overall at the lower end of the spectrum, however in many application contexts, this can be compensated through the sophisticated multi-level memory hierarchies.}
Regarding power consumption, CPUs are at the upper end of the spectrum with high-end devices range up to 400\,W~\cite{cascadelake}.
In the embedded space, ARM processors provide generally popular solutions, in particular when performance requirements are very low and when functionality is required that is not supported by the specialized device variants. In particular, the 
%\textcolor{red}{[S.H. Missing reference]
Ethos~\cite{skillman2020technical} family of processing cores is specialized for CNN workloads and as such is considered under the DPU category below.
The advantages of CPUs are the generality of the hardware, as well as the ease of programming where design environments have matured over decades. 
As expected this comes at the cost of lower peak performance and less efficiency compared to the more specialized device families.
In regards to quantization, CPUs can only leverage this optimization technique for INT8 and INT16 if supported.

\paragraph*{GPUs} GPUs are SIMD-based (Single Instruction, Multiple Data) vector processors that support smaller floating point formats (FP16) natively, as well as fixed point 8-bit and 4-bit integer formats more recently, and have a mix of implicitly and explicitly managed memory. 
NVIDIA GPUs are some of the most popular hardware targets for machine learning, and newer families of chips have been introduced to specifically accelerate this workload, with AMD not far behind. 
The latest devices in NVIDIA's Volta and Turing architecture families, introduced in 2018 and 2019 respectively, offer up 130TOPs in FP16, which is beyond the capabilities of the latest CPU generations. 
As such they are amongst the highest performant devices in the market for the acceleration of DNNs as they can exploit the high degree of parallelism inherent in this application via increasingly specialized architectural features.
For example, NVIDIA's Volta is the first generation to incorporate tensor cores as a new feature, as well as improved FP32 and FP64 support for training in a data center setting~\cite{NVIDIAv100}, and also introduced a deep learning accelerator (DLA) in their embedded devices to further reduce power consumption. This specialization brings additional challenges for their usage; there are up to 3 distinct execution units now, namely CUDA cores, tensor cores, and the DLA, which don't operate on the workload simultaneously (at least not easily or by default). 
We, therefore, don't sum up the peak performance of different execution units, but use only the maximum.
AMD announced the Vega GPU~\cite{RadeonInstinctGPU} with new deep learning instruction set operations, with the goal of obtaining parity with NVIDIA's high-end Tesla V100 datacenter GPUs.  
Also, AMD's  most recent EPYC family supports customized instructions for deep learning~\cite{epyc}. 
Both companies offer also low power GPUs for the embedded space, namely the AMD Vega mobile GPU~\cite{radeon-mobile} and NVIDIA Jetson TX2~\cite{nvidia-jetson} and AGX family~\cite{agx}.

In regards to memory, GPUs leverage specialized and highly pipelined GDDR memory, which reduces capacity, but offers much higher bandwidth (up to 732GBps). With NVIDIA's Turing family the latest devices include HBM2 DDR memory stacks~\cite{turing}, which scales the memory access bandwidth to 1TBps and beyond. 
Again this is particularly important to address the needs of training workloads.
For the same reason, some of the DPUs introduce HBM2 as well, as discussed below. 
In regards to power consumption, GPUs are high, up to 345\,W.

One general challenge for GPUs is that they need to leverage input parallelism to achieve high utilization of their large compute arrays. Therefore before execution inputs need to be grouped into batches, which has adverse effects on end latency. 
% This is discussed in more detail below in Section~\ref{sec:deploy}. 
Further, GPUs are relatively high in power consumption.
Regarding quantization, support is limited to the inherent datatypes, which are INT4 at smallest in the context of NVIDIA's Turing family, and INT8 for many of the others.
Finally, the corresponding software environments for GPUs, while not on the same level as CPUs, have matured significantly and provide increased ease of use.

\paragraph*{FPGAs and ASICs}
FPGA and ASIC customize hardware architectures to the specifics of a given application.
They can be adapted in all aspects to suit a use case's specific requirements.
This includes their IO capability, their functionality, or even to suit specific performance or efficiency targets. 
FPGAs can be reprogrammed whereas ASICs are fully hardened.
This flexibility allows for amortizing the design costs of the circuit across many applications but comes at the expense of hardware resource cost and performance.

FPGAs are a popular choice for the acceleration of CNNs. 
Traditionally, an FPGA compute fabric consist of a sea of lookup tables (LUTs) which are interconnected through a programmable interconnect. The latest generations host millions of LUTs. Furthermore, the fabric is interspersed with specialized hardened compute blocks (DSPs) which accelerate n-bit multiply accumulate operations (MACs), as well as SRAM blocks. The latter are referred to as block RAMs (BRAMs), which hold 36~kbits, and Ultra RAMs (URAMs) which store 288~kbits.
More recent FPGA generations combine multiple FPGA dies, referred to as super logic regions (SLRs), and leverage a silicon interposer to provide connectivity between SLRs. 
This technology is referred to as stacked silicon interconnect (SSIT) and helps scale device capacity.

\paragraph*{DPUs} As mentioned at the beginning, the term DPU (short for deep learning processing unit) refers to a new type of compute architecture, specialized for the acceleration of CNNs. 
DPUs are customized for these types of applications in a number of ways: types of operations supported, direct support of tensors or matrices, inherent data types and supported numerical representations, macro-architecture, explicitly managed and specialized memory hierarchies, and which levels of parallelism they exploit (input, output pixel, IFM, OFM, bit, and layer and branch parallelism) as was introduced in the first part of this chapter.
We differentiate two types of DPUs, which can be implemented with both ASIC technology and FPGAs.

\begin{figure*}[h!]
\centering
\includegraphics[width=0.8\linewidth]{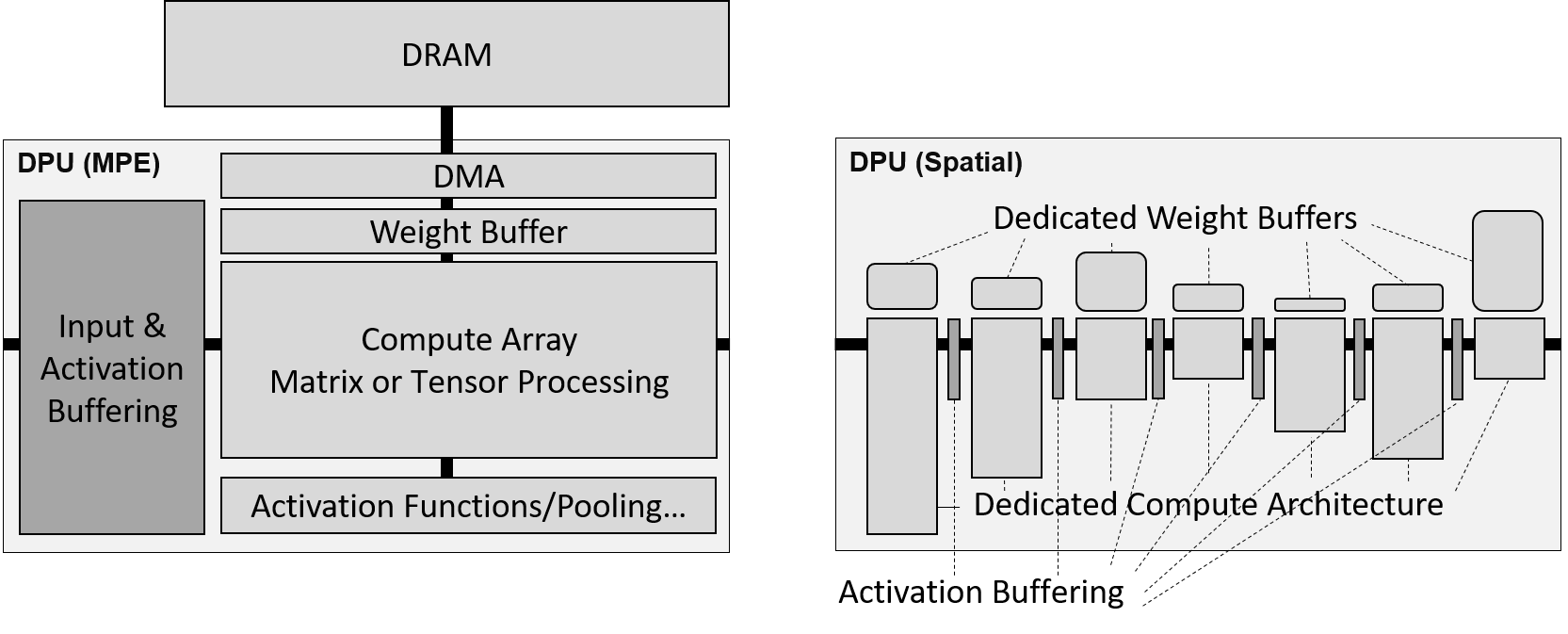}
\caption{DPU architectures: Matrix of Processing Engines (MPE) on the left, and spatial architecture on the right}
\label{fig:dpu}
\end{figure*}

\paragraph*{Matrix of Processing Elements (MPE)} The first type, as shown on the left side of Figure~\ref{fig:dpu}, consists of an MPE that operates on matrices or higher dimensional tensors. The processing engines can be simple MACs, vector processors, or more complex VLIW (Very Long Instruction Word) cores that can support concurrent execution of different instructions.
A popular example in this category is Google's Tensor Processing Unit (TPU).  
Introduced in 2016~\cite{tpu1}, it was originally designed to accelerate Google's TensorFlow framework.
The first generation supported integer arithmetic with a massively parallel INT8 matrix-multiply engine.
The second generation TPU was announced in May 2017~\cite{tpu}, and the third generation in May 2018~\cite{tpu3}.
These newer chips boast improved memory performance as well as support for floating point specifically aimed at training.  
There are a number of startups introducing custom hardware that fall into this category.
Within the cloud , there are Graphcore, Groq, and Wave Computing.
Within the embedded space, where the design constraints are even more stringent, we find even more solutions. %, as are listed in Table~\ref{tableHW-embedded}.
Most are secretive about the details of their designs. Intel is investigating several custom accelerators and has for that purpose acquired a number of startups, namely Nervana, Habana, and Movidius. Fathom~\cite{movidius-tom} is Movidius' ultra low power Neural Compute Stick (NCS) which operates at about 1\,W.  
Also, ARM offers specialized CNN processors in the form of their Ethos family, boosting performance up to 4TOPs with support for INT8 and INT16 datatypes.

%\textcolor{red}{[S.H. Edits needed to the DPU architectures figure: (1) move the left and right further apart - hard to see the separation.  Remove the black box around everything.  Adjust spacing on / between Functions and Pooling.  Remove ... from Pooling]}

As mentioned above, DPUs provide specialized datatypes to execute heavily quantized, reduced precision CNN implementations.
At the extreme, binarized neural networks (which are very high throughput at extremely low power) are exploited in the following ASICs: BinarEye~\cite{binareye}, BNN Custom Fabric~\cite{ando2017brein}, and IBM AI Accelerator~\cite{IBMAI}. Also, Lattice has announced binarized neural network libraries targeting low power FPGA and achieving 1\,TOPs/W~\cite{lattice-bnn}.
Custom floating point representations are also considered. For example, Microsoft's Brainwave project~\cite{chung2018serving} uses this approach with the aim of applying FPGAs to CNNs at datacenter scale.
However, typically the hardened versions in ASICs only support INT8, as lower precisions could potentially limit their application scope. FPGA-based MPE implementations such as Xilinx's xDNN are less constrained and in principle can be customized as needed.

Similar to the GPU, but perhaps to a lesser degree, DPUs leverage input, IFM (input feature map) and OFM (output feature map) parallelism, which requires buffering of inputs and may have adverse effects on latency as well.
A particular challenge arises in the context of software environments, which differ for all vendors and are less mature than what we have observed for CPUs and GPUs. Typically, they are limited to support execution of very specific layer types (sometimes even restricted in regards to parameter ranges) and neural networks, whereby the range of layer types and neural network models is continuously expanding.
%\textcolor{red}{[S.H. This statement is a bit awkward.  Please rewrite]A given and increasingly expanding modelzoo is supported, however many limitations exist in the form of layer types and are provided as black box solutions which lack transparency and flexibility.} 

In summary, through their specialization, these implementations minimize hardware cost, maximize performance and optimize efficiency by exploiting specific precision arithmetic with a specialized instruction set and customized memory system. However, in order to gain a performance advantage, the algorithms need to be adapted to leverage these features. 

\paragraph*{Spatial DPUs.} The second type of DPU leverages spatial acceleration and exploits layer and branch parallelism. Popular examples are hls4ml~\cite{hls4mldata_150p} and FINN~\cite{umuroglu2017finn,blott2018finnr}.
To that extent, the hardware architecture is even further specialized to the specifics of a given deep learning topology. 
This is visualized on the right side of Figure~\ref{fig:dpu}. 
The hardware architecture actually mimics the given deep learning topology and the inputs are streamed through the architecture. 
Every layer is instantiated with a dedicated compute datapath. 
Each layer has a dedicated weight buffer, and activation buffers in-between layers are FIFOs of minimal size. They buffer just enough data to feed the next set of convolutions in the next layer. 
This is substantially more efficient compared to the first type of DPUs or GPUs and yields reduced latency. 
%, where all activations between 2 layers for a full batch of images need to be double-buffered. 

DPUs and GPUs generally perform a layer-by-layer compute, where a sequence of images has to be buffered in order to extract maximum compute out of the platform (input, IFM and OFM parallelism). 
For this, the device buffers a batch of images before computing the first layer of all images. 
Then all intermediate results are buffered, and then the next layer is computed, and so on. Hence the latency is heavily dependent on the size of the input batch.

As a result, spatial DPUs have an advantage in regard to latency.
This level of customization is only possible with programmable hardware architectures such as FPGAs, as they can adapt the hardware architecture for different use cases. 
This generally wouldn't make sense in the context of an ASIC accelerator, as that would yield an ASIC only capable of accelerating one specific topology, which would be far too restrictive in scope.
The limitation in spatial architectures is the scalability in the numbers of layers. 
Each layer comes at a resource cost overhead and there is a maximum number of layers that can be created within a single device. As a result, some extremely deep CNNs might not be able to fit into a single device. Microsoft's Brainwave project leverages spatial computing and overcomes this limitation with a distributed approach~\cite{chung2018serving}.

Once a spatial DPU has been leveraged and the architecture is specialized for a very specific CNN, the architecture can be further customized in regards to minimum precision.  By supporting only the bits as needed per layer of the CNN they can achieve even higher performance and efficiency, while in an MPE, the hardware will support the maximum precision that is required over the whole network. In regards to customized precisions and spatial architectures, FINN has pioneered the first binarized neural network accelerators \cite{umuroglu2017finn,fraser2017scaling} and provided many proof points for customized reduced precision implementations~\cite{blott2018finnr}. 
This flexibility comes at a cost, in the form of programming complexity, and they are extremely difficult to characterize in general, as the performance characteristics depend on the specifics of the hardware architecture that has been implemented.

\paragraph*{Further Variants of DPUs}
Beyond the previously discussed spatial DPUs and MPEs, there are many more variants.
%\textcolor{red}{[S.H. This section is kinda abrupt.  Cut, or smooth the transitions into it.  Also, how does the next paragraph relate to this section?]
Some exploit sparse computing engines for example, such as EIE and its successor ESE~\cite{han2017ese}, SCNN~\cite{parashar2017scnn}, Cnvlutin~\cite{cnvlutin}, Cambricon-S and Cambricon-X~\cite{zhang2016cambricon}. 
These are the only architectures that can benefit from irregular sparsity.
Finally, another dimension for customization of precision is to optimize over the execution- or run-time of a CNN.
In other words, beyond using statically fixed reduced precision, where the hardware operates with a fixed precision for all variables, some approaches explore run-time configurable bit precision which allows for the exploitation of bit-parallelism in the arithmetic. 
% This is also referred to as transprecision computing and illustrated in Figure~\ref{fig:bismo}. 
% On the x-axis, we show the execution time of a CNN, where we first execute all layers (layer 1, layer 2 and so on) for input i0.
% Then we progress to the second input i1 and so on.
% The y-axis annotates the minimum precision required.
% In the shown example, layer 1 for i0 requires 3bits, while layer 2 for i0 requires 2bits.
% This can vary for every input.
% In case of a FP32 implementation, marked by the top line, the area between the red boxes and line would be representative of the wasted computation.
% When leveraging hardened customized arithmetic, for example INT8, then the wasted computation is significantly reduced, but there is further scope for optimization.
On the hardware implementation side, this can be exploited with run-time programmable precision and is effective with \textbf{bit-serial} implementations. 
For example Umuroglu et al.~\cite{umuroglu2018BISMO} demonstrate with BISMO that bit-serial can provide highly attractive performance with minimal overhead on FPGAs, while Judd et al. show the same is true for ASICs with their prototype ASIC called Stripes~\cite{judd2016stripes}. 
While this concept can be applied to both MPE and spatial architectures, it makes the most sense for MPEs.

\begin{table}[!ht]
\resizebox{\linewidth}{!}{
\begin{tabular}{|c||c|c|c|c|c|c|c|}
\hline
\textbf{Server-class} & \textbf{Throughput} & \textbf{Latency} & \textbf{Power} & \textbf{Ext. Mem. Bandwidth} & \textbf{HW specialization} & \textbf{Ease of Use} & \textbf{Training/Inference} \\
\hline \hline
\textbf{Conventional} & & & & & & & \\
CPU & Medium & High & High & Medium & Low & High & Both \\
DPU-MPE & High & Medium-High & Medium & High & Medium & Low-Medium & Inference \\
% e.g. DPU-MPE (Gaphcore IPU) & Medium & Medium & Medium & High & Medium & Medium & Inference \\
DPU-Spatial & High & Low & Medium & High & High & Low & Inference \\
GPU (NVIDIA A100) & High & High & High & High & Medium & High & Both \\
\hline \hline
\textbf{Speculative} & & & & & & & \\
Cerebras CS-1 & Very High & Medium & High & Very High & Medium & Medium & Both \\
\hline
\end{tabular}}
\caption{Characterization of types of hardware based on important metrics.% -- these are used as input to Fig.~\ref{fig:hwcompare}.
\label{tab:hwcompare}}
\end{table}

%\begin{figure}[h!]
%\centering
%\includegraphics[width=0.8\linewidth]{figures/radar_chart.png}
%\caption{Qualitative comparison of popular server-class hardware platforms.}
%\label{fig:hwcompare}
%\end{figure}

%\textcolor{red}{[S.H. Table: seems strange to have two different MPU's listed.  Diagram - seems superfluous, since it has the same info as the table - I'd cut it.] [MB: Agreed, I'm okay with removing the graphcore IPU example. Regarding the radar chart, you're right of course, this is just a visualization of the same thing. If you're stuck for space, feel free to cut it]}

\paragraph*{Summary of Conventional CMOS Hardware Architectures}

We analyzed three categories of hardware architectures that are leveraged for CNN inference, namely common CPUs, SIMD-based vector processors such as GPUs, and DPUs which are specialized architectures for the acceleration of deep learning workloads. 
An overview of the architectures is visualized in Table~\ref{tab:hwcompare}.
Please note, "Ease of Use" includes compute kernel programmability as well as general ease of use.
The degree of specialization includes operators, precision support, and customization towards topologies.
In summary, for DPUs, we distinguish between tensor processors which leverage a matrix of processing engines and spatial architectures which can be further specialized for specific topologies using FPGAs.
CPUs are the most general solution but high in power. GPUs and DPUs offer the highest performance, though GPU are more expensive in energy cost. Spatial DPU architectures excel at low latency and provide the highest compute efficiency through maximized customization. 
CPUs, GPUs, and DPUs (MPE) use a sequential layer-by-layer compute model whereas spatial DPUs execute all layers of the network concurrently. 
Hardened topologies in form of ASICs, CPU and GPU offer a fixed set of native dataypes, whereas FPGAs can adopt any precision and numerical representation, which provides the utmost flexibility and leverages optimization with quantization to the maximum, whereas hardened approaches need to default to the next higher supported precision into which the reduced precision variable can be embedded. However, the programmability in the FPGA fabric also comes at a speed and energy cost.
All architectures can benefit from coarse-grained pruning optimization techniques. Only sparse execution engines can benefit from irregular pruning, such as synaptic pruning.
We also discussed the various deployment options. 
Many devices offer different power and operating modes as different compromises between throughput and power consumption to adapt to the potentially very different optimization targets of different application settings. Similarly, batch sizes, thread counts and stream sizes offer another compromise in regards to throughput versus latency. Again this is to facilitate a spectrum of different use cases.
Finally, the table shows that speculative approaches such as Cerebras can bring fundamental performance scalability.
Overall, each approach comes with its own advantages and disadvantages and the best solution greatly depends on the specifics of a given use case.

%\subsubsection{Conventional ML Hardware}
%\subsubsection{Emerging Beyond CMOS Hardware}
%%%%%%%%%%%%%%%%%%%%%%%%%%%%%%%%%%%%%%%%%%%%%%%%%%%%%%%%%%%%%%%%%%
\subsection{Hardware/Software Codesign Example: FPGA-based Systems}
\label{sec:codesign}

In the last decade, we have observed the rise of two significant paradigms that have come to scientific applications: heterogeneous-computing systems and machine learning. Heterogeneous computing can overcome the decline of Moore's Law and Dennard Scaling and achieve the desired computational cost and performance by executing portions of the applications on the best-matched hardware, e.g., CPU, GPU, ASIC, and FPGA. On the other hand, machine learning is an automatic process that creates programs that can solve classes of problems. As with traditional programming, machine learning can significantly benefit from heterogeneous computing; in addition, designers can tailor specialized but reprogrammable hardware to fit ever-changing machine learning requirements. This section examines tools and methodologies that can automatically deploy and orchestrate machine learning on FPGA systems in larger scientific applications.  FPGAs are a particularly compelling example to explore because the efficiency of the hardware coupled with their programmability makes for an interesting case study in hardware/software codesign.   

Traditional software programming is complicated, and parallel high-performance programming is even more challenging. Programming heterogeneous systems that integrate FPGAs bring the challenge to the next level: the programmer must deal with a multi-objective optimization problem that involves performance and costs, i.e., hardware resources.
For machine learning applications, a common practice is to profile the application on CPU (or GPU) to identify the bottlenecks to be offloaded onto the reprogrammable logic to improve latency, throughput, or energy efficiency of the application as a whole. Then, part of the application can remain on the CPUs to control the execution and interact with the rest of the scientific setup.

\paragraph*{FPGA Programming}
FPGA are configurable integrated circuits that provide a good trade-off in
terms of performance, power consumption, and flexibility with
respect to other hardware paradigms. However, it is a challenging and lengthy task to program FPGAs. FPGA programming has traditionally been a job for hardware designers familiar with digital design and computer architecture. These requirements lead to a steep learning curve for software developers and other domain experts. In order to lower the entry barrier, there has been a growing focus on designing FPGA hardware at a higher level of abstraction. As a result, various approaches have brought FPGA development into the mainstream by allowing developers to design for FPGAs at a higher level using familiar languages such as C, C++, OpenCL, and in some cases, even C\# \cite{kiwiHLS}. Here an important question arises: what are the additional advantages of designing the hardware at a higher level of abstraction? High-level languages (HLLs) include various constructs and design patterns that are more functionally expressive. Furthermore, the amount of time spent in the verification of the design is also a crucial factor. Hardware-description languages such as Verilog or VHDL focus on the final implementation details and, because of that, are more verbose. Bigger code repositories are not easy to verify for functional correctness. On the other hand, HLLs are more compact and simulate faster. Thus, a designer can do more verification in the same span of time. Despite these advances, FPGA programming remains complex. This has compelled academia and industry to develop new compilers, frameworks, and libraries to facilitate hardware design.

\paragraph*{High-Level Synthesis and Languages}

High-level synthesis (HLS), also known as behavioral or algorithmic synthesis, is an automated design process that takes as input a functional description of a design and outputs an RTL implementation. It transforms an untimed (or partially timed) high-level specification into a fully timed implementation. The process of HLS starts by analyzing the data dependencies between the various operations in the functional description. This analysis leads to a Data Flow Graph (DFG) representation. After the DFG generation, during the allocation phase, HLS maps each operation onto a hardware resource with latency and area characteristics. Then, HLS adds the notion of time to the design during the scheduling phase. Scheduling takes the operations and resources of the DFG and decides in which clock cycle to execute them, given their latency information. This step infers sequential logic by adding registers between operations and creating finite state machines~\cite{hlsbluebook}. 

Over the past three decades, many HLS tools have been proposed. The work in \cite{survey_evaluation} presents an evaluation of different academic and commercial HLS tools tested on the same set of benchmarks. These tools have different input languages, perform different internal optimizations, and produce different quality results, even for the same input languages. The results show that each HLS tool can significantly improve performance once the designer has mastered benchmark-specific optimizations and constraints. However, academic HLS tools have a higher learning curve because of a minor focus on usability. Commercial HLS tools have an advantage because of their better documentation, robustness, and design verification integration.

In terms of input languages for HLS, most of the HLLs are variants of the C language. However, there are a few limitations to generate hardware from a pure C specification. First, C lacks the notion of timing and concurrency. The designer must rely on the HLS tool to create clock-based timing. Similarly, the designer must specify the concurrency model or rely on HLS to extract the parallelism among operations or processes. Second, C lacks bit-accurate data types. It only provides ``native'' data types such as \texttt{char}, \texttt{int}, and \texttt{long}, whose size is a multiple of a byte. Third, it lacks the concepts of hardware interfaces and communication channels. SystemC was adopted as HLS language to address all of these limitations~\cite{6838614}. However, SystemC still has not entirely made inroads in the FPGA community. Another common problem with all C-based languages, including SystemC, is memory access and modeling. These languages have a flat memory model, and memory access is done through pointers. Either HLS has to decide how to implement the memories in hardware, or the designer must leverage additional HLS directives or libraries to model the memory sub-system properly. Finally, in the family of the C-based specification languages for HLS, the SYCL language is emerging. SYCL (pronounced sickle) is an industry-driven standard that adds parallelism to C++ to design heterogeneous systems. SYCL programs perform best when paired with SYCL-aware C++ compilers such as the open-source data-parallel C++ (DPC++) compiler \cite{dpcplus}.

Apart from the variations of C, Bluespec is an open-source language for the description and synthesis of hardware based on SystemVerilog. It provides levels of abstraction with a clean semantic that highlights aspects of the architecture. It can be considered a high-level functional HDL, where modules are implemented as rules using SystemVerilog syntax. Those rules are called guarded atomic actions and express behaviors as concurrently cooperating finite state machines (FSMs). Another recent language among FPGA designers is Chisel. It is based on Scala and supports hardware definition using highly parameterized generators, object-oriented and functional programming. Similar to an HLS flow, it compiles into an RTL Verilog implementation.

\begin{table}[h]
	\centering
		\caption {A brief taxonomy of domain-specific languages and frameworks for FPGA applications~\label{DSLs}}
		\begin{tabular}{ | l | l | l | p{0.1cm}|}
			\hline
			\textbf{Domain and Interfaces} &  \textbf{DSLs and Frameworks} \\ \hline  
			Signal-Processing &  HDLCoder \cite{dsl}, LabView \cite{dsl}, Spiral \cite{SPIRAL}, VSIPL \cite{VSIPL}  \\            \hline
			Networking  &  SNORT \cite{snort}, Click \cite{click}, P4 \cite{p4lang}, Floem \cite{floem}\\            \hline
			Databases & Glacier \cite{glacier}  \\ \hline
			Machine Learning & OptiML \cite{optiML}  \\ \hline 
			Numerics &Verilog
			AMS \cite{verilogAMS}  \\            \hline
			Streaming & Maxeler \cite{maxeler}, SCORE \cite{score}, Lime \cite{lime}, Aetherling \cite{aetherling}  \\            \hline
			Dataflow & OpenDF \cite{openDF}, OpenSpatial \cite{dsl}  \\            \hline
			Graphs & GraphStep \cite{graphstep}, GraphGen \cite{graphgen}  \\            \hline
			Data Parallel & MapReduce \cite{dsl}, Accelerator \cite{accelerator}, FCUDA \cite{fcuda}, SuSy \cite{susy}  \\            \hline
			Circuit Generators & Flopoco \cite{flopoco}, JHDL \cite{jhdl}, PAMDC \cite{pmdc}  \\            \hline
			Image processing & HIPACC \cite{hipacc}, FROST \cite{frost}, Darkroom \cite{darkroom}, RIPL \cite{ripl}, PolyMage \cite{polymage} \\            \hline
 			Static &  JBits \cite{jbits}, TVM \cite{tvm}\\            \hline
 			Task based & TAPAS \cite{tapas} \\            \hline
				Dynamic &  PyRTL \cite{pyrtl}, APARAPI \cite{arapa}, TornadoVM\cite{tornado}, Caldeira et al. \cite{caldeira}, \\ & LINQits \cite{linqits}, DHDL \cite{dhdl}, Spatial \cite{spatial} \\            \hline
		Type Systems & DAHLIA \cite{dahlia}\\ \hline
		Verification & Kami \cite{kami} \\ \hline
		Virtualization & Cascade \cite{cascade} \\ \hline
		\end{tabular}
\end{table}

Although all these languages have helped create efficient hardware and significantly shorten the development time, specific coding techniques are still necessary.
Also, the growth and diversification of the application domains have shown the limitations of these programming languages. This has further pushed the level of abstraction to domain-specific languages (DSLs). 
In recent years, we are observing the growth of a considerable corpus of DSLs and frameworks for FPGA designs~\cite{dsl, transparent}. 
In a DSL-based approach, the users and the tools can use domain knowledge to apply static and dynamic optimizations. 
However, a domain-specific HLS tool requires an appropriate compiler and a development environment that caters to the target domain. 
Table~\ref{DSLs} shows some of the DSLs and frameworks developed over the years for FPGA computing organized by domains of application. 
Although all the approaches in the table are diverse in terms of applications, the interesting question is, what are the common denominators? 
To the best of our knowledge, most of the approaches are broadly based on two approaches: either the DSL specification gets directly compiled into the RTL implementation, or the approach leverages source-to-source compilers. 
In the latter case, the DSL compiler produces an equivalent source code in a different programming language, for example, C++, for a more standard HLS flow.   
As a final concluding remark for this paragraph, the efforts for designing 
better HLS compilers and languages are a significant part of present FPGA research. 
Furthermore, the work in Table 5 by no means is an exhaustive list. 
The area of DSLs for FPGA easily outnumbers the work presented in the table.

\paragraph*{Software and Hardware Integration}

Running an application as software on a microprocessor is more accessible than designing and running specialized hardware, but it may result in poor performance and higher power costs. On the other hand, partitioning an application into software and hardware components is challenging. This process, also known as hardware/software codesign, divides an application between software running on the microprocessor and one or more custom hardware or co-processors components to achieve desired performance goals. Understandably there exists a plethora of research work in this area. The authors in \cite{rfu} have provided background information on notable aspects of older FPGA technologies and simultaneously explained the fundamental architectures and design methods for codesign. Furthermore, the work in \cite{microarchitecture} is another comprehensive study that aims to evaluate and analyze the microarchitectural characteristics of state-of-the-art CPU-FPGA platforms in depth. That paper covers most of the shared-memory platforms with detailed benchmarks.

%Nonetheless, partitioning involves various design decisions. One of the critical challenges is enabling higher design productivity and a more accessible way to use FPGAs for users unfamiliar with the underlying concepts. One way of doing this is to provide standardization and abstraction, usually supported and enforced by an operating system. Operating systems simplify the programming interface through an abstracted programming model.

%\textcolor{red}{[S.H. This section is a but choppy, and needs smoothing. Also odd to be exclusively Xilinx.][MB: I can't help with that}
The two leading FPGA vendors, Xilinx and Intel, have their own solutions. The Xilinx Runtime Library (XRT) \cite{XRT} is implemented as a combination of userspace and kernel driver components. It supports both PCIe-based boards and MPSoC based embedded platforms. Similarly, Xilinx SDSoc \cite{sdsoc} and SDAccel \cite{sdaccel} became publicly available later in late 2015; the former works only on select boards of the Zynq family of FPGAs, the latter only on selected PCIe-based boards for OpenCL computing. Since 2020 Xilinx has introduced Vitis \cite{vitis} as a unified platform. Vitis Unified Software Platform is a comprehensive development environment to build and seamlessly deploy accelerated applications on Xilinx platforms, including on-premises Alveo cards, FPGA-instances in the cloud, and embedded platforms.
In addition, the recent efforts of Xilinx under the flagship Versal \cite{versal} is also a step towards codesign applications. Intel has the Open Programmable Acceleration Engine (OPAE) \cite{intelopae} which is the API library for programmers writing host applications that will leverage the FPGA acceleration. Likewise, Intel oneAPI \cite{inteloneapi} is an open, unified programming model built on standards to simplify the development and deployment of data-centric workloads across CPUs, GPUs, FPGAs, and other accelerators.

Apart from vendor solutions, academia and the open-source community have also attempted to simplify the integration of applications, operating systems, and hardware acceleration. For a comprehensive analysis, the reader is referred to the works in \cite{reconfig_os, connectal}, which give a historical review and summary on ideas and key concepts to include reconfigurable computing aspects in operating systems. They also present an overview of published and available operating systems of the last 30 years targeting reconfigurable computing. Similarly, the design exploration and engineering of FPGA drivers that are portable across multiple physical interfaces (PCIe, Ethernet, optical links) have remained a significant part of HW/SW codesign research. The challenges come from the variety of FPGA boards, the plethora of interfaces, and the diverse user requirements. Fundamentally, the FPGA drivers should allow the designer to load or reconfigure an application bitstream and support data transfers between the FPGA and host.

A significant engineering challenge is to consider how to partition driver functionality between the hardware and software components. One growing research focus is to exploit the spatial parallelism of FPGA technology through implementing multiple queues on FPGA drivers. A thorough analysis of system-level drivers for FPGA is out of the scope of our white paper. Readers interested in FPGA system-level drivers are referred to the work in \cite{fpga_drivers, fpgadrivers2}. The authors of those papers have provided benchmarks of various mainstream academic and vendor solutions regarding system-level drivers in the FPGA domain. 

Despite various existing OS and driver solutions, an open problem that remains is standardization. An industry-wide standardization would allow for faster development and better portability, and (re)usability of FPGA applications. There is already ongoing work in this area. Standards like the CCIX consortium \cite{ccix} and the Heterogeneous System Architecture (HSA) foundation \cite{hsa}  have already made good progress.

\paragraph*{The Case for ML Frameworks for FPGA Design}

\begin{figure}
\centering
\includegraphics[width=0.8\linewidth]{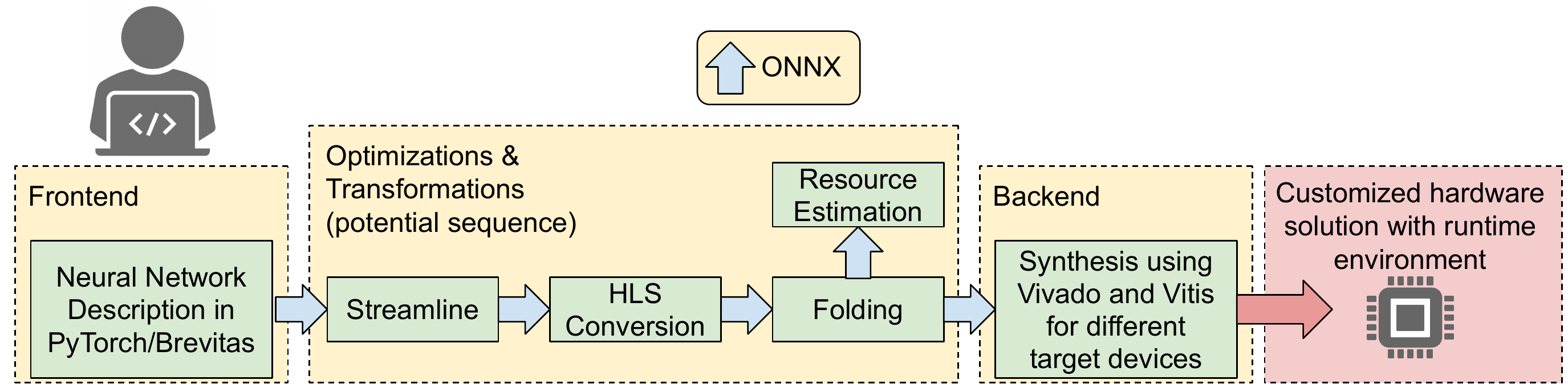}  
\caption{FINN Compiler Flow}\label{figures:finnflow}
\end{figure}

Machine learning is one of the fastest growing application domains and over the years there has been an increasing demand for FPGA-based implementations, as the FPGA can achieve latency and throughput and efficiency requirements through extreme customization of the hardware design leveraging reduced precision arithmetic, streaming dataflow implementations (as were introduced as spatial architectures), and fine-granular sparsity.  In order to enable a broad spectrum of users with these customizations and to reduce the significant engineering effort, compilers and tools are needed that cater to the needs of ML researchers and domain experts working with FPGAs. Two main ML frameworks are making the effort to fill this vacuum: hls4ml and FINN. Considering the aforementioned tools, compilers, programming languages, and codesign solutions, both hls4ml and FINN have the potential to reach a broader scientific community. 
%\textcolor{red}{Add a description of FINN, while hls4ml has been described in Sec. 2.1.4}.
To get a better understanding of how such a tool flow works, we consider the FINN compiler in more detail in the following paragraphs.

The \textbf{FINN compiler}~\cite{umuroglu2017finn} is an open-source framework to generate spatial DPU or streaming dataflow accelerators on FPGAs. 
The FINN compiler has a highly modular structure as shown in Figure~\ref{figures:finnflow}, which allows the user to interactively generate a specialized architecture for a specific DNN. 
The framework provides a frontend, transformation and analysis passes, and multiple backends to explore the design space in terms of resource and throughput constraints.
Brevitas~\cite{brevitas}, a PyTorch library for quantization-aware training, is the \textbf{frontend} used in this work.
It enables training DNNs with weights and activations quantized down to a few bits, then exports the trained network into the intermediate representation (IR) used by the FINN compiler. 
The \textbf{transformation and analysis passes} help to generate an efficient representation of the DNN. Finally, the \textbf{backend} contains a code generator that creates synthesizable accelerator descriptions, which can be implemented as either a standalone Vivado IPI component or integrated into various shells, including Xilinx Alveo boards and PYNQ embedded platforms.

For further processing, the DNN model must be converted into the IR of the FINN compiler first. The frontend stage takes care of this by converting the PyTorch description into the IR, called FINN-ONNX. This IR is based on ONNX~\cite{onnx_github}, an open-source interchange format that uses a protobuf description to represent DNNs. It comes with several standard operators and allows the user to easily create their own operators to customize the model. The nodes represent layers and edges carry outputs from one layer to become inputs to another. The feature to customize the ONNX representation is used in the framework to add application-specific nodes and attributes. Each node is tagged with the quantization of its inputs, parameters (weights and activations), and outputs to enable quantization-aware optimizations and the mapping to backend primitives optimized for quantized computation. During the compiler flow the nodes will be transformed into a backend-specific variants via a series of transformation passes. 

The main principle of the \textbf{FINN compiler} is \textbf{graph transformation} and \textbf{analysis passes}, which change or analyze the IR of the model. A pass is a function that takes the IR graph as input and either (a) transforms the DNN by looking for a certain pattern, changing the graph in a specific manner and outputs the modified graph, or (b) \textbf{analyzes} the DNN to produce metadata about its properties. To bring the model into a representation from which code can be produced and finally the hardware accelerator can be generated, various transformations must be applied. The main transformations involved  are summarized below.

Although the PyTorch description of the network is mostly quantized, it may still contain some floating-point operations from e.g. preprocessing, channelwise scaling or batchnorm layers. In order to generate a hardware accelerator from the model, these floating-point operations must be absorbed into multi-level thresholds, so that a functionally identical network of integer operations is created. The transformation to achieve this is called \textbf{streamlining}, as described by Umuroglu and Jahre~\cite{DBLP:journals/corr/abs-1709-04060}. During streamlining, floating-point operations are moved next to each other, collapsed into a single operation, and absorbed into succeeding multi-thresholding nodes.

Next, high-level operations in the graph are \textbf{lowered} to simpler implementations that exist in the FINN HLS-based hardware library. For instance, convolutions will be lowered to a sliding window node followed by a matrix-vector node, while pooling operations will be implemented by a sliding window followed by an aggregation operator. The resulting graph now consists of layers that can be converted to hardware building block equivalents. Each node corresponds to a Vivado HLS C++ function call, from which an IP block per layer can be generated using Vivado. The resources utilized by each hardware building block can be controlled through specific attributes passed from FINN to Vivado. For example, multiplications can be performed with LUTs or DSP blocks, and parameters can be stored in distributed, Block, or Ultra RAM.

Finally, the \textbf{folding} process assigns compute resources to each layer to obtain the desired throughput with a balanced pipeline by fine-tuning their degree of parallelism. 
To enable per-layer specialization without reconfiguration and minimize latency, FINN creates dedicated per-layer hardware interconnected with FIFO channels, thus the outermost loop across $L$ layers is always fully pipelined.
Once the folding is specified, resource estimates can be produced for each node. 
There are several ways to estimate the resources. Even before IP blocks are generated from the HLS layers, an estimate of the resources per layer can be made by using analytical models based on the concepts from the FINN-R paper~\cite{blott2018finnr}. 
Estimations can also be extracted from Vivado HLS after IP generation, though these results are still estimations that may differ from the resource usage of the final implementation due to synthesis optimizations.

The \textbf{Backend} is responsible for consuming the IR graph and backend-specific information to create a deployment package, also implemented using the transformation concept.
To get the inference accelerator, between the layers FIFOs are inserted, which can be sized automatically by the FINN compiler. Afterwards, the single IP blocks are stitched together and synthesized. The stitched IP can be manually integrated into a system, or inserted into an appropriate shell for the target platform. If the target platform is an Alveo card, the design is exported as a Vivado Design Checkpoint (DCP), followed by generation of Xilinx Vitis~\cite{kathail2020xilinx} object files and linking.

%\textcolor{red}{[S.H. Need a section to wrap up 4.4.  Summary, conclusion, what have you.]}
\paragraph*{Summary of Hardware/Software Codesign and FPGA-based Systems}
%In summary, we analysed three categories of standard CMOS-based hardware architectures that are leveraged for CNN inference, namely common CPUs, SIMD based vector processors such as GPUs, and DPUs which are specialized architectures for acceleration of deep learning workloads. For DPUs, we distinguish between tensor processors which leverage a matrix of processing engines and spatial architectures which can be further specialized for specific topologies using FPGAs.
In summary, CPUs are the most general solution for CNN inference but high in power. GPUs and DPUs offer highest performance, whereby GPU are more expensive in regards to energy cost. FPGAs offer several tradeoffs that may well fit rapidly moving application domains. %Spatial DPU architectures excel at latency and provide highest compute efficiency through maximized customization. 
%CPUs, GPUs and DPUs (MPE) use a sequential layer by layer compute model whereas spatial DPUs execute all layers of the network concurrently. 
%Hardened topologies in form of ASICs, CPU and GPU offer a fixed set of native dataypes, whereas 
FPGAs can adopt any precision and numerical representation, which provides utmost flexibility and leverages optimization with quantization to the maximum, whereas hardened approaches need to default to the next higher supported precision where the reduced precision variable can be embedded.
Furthermore, through the spatial dataflow approach, much lower latency can be achieved.
However, the complexity of programming FPGAs limits their deployment.
Tools such as hls4ml and FINN are frameworks specifically created for the ML domain where they automate the process of hardware generation for the end-user thus hiding the associated design complexity of FPGAs and enabling them for the previously discussed end applications.

\subsection{Beyond-CMOS neuromorphic hardware}
\label{sec:beyondcmos}

With rapidly growing machine learning applications comes the acute need for their efficient hardware implementations. Most of the efforts are focused on digital CMOS technology, such as implementations based on general-purpose TPUs/GPUs, FPGAs, and more specialized ML hardware accelerators.  The steady improvements in such hardware platforms' performance and energy efficiency over the past decade are attributed to the use of very advanced, sub-10-nm CMOS processes and holistic optimization of circuits, architectures, and algorithms.  It includes, for example, taking advantage of aggressive voltage supply scaling \cite{Moons2017}, very deep pipelines and extensive data reuse in architectures \cite{Chen2017}, and lowering the precision of weights and activations of the algorithms \cite{Simons2019}.  As a result, very compact state-of-the-art neural networks, such as MobileNet based on 3.4M parameters and 300M multiply-and-add operations per inference \cite{Sandler2018}, can now be fitted entirely on a single chip. However, on all these fronts, advances are saturating and cannot rely on the faltering Moore's law. 

On the other hand, further progress would be essential because ML algorithms are getting increasingly more complex. For example, transformer networks \cite{Vaswani2017}, the state-of-the-art approach for many ML tasks today \cite{Vaswani2017, Vinyals2019, Dosovitskiy2020}, could have hundreds of billions of parameters and perform hundreds of trillions of operations per inference. Moreover, the transformer's functional performance typically improves with the model size \cite{Rajbhandari2020,Brown2020}. Training such models requires enormous, data-center-scale (e.g., kiloTPU-year) resources while performing inference on resource-constrained edge devices would be extremely challenging. 

The opportunities for building more efficient hardware may come from biological neural networks. Indeed, it is believed that the human brain, with its $>$1000$\times$ more synapses than the weights in the largest transformer network, is extremely energy efficient \cite{Hasler2013}, which serves as a general motivation for developing neuromorphic hardware \cite{Mead1990}. There is a long history of CMOS neuromorphic circuits \cite{Indiveri2011}. However, unleashing the full potential of neuromorphic computing might require novel, beyond-CMOS device and circuit technologies \cite{Berggren2020} that allow for more efficient implementations of various functionalities of biological neural systems. 

In this section, the most prominent emerging technology proposals, including those based on emerging dense analog memory device circuits, are grouped according to the targeted low-level neuromorphic functionality - see, e.g. reviews in \cite{Burr2017, Bavandpour2018, Yang2013NatureNano, Yu2018IEEE} and original work utilizing volatile \cite{Sheridan2017, Cai2019NatElec, Chu2014Neuro, Yeon2020, Ohno2011, Wang2017NatMat, Pickett2013, Wang2018NatElec, Zhang2018Small, Lashkare2018, Adda2018} and nonvolatile \cite{Alibart2012, Adam2017,Govoreanu2013, Prezioso2015, Prezioso2016, Prezioso2018, MerrikhBayat2018, Lin2020NatElec, Hu2018AdvMat, Yao2020Nature, Liu2020ISSCC, Kim2019XBAR, Cai2020NatElec, Mahmoodi2019IEDM, Mahmoodi2019NatComm, Li2016IEDM, Wang2018NatElec, Pedretti2017}  memristors, phase change memories (PCM) \cite{Burr2015, Tuma2016, Ambrogio2018, Karunaratne2020, Joshi2020, Kuzum2011, Rios2019}, and nonvolatile NOR \cite{Guo2017CICC, Guo2017IEDM, MerrikhBayat2015, Mahmoodi2019NatComm}, and NAND \cite{Bavandpour2019NAND, Bavandpour2020, Lee2019NAND}, and organic volatile \cite{Fuller2019} floating gate memories, as well as multiferroic and spintronic \cite{Grollier2020, Ostwal2018, Sengupta2016, Romera2018, Ni2018IEDM}, photonic \cite{Shasti2021, Goi2020, Rios2019, Lin2019Sciece, Hamerly2019PhysRevX, Hamley2019, Shen2017NatPhot, Tait2016, Feldmann2019, Buckley2017, Bruiner2013, Vandoorne2014}, and superconductor \cite{Segall2017, Buckley2017, Rowlands2021} circuits. More discussion is devoted to analog vector-by-matrix multiplication circuits in the following subsection because of their immediate value for today's state-of-the-art algorithms. More biologically-realistic proposals described in the subsequent sections are less emphasized because they target algorithms with inferior performance. The least mature though very intriguing quantum neuromorphic computing \cite{Yamamoto2020, Markovich2020} is not discussed in this brief review.

\paragraph*{Analog Vector-by-Matrix Multiplication}

\begin{figure}[!t]
\centering
\includegraphics[width=0.30\textwidth]{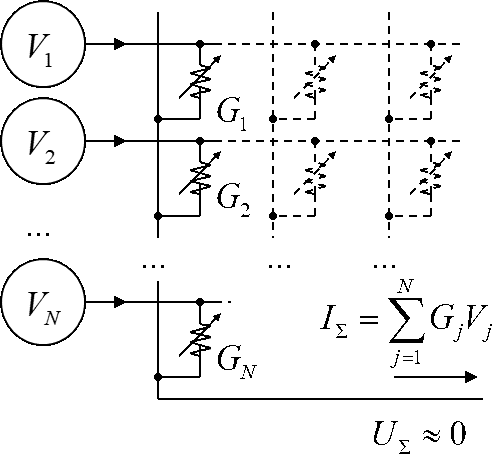}
\caption{Analog vector-by-matrix multiplication (VMM) in a crossbar circuit with adjustable crosspoint devices. For clarity, the output signal is shown for just one column of the array, while sense amplifier circuitry is not shown. Note that other VMM designs, e.g. utilizing duration of applied voltage pulses, rather than their amplitudes, for encoding inputs/outputs, are now being actively explored – see, e.g., their brief review in Ref. \cite{Bavandpour2018}}.
\label{fig:VMM}
\end{figure}

The emergence of dense analog-grade nonvolatile memories in the past two decades renewed interest in analog-circuit implementations of vector-by-matrix multiplication (VMMs) \cite{Mead1990, Holmes1993, Alibart2012, Widrow1962, Chawla2004, Guo2017CICC, MerrikhBayat2015}, which is the most common and frequently performed operation of any neural network in training or inference \cite{Hertz1991, Gerstner2002}. In the simplest case, such a circuit is comprised of a matrix of memory cells that serve as configurable resistors for encoding the matrix (synaptic) weights and peripheral sense amplifiers playing the role of neurons (Fig. \ref{fig:VMM}). The input vector is encoded as voltages applied to rows of the memory matrix so that the currents flowing into virtually grounded columns correspond to VMM results. Because addition and multiplication are performed on the physical level, via Kirchhoff's and Ohm's laws respectively, such an approach can be extremely fast and energy-efficient, provided that memory devices are dense and their conductances are adjustable (i.e., multi-state). The energy efficiency in part comes from performing ``in-memory" computing that reduces the amount of data (corresponding to the synaptic weights) that are moved across or in-and-out of the chip during computation. Such communication overhead could dominate the energy consumption in the most advanced digital CMOS implementations.

The general challenge towards practical adoption of such circuits, especially when using the most prospective emerging memory technologies, is variations in $I$-$V$ characteristics, e.g., in the switching voltages applied to change the memory state.  In light of this challenge, the most straightforward application is ex-situ trained inference accelerators for the earlier firing-rate neural networks \cite{Bavandpour2018}, i.e., the so-called second generation of artificial neural networks (ANNs) with graded-response neurons. In such applications, memory devices are updated infrequently, only when new inference functionality should be programmed. Thus, crosspoint devices' conductances can be tuned with slower, more tolerant to device variations write schemes. For example, after the weights have been found in the software, memory cells are programmed, one by one, using feedback write-verify algorithms that can adapt to the unique $I$-$V$ characteristics of each device \cite{Alibart2012}.  For the same reason, the switching endurance, i.e., the number of times the memory devices can be reliably programmed, and the write speed/energy are less critical.  Additionally, VMM operations in the inference of many neural networks could be performed with moderate, less than 8-bit precision, without incurring accuracy loss \cite{Yang2019IEDM}, which further relaxes requirements for analog properties and permits more $I$-$V$ non-idealities and noise.

The most advanced neuromorphic inference circuits have been demonstrated with more mature floating-gate transistor memory circuits. Up until recently, such circuits were implemented primarily with ``synaptic transistors"~\cite{Diorio1996}, which may be fabricated using the standard CMOS technology, and several sophisticated, efficient systems were demonstrated \cite{Hasler2013, Chawla2004, George2016}.  
However, these devices have relatively large areas ($>$10$^3$ $F^2$, where $F$ is the minimum feature size), leading to higher interconnect capacitance and hence larger time delays. 
More recent work focused on implementing mixed-signal networks with much denser ($\sim$40\unit{F$^2$}) commercial NOR-flash memory arrays redesigned for analog computing applications~\cite{MerrikhBayat2015, Guo2017CICC}. 
For example, a prototype of a 100k+-cell two-layer perceptron network fabricated in a 180-nm process with modified NOR-flash memory technology was reported in Ref.~\cite{Guo2017IEDM}. 
It performed reliably, with negligible long-term drift and temperature sensitivity, and reproducible classification of the MNIST benchmark set images with $\sim95\%$ fidelity and sub-1-$\mu$s time delay and sub-20-nJ energy consumption per pattern. 
The energy-delay product was six orders of magnitude better than the best (at that time) 28-nm digital implementation performing the same task with a similar fidelity~\cite{Guo2017IEDM}. 

Recent theoretical studies showed that neuromorphic inference circuits could be also implemented with much denser 3D-NAND flash memories \cite{Bavandpour2019NAND, Bavandpour2020, Lee2019NAND}, projected to scale eventually to 10 terabits per square inch density. 
In the long term, the most promising are perhaps circuits based on metal-oxide resistive switching random access (ReRAM for short, which are also called metal-oxide memristors)~\cite{Yang2013NatureNano, Yu2018IEEE}, especially their passively integrated (0T1R) technology variety \cite{Kim2019XBAR}. Indeed, due to the ionic switching mechanism, ReRAM devices with dimensions below 10 nm still retain excellent analog properties and year-scale retention \cite{Govoreanu2013}. Furthermore, a low-temperature fabrication budget allows monolithic vertical integration of multiple ReRAM crossbar circuits, further increasing effective density \cite{Adam2017}. There has been rapid progress in scaling up the complexity of ReRAM-based neuromorphic circuit demonstrations over the past several years \cite{Prezioso2015, MerrikhBayat2018, Lin2020NatElec, Hu2018AdvMat, Yao2020Nature, Liu2020ISSCC, Kim2019XBAR}. However, the ReRAM technology is still in much need of improvement. In addition to high device variations, another remaining issue is high write currents and operating conductances, which must be decreased by at least one order of magnitude to reduce the significant overhead of peripheral circuits \cite{Kim2019XBAR}.

The device requirements for training hardware accelerators are different and much more stringent. For instance,  long retention is not required because weights are frequently updated. That allows using volatile memories in analog VMM circuits, such as interfacial memristors based on electron trapping/detrapping switching~\cite{Sheridan2017, Cai2019NatElec, Chu2014Neuro} and solid-state-electrolyte memories~\cite{Fuller2019, Yeon2020, Berggren2020}, or even capacitor-based memories controlling current via crosspoint transistors~\cite{Ambrogio2018}. 
However, the toughest challenge is much higher computing and weight precision required for training operation and the need for efficient schemes for weight updates, which in turn necessitate drastically tighter device variations. 
The additional related requirement is that the change in device conductance upon applying the write pulse should not depend on its current state (the so-called linearity of update property). 
Otherwise, accurate conductance adjustment would require sending a unique write pulse based on the current device state, which would be hardly compatible with fast (in parallel) weight update. 

Phase change memories have also been investigated as candidates for variable resistors in analog VMM circuits \cite{Burr2015, Joshi2020}, though their main drawback is significant drift in the conductive state over time. 
High write endurance, high density (with vertical 3D-NAND-like integrated structure), and long retention are demonstrated in 1T Ferroelectric RAM devices. 
There is much excitement about such devices' applications in training and inference accelerators \cite{Ni2018IEDM}, though their analog properties are probably inferior to ReRAM.  
The significant drawbacks of magnetic devices, such as magnetic tunnel junction memories, are smaller on/off current ratios, insufficient for practical VMM circuits, and poor analog properties for scaled-down devices \cite{Grollier2020}. 

The potentials of using light for implementing fast and large-fanout interconnect and linear computations, such as multiply-and-add operation, have motivated photonic neuromorphic computing research~\cite{Berggren2020,Goi2020,Shasti2021,Hamley2019}. 
Different implementation flavors, e.g., with fixed~\cite{Lin2019Sciece} and programmable~\cite{Rios2019, Hamerly2019PhysRevX, Shen2017NatPhot, Tait2016} functionalities, have been recently suggested in the context of modern neural networks. 
Specifically, Ref.~\cite{Lin2019Sciece} reports a system of multiple 3D-printed optical layers, each being a mesh of regions (neurons) with specifically chosen transmission-reflection properties, which can perform pattern classification inference similar to the convolutional neural networks. By sending a coherent light with amplitude-encoded input, a useful computation is performed at the speed of light. Specifically, the light diffracts and interferes when passing through the optical system and is ultimately steered to the specific region at the output layer corresponding to the pattern class. 
Refs.~\cite{Rios2019, Hamerly2019PhysRevX, Shen2017NatPhot, Tait2016} report optical neuromorphic systems with configurable weights. 
The inputs are encoded in the light's energy, and the weights are encoded by optical attenuation in PCM devices in Ref.~\cite{Rios2019} so that a product is computed by passing the light via PCM device. 
Ref.~\cite{Tait2016} proposes encoding inputs with light amplitude and uses specific frequency for different VMM inputs. 
The light from inputs is combined and passed to the frequency selective weight banks based on a microring resonator (MRR) that features metal heaters to perform multiplication. In particular, the MRR coupling (i.e., weight) is controlled via heating by adjusting currents supplied to each MRR. In these reconfigurable implementations, the product accumulation (i.e., the summation operations in the VMM) is performed by integrating the light-induced charges on the photodetector. A very aggressive time-division multiplexing scheme for calculating VMM in which both weights and inputs are encoded in the coherent light's amplitude is proposed in Ref. \cite{Hamerly2019PhysRevX}. 
At one step of such scheme, the input light is fanned out into $n$ channels and combined with the light-encoded $n$ weights using a beam splitter and then sent to $n$ homodyne photodetectors to compute $n$ products in parallel. 
All-optical feed-forward inference based on Mach-Zehnder interferometer meshes utilizes single-valued decomposition for the weight matrix \cite{Shen2017NatPhot}. Unitary matrix transformations are implemented with optical beam splitters and phase shifters, while the diagonal matrix is implemented with optical attenuators. 

In principle, sub-aJ energy and sub-ps latency for a single multiply-and-add operation might be possible with optical computing~\cite{Hamley2019}. 
However, the main challenge remains much large dimensions of the optical components and the very high I/O overhead of converting to and from optical domains~\cite{Berggren2020, Shasti2021, Hamley2019}. The designs that rely on conversion to the electrical domain would be especially affected by poor integration density of optical devices due to larger electrical communication overheads, which were shown to overwhelm system-level performance of (much denser) ReRAM based circuits \cite{Bavandpour2018}. 
Optical systems would ultimately benefit from very wide ($\gg$10,000) dot-products and/or utilizing deep time-division multiplexing to amortize the I/O overhead. However, the possible issues of nonlinearities in charge integration and utility of such wide dot-product computations remain unclear \cite{Hamley2019}.  

\paragraph*{Stochastic Vector-by-Matrix Multiplication}

Computations performed by the brain are inherently stochastic, in that, e.g. substantially different neural responses are observed to the repeatable presentation of identical stimuli~\cite{Rolls2010}. 
Such noisy operation is mimicked by probabilistic neural networks, such as Boltzmann machines~\cite{Hinton1983} and deep belief neural networks~\cite{Hinton2009}. 
In the simplest case, such a network is comprised of binary neurons that compute stochastic dot products, i.e., probabilistically generate output according to their pre-activation (dot-product) values. 

The stochastic functionality can be realized at either the synapse or the neuron side. In the latter, more straightforward scenario, the neuron first computes a dot-product of its inputs and corresponding weights deterministically. 
The result is then passed to some ``probabilistic" activation function, e.g., used as an argument in the sigmoid probability function, to determine the probability of generating high output. 
Because of the typically large ($>100$) ratio of synapses to neurons, the efficient deterministic dot-product implementations, e.g., with the already discussed analog VMM circuits, is of primary importance for realizing high-performance probabilistic neural network hardware. Still, earlier work showed that even the simplest, deterministic neurons may incur substantial overhead, e.g., occupy up to $30\%$ of the area and consume up to $40\%$ of energy for some neural network models~\cite{Bavandpour2018}. Hence neuromorphic hardware would also benefit from the efficient realization of stochastic neurons.

Emerging devices can be broadly employed in two ways to achieve stochastic functionality, namely by using either dynamic or static $I$-$V$ characteristics of memory devices. Specifically, the former approach is to utilize intrinsically stochastic switching between memory states in emerging memory devices. 
For example, in MTJ memories, thermal fluctuation causes stochastic transition between the low resistance parallel and high resistance antiparallel states so that the probability of the final memory state upon switching could be controlled by the spin-torque current~\cite{Grollier2020}. The melt-quench-induced reconfiguration of the atomic structure is intrinsically stochastic in phase-change memories (PCMs)~\cite{Tuma2016}. 
These phenomena were suggested for implementing MTJ~\cite{Ostwal2018} and PCM~\cite{Tuma2016} stochastic neurons. 
The second approach is to utilize intrinsic and extrinsic current fluctuations in memory devices, e.g., random telegraph~\cite{Cai2020NatElec} and thermal noise~\cite{Mahmoodi2019IEDM} in ReRAM devices, or shot-noise in nanoscale floating gate transistors~\cite{Mahmoodi2019IEDM, Mahmoodi2019NatComm}. 
In such an approach, the noisy current flowing into the neuron is compared against a reference value, e.g. using a simple latch, to implement a probabilistic activation function~\cite{Mahmoodi2019NatComm}. 

The primary concern for the former approach is the limited endurance of many memories and the drift in the stochastic switching properties upon repeated switching. 
An additional drawback is a necessity for the co-integration of multiple memory device technologies for scalable stochastic dot-product circuits, e.g., integrating ReRAM-based artificial synapses and MTJ-based neurons. 
On the other hand, analog circuits based on ReRAM devices only (Fig.~\ref{fig:VMM}), though operating at a much lower signal-to-noise ratio (SNR), can be utilized to implement stochastic VMM of the second approach. 
Furthermore, adjusting read voltages in such a circuit allows for controlling SNR. 
Hence, the control of effective temperature, i.e. the slope of sigmoid probability function, enables efficient implementation of stochastic annealing in Boltzmann machines during runtime. 
The second approach's possible downside is slower operation because of lower read currents (which can be potentially addressed by utilizing external noise instead~\cite{Mahmoodi2019NatComm}).
Finally, the impact of noise quality on functional performance is another common concern. 
This issue has not been systematically studied yet, though Gaussian-like thermal or shot noise should be more advantageous for truly random operation.

\paragraph*{Spiking Neuron and Synaptic Plasticity}

Despite much recent progress in algorithms~\cite{Neftci2019, Tavanaei2019}, the most biologically plausible, spiking neural networks (SNNs)~\cite{Gerstner2002} are still inferior in the functional performance to simpler ANNs. 
If simpler ANNs would remain superior, the work of efficient SNN hardware could still be justified by the need to efficiently interface to the brain and/or model it, which in turn could lead to the development of higher-cognition artificial intelligence algorithms. 
An additional intriguing feature of SNNs is local weight update rules, requiring only information from pre- and post-synaptic neurons that could enable large-scale neuromorphic hardware with real-time training capabilities~\cite{Thakur2018}. 

In the simplest SNN models, the information is encoded in spike-time correlations~\cite{Gerstner2002}, while the network function is defined by the synaptic weights, which are adjusted based on the relative timing of spikes that are passed via synapses. 
In addition to VMM, the essential operations in SNNs are leaky-integrate-and-fire (LIF) functions performed by neurons and various types of synaptic plasticity, such as short-term plasticity (STP) and long-term potentiation (LTP), and spike-timing-dependent-plasticity (STDP)~\cite{Gerstner2002}. 
LIF neurons mimic the dynamic processes in the neuronal membrane, while synaptic plasticities mimic learning and memory mechanisms in biological networks. 
For example, STP is a temporary change in the synaptic strength implementing a short-term memory. 
Without immediate reinforcement of synaptic weight adjustment, the memory would be lost, i.e., the synaptic weight would relax to the original equilibrium state. On the other hand, the frequently repeated spiking stimulus causes long-term memory, e.g., permanent potentiation via the LTP mechanism. 
STDP is a time-dependent specialization of Hebbian learning. 
Its specific goal is to strengthen the synaptic efficiency when pre- and post- synaptic spikes happen in the expected causal temporal order and weaken it otherwise. 

A compact implementation of LIF neurons with biological, ms-scale integration times using conventional circuit technology is challenging because of the large capacitors that are required. Leaky integration circuits utilizing volatile memristors (e.g., based on filamentary~\cite{Zhang2018Small}, interfacial~\cite{Lashkare2018}, and Mott insulator \cite{Adda2018} switching mechanisms) have been suggested to address this problem. 
In such implementations, the integrated current is encoded with a conductive state of the volatile memory device. 
Neuron spiking functionality was demonstrated with threshold-switching (volatile) memory devices that feature S-type negative differential resistance (NDR) $I$-$V$ characteristics~\cite{Pickett2013}. 
This approach's general idea is similar to the oscillator circuits based on S-type (NDR) device connected to a resistor-capacitor circuit \cite{Kesim2019}. 
LIF neurons based on spin-torque magnetic memories were simulated in Ref.~\cite{Sengupta2016}. 
In such a neuron,  spin-torque oscillations are employed to generate spikes, while incremental magnetization and its relaxation mimic integration and leakage, respectively.

STP to LTP transition has been emulated with solid-state-electrolyte devices – see, e.g., original work in Ref.~\cite{Ohno2011} and more recent work on ``diffusive" memristors~\cite{Wang2017NatMat}. 
Specifically, the short and infrequent write pulses result in the formation of thin filaments, which are unstable and quickly dissolve, representing a short memory. 
However, a thicker and more stable filament can be formed by applying repeated and/or longer write pulses, thus mimicking transition to the LTP.  
Different STDP window implementations, e.g., using PCM~\cite{Kuzum2011} or metal-oxide ReRAM~\cite{Prezioso2016} devices, have been suggested by carefully selecting the shape of pre and post-synaptic write voltage pulses---see a comprehensive review of the emulated synaptic plasticity with memristive devices in Refs.~\cite{Serrano-Gotarredona2013, Saighi2015}.

Several small-scale spiking neuromorphic systems based on emerging device technologies were demonstrated, including coincidence detection via STDP mechanism based on metal-oxide memristors~\cite{Prezioso2018, Pedretti2017} and temporal data classification with diffusive memristors~\cite{Wang2018NatElec}. 
However, the overall progress in such advanced hardware has been much slower compared to simpler ANNs inference accelerators. 
The main reason is more demanding functionality from emerging devices in such applications and hence the more severe impact of device variations on the SNN operation and performance. 
For example, SNNs rely on fixed-magnitude spikes to update the conductance of multiple devices in parallel. 
Because of that, change in the conductances could vary drastically even with minor variations in $I$-$V$'s switching voltages, which in turn leads to very significant variations in STDP characteristics~\cite{Prezioso2018}. 
On the other hand, as already mentioned above, the implementation of simpler ex-situ trained ANNs is much less challenging because the write amplitude voltages in such networks can be adjusted uniquely for each device based on the feedback information during conductance tuning~\cite{Alibart2012}. 

Superconductor circuits, e.g., based on rapid single flux quantum (RSFQ) variety~\cite{Likharev1991}, are naturally suited for spiking circuits due to information encoding in SFQ voltage pulses. 
For example, Josephson Junction spiking neurons operating at up to 50\unit{GHz} range have been demonstrated in Ref.~\cite{Segall2017}. 
The historical challenges of such an approach include inferior fabrication technology (which may finally change given the enormous investments in superconductor quantum computing), the low-temperature operation that limits its applications, and the lack of efficient analog memory circuits \cite{Likharev2012}. 
The photonic spiking neural networks (e.g., Ref.~\cite{Feldmann2019}) and hybrid superconductor / optoelectronic neuromorphic circuits~\cite{Buckley2017} share the same challenges of the already discussed photonic neuromorphic inference approaches.

\paragraph*{Reservoir Computing}

Due to intrinsic memory properties, recurrent neural networks, such as Google Neural Machine Translation model, are especially suitable for processing sequential or temporal data. Reservoir computing (RC) networks are a special type of efficiently learning recurrent networks~\cite{Lukosevicius2009}, that were motivated by cortical information processing~\cite{Maas2004}. 
Among its variants are liquid state machines~\cite{Maass2002}, which is a spiking RC network, and echo state networks~\cite{Jaeger2001}, an RC based on a very sparse recurrent network. 
The main component in RC networks is a reservoir, which is a nonlinear recurrent network that maps inputs into a higher-dimensional spatio-temporal representation and has the property of a fading memory of the previous inputs and network states. 
Another component is a readout layer, which maps the intermediate state to the outputs. 
All connections in the reservoir are fixed and only weights in the readout layer are trainable. 
Because of that and sparse intermediate representation, faster and online algorithms can be employed for training such networks, which is a primary strength of this approach.

Though both readout and the reservoir can also be realized with the discussed analog VMM circuits, intriguing opportunities for implementing the reservoir are presented by nonlinear physical phenomena in superconductor, magnetic, and photonic devices~\cite{Tanaka2019}. 
For example, spoken vowel recognition was demonstrated with RC in which the reservoir was implemented with four coupled MTJ-based spin-torque oscillators (STO)~\cite{Romera2018}. 
In such a demo, the temporal input corresponding to spoken vowels is first converted to the frequency domain, which is in turn mapped to the corresponding DC bias currents that are applied to the MTJ devices. 
The induced voltage on the STO devices is used as an output of the reservoir. 
The reservoir utilizes the nonlinear dependence of the frequency of STOs on the DC current and the history-dependent transient motions of the MTJ's free layer spins spin. 

Various photonic reservoirs have been suggested~\cite{Shasti2021}, e.g. utilizing transient properties of optical systems with time-delayed feedback~\cite{Bruiner2013}, or relying on superimposing lights that passively circulates via waveguides, splitters and combiners, and nonlinear conversion to the electronic domain~\cite{Vandoorne2014}, to achieve high-dimensional response. The dynamics in the superconductor circuits are recently studied for efficient and extremely fast reservoir implementation~\cite{Rowlands2021}. 
Specifically, the proposed reservoir is based on a Josephson transmission line (JTL) formed by a chain of biased JJs. 
An input pulse from one end of the JTL causes a rapid cascade of junction phase slips that propagate SFQ pulse to the other end. Because JJs modulate each others' currents, a complex dynamical state is achieved.

There are several general concerns with RC computing approaches. On the algorithmic level, RC is inferior in performance to state-of-the-art approaches and it is unclear whether without further algorithm improvements such a handicap can be outweighed by the advantages of online training. 
The main concern for various hardware implementations is again related to the device variations, e.g., whether the hardware would be able to produce repeatable results when applying the same input. 
An additional concern for magnetic devices is the limited coupling between devices which could impact the effectiveness of the reservoir. 

\paragraph*{Hyperdimensional Computing / Associative Memory}

Hyperdimensional computing~\cite{Kanerva2009} circuits have been recently demonstrated with ReRAM~\cite{Li2016IEDM} and PCM~\cite{Karunaratne2020} devices. 
The low-level operation in hyperdimensional computing is closely related to that of associative or content addressable memory~\cite{Hertz1991}. 
Specifically, at the core of such an approach is an associative memory array circuit that outputs the closest, in a Hamming distance sense, memory row entry to a binary input vector serving as a search key. 
Assuming symmetric binary representation, with $-1$ and $+1$ encoding, Hamming distance is linearly related to a dot product, i.e., equal to output vector length minus dot product between the input vector and the stored memory row values. Therefore, the critical functionality in hyperdimensional computing is again a VMM operation. After the VMM operation has been completed, its results are passed to the winner-take-all circuit~\cite{Hertz1991} (which is a harder version of a softmax function~\cite{Bridle1989}) that determines the element with the smallest Hamming distance while discarding all other outputs. 
The additional simplification is that both input and weights in VMM are binary. 

In principle, binary VMM can be more efficiently implemented in hardware than its fully analog version. Similar to binary neural networks~\cite{Simons2019}, the apparent tradeoff is a worse functional performance of hyperdimensional computing.
Another essential feature of hyperdimensional computing is the suitability for fast ``one-shot" or incremental learning~\cite{Kanerva2009} though at the cost of having a much more redundant memory array. 
Note that fast ``one-shot” learning is not unique to hyperdimensional computing. For example, Hebbian learning and its many variants used in training associative neural networks have recursive form and are naturally incremental in that the weights can be modified only based on current weight values and the new pattern stored in the network~\cite{Hertz1991}.  

\paragraph*{Concluding Remarks} Many emerging devices and circuit technologies are currently being explored for neuromorphic hardware implementations. 
Neuromorphic inference accelerators utilizing analog in-memory computing based on floating gate memories are perhaps the closest to widespread adoption, given the maturity of such technology, the practicality of its applications, and competitive performance as compared to conventional (digital CMOS) circuit implementations. 
Comparing the performance prospects of other neuromorphic approaches is not straightforward because many proposals target algorithms with inferior functional performance, especially those closely mimicking the brain's operation. 
Baring a substantial breakthrough in ML algorithms or the emergence of new applications that could benefit from high-performance low-accuracy neuromorphic hardware, the inferior functional performance may limit the practicality of other approaches. 
The main challenge, much more so for advanced neuromorphic computing concepts, remains significant variations in the operation of emerging devices.

\clearpage

%\textbf{ASIC}
%
%It may be similar to FPGA, but the general impression is that the abstraction push may stop at HLS because of a more significant emphasis on verification (too many abstraction levels complicate the equivalence checking between those levels)
%
%\subsubsection{Software Integration}
%This section should answer the following questions:
%\begin{itemize}
%    \item How to integrate a heterogeneous-computing system in a broader scientific setup.
%    \item SW eng.
%    \item HPC way of things
%\end{itemize}
%heterogeneous-computing system is a good terminology. One good example is of Versal ACAP approach from Xilinx. May be helpful for FastML community as well.

%%%%%%%%%%%%%%%%%%%%%%%%%%%%%%%%%%%%
\pagebreak
% \addtocontents{toc}{\setcounter{tocdepth}{5}}
\section{Outlook}
\label{sec:outlook}
% As we discussed, there are lots of works on NN for science.  We have ways of doing X, Y, and Z.  And it is clear that, as we go forward, high throughput and low latency machine learning techniques will be critical for future Science deployments.  However, we still need innovation on A, B, and C.

This report has laid out exciting applications of fast ML to enable scientific discovery across a number of domains.  
This is a rapidly developing area with many exciting new studies and results  appearing often.  
However, this is a relatively young area rich with potential and a number of open challenges across a number of fields.  
Beyond what has been laid out in the report, we hope that the discussion of scientific use-cases and their overlaps will provide readers with the inspiration to entertain and pursue additional applications.  

In Section~\ref{sec:technolog_sota}, we provided an overview of techniques for developing powerful ML algorithms that need to be operated in high throughput and low latency environments.  
This includes both system design and training as well as efficient deployment and implementation of those ML models.  
Implementation in hardware is discussed under two main categories---current conventional CMOS and more speculative beyond CMOS technologies.  In the conventional CMOS case,  in light of the end of Moore's Law, the recent emphasis has been focused on advanced hardware architectures designed for ML.  
We gave an overview of popular and emerging hardware architectures and their strengths and shortcomings. 
A key area of importance for the multitude of hardware is their codesign of a given ML algorithm for specific hardware including the architecture and programmability of that algorithm.  
An example of a particularly relevant and important hardware platform is for FPGAs and that is the use-case discussed in Section~\ref{sec:codesign}.  
Finally, we concluded with an overview of beyond CMOS technologies which offer exciting and ultra-efficient technologies on which we can implement ML models.  
While these technologies are speculative, they offer potential orders of magnitude improvement over conventional technologies.

Both ML training and deployment techniques and computer architectures are extremely rapidly moving fields with new works appearing at a pace difficult to keep up with, even for this report.  
While new methods are being introduced continuously in both spaces, it is particularly important to understand the codesign of new algorithms for different hardware and the ease of use of the tool flows for deploying those algorithms.  
Innovations here will allow rapid and broad adoption of powerful new ML hardware.  
In the case of beyond CMOS technologies, these practical considerations are important as well as considering the maturity of the technology, integration into computing architectures, and how to program such devices.  

We look forward to revisiting these topics in the near future to see how quickly advances may come in applications, ML techniques, and hardware platforms---and most importantly their confluence to enable paradigm-shifting breakthroughs in science.

%%%%%%%%%%%%%%%%%%%%%%%%%%%%%%%%%%%%
\pagebreak
\section*{Acknowledgments}
We acknowledge the Fast Machine Learning collective as an open community of multi-domain experts and collaborators. 
This community was important for the development of this project.

%%%%%%%%%%%%%%%%%%%%%%%%%%%%%%%%%%%%
%%%%%%%%%%%%%%%%%%%%%%%%%%%%%%%%%%%%
\clearpage
\bibliographystyle{frontiersFPHY} %needed for correct ordering
\bibliography{references,references-amir,references-dmitri,references-guo}

\end{spacing}

\end{document}